\documentclass[10pt]{article}
\usepackage[accepted]{tmlr}

\usepackage{amsmath,amsfonts,bm}

\def\eqref#1{equation~\ref{#1}}

\def\1{\bm{1}}

\DeclareMathAlphabet{\mathsfit}{\encodingdefault}{\sfdefault}{m}{sl}
\SetMathAlphabet{\mathsfit}{bold}{\encodingdefault}{\sfdefault}{bx}{n}

\usepackage{hyperref}       
\usepackage{url}            
\usepackage{booktabs}       
\usepackage{nicefrac}       
\usepackage{microtype}      
\usepackage{xcolor}         
\usepackage{graphicx}       
\usepackage{adjustbox}      
\usepackage{enumitem}       
\usepackage{listings}
\usepackage{subcaption}
\usepackage{wrapfig}
\usepackage{caption}
\usepackage{amssymb}
\usepackage{multirow}
\usepackage{amsthm}
\usepackage{longtable}
\usepackage{array}
\usepackage{arydshln}
\usepackage{tcolorbox}
\usepackage{mathtools}

\newsavebox{\leftcolbox}
\newcommand{\refAppendix}[1]{Appendix~\ref{#1}}

\title{The Scaling Properties of Implicit Deductive Reasoning in Transformers}

\author{\name Enrico Vompa \email envomp@taltech.ee \\
\name Tanel Tammet \email tanel.tammet@taltech.ee \\
\addr Applied Artificial Intelligence Group \\
Tallinn University of Technology, Estonia}

\def\month{9}  
\def\year{2026} 
\def\openreview{\url{https://openreview.net/forum?id=M91oDrW4i7}}

\begin{document}

    \maketitle

    \vspace{-1em}
    \begin{abstract}
        We investigate the scaling properties of implicit deductive reasoning over Horn clauses in depth-bounded Transformers.
        By discouraging the reliance on statistical shortcuts via counterfactual data augmentation,
        and promoting the learning of shared reasoning primitives across direct and CoT modes,
        we find that in sufficiently deep models with a bidirectional prefix mask,
        implicit reasoning approaches explicit CoT performance across graph topologies and problem widths,
        though CoT remains necessary for depth extrapolation.
        These findings represent a step toward achieving better compositional reasoning in Transformers.
        The code and models to reproduce this work are available at: \href{https://github.com/envomp/Implicit-Deductive-Reasoning-in-Transformers}{https://github.com/envomp/Implicit-Deductive-Reasoning-in-Transformers}
    \end{abstract}

    \section{Introduction}

    Formal logic, such as deduction over Horn clauses,
    provides a foundation for systematic reasoning, offering the guarantees of truth-preservation and compositionality.
    To enforce such guarantees, classical logic solvers, such as automated reasoning systems,
    often rely on unbounded search and global planning to navigate complex or undecidable spaces~\citep{RobinsonVoronkov2001, russel2010, Boolos2007Computability}.
    Early work suggested that Transformers could simulate logical algorithms used within these systems directly.
    \cite{clark2020transformers} and~\cite{tafjord2021proofwriter} provided the first empirical demonstration that Transformers could operate as soft theorem provers,
    successfully generalizing to reasoning depths unseen during training,
    and~\cite{talmor2020leap} bridged explicit logical rules with implicit, pre-trained world knowledge.

    However, subsequent work revealed that these models often relied on spurious features specific to their training distributions,
    failing to achieve true out-of-distribution generalization~\citep{2023paradox}.
    Fundamentally, rather than faithfully executing sequential algorithms,
    where the causal structure mirrors the underlying logical structure of the problem~\citep{creswell2022faithful},
    neural networks tend to learn simple,
    low-complexity approximations~\citep{kalimeris2019sgd, perez2018deep}.
    This bias manifests as shortcut learning~\citep{geirhos2020shortcut},
    where models exploit spurious features or rely on parallelizable computations to bypass sequential execution~\citep{marconato2023not, liu2023transformers, wang2024do}.
    To address these shortcut-inducing biases, we introduce three complementary interventions:
    the \texttt{r2} heuristic to decorrelate spurious features from provability;
    bidirectional prefix masking for problem-wide visibility;
    and a single-sequence corrective training objective to align the reasoning primitives shared by direct and chain of thought (CoT)~\citep{wei2022cot} predictions.

    Under standard complexity assumptions~\citep{merrill2022saturated},
    depth-bounded models cannot perfectly solve Horn-satisfiability or do AI planning directly~\citep{merrill2023parallelism},
    yet their specific scaling properties remain unclear.
    We investigate these properties on deductive reasoning over Horn clauses
    and identify scaling requirements for faithful approximations as a function of attention~\citep{vaswani2017attention} layer count.
    Understanding these properties is important for assessing how these models might extend to first-order logic
    or faithful natural language reasoning~\citep{turpin2023language, luo2024reasoning, sui2025fidelis}.

    Motivated by evidence that attention layers can recover latent causal structures~\citep{nichani2025causal},
    we find that scaling model depth is essential for achieving robust generalization to unseen distributions.
    While directly learning functions that require serial computation in transformers is computationally intractable for gradient descent (requiring exponential iterations)~\citep{feng2023towards},
    they can be learned efficiently with CoT~\citep{li2024chain, kim2025transformers}.
    In our setting, we observe that the corrective objective makes this optimization tractable for direct prediction,
    allowing CoT to act as an empirical upper bound.

    Our contributions are as follows:
    \begin{itemize}[noitemsep, topsep=0pt, parsep=2pt, partopsep=0px, leftmargin=*]
        \item[-] We systematically mitigate shortcut-inducing biases,
        and improve out-of-distribution performance through the \texttt{r2} heuristic, bidirectional prefix masking, and a corrective objective.
        \item[-] Within this setting, we find increasing model depth closes the implicit--explicit reasoning gap across graph topologies and problem widths,
        though CoT remains necessary for depth extrapolation.
        \item[-] To advance mechanistic interpretability, we introduce a Procrustes alignment method that factors out orthogonal transformations across layers,
        revealing that linearly separable features remain topologically consistent and decodable throughout the network.
    \end{itemize}

    \vspace{-0.5em}
    \section{Related work}\label{sec:related_work}

    A complementary line of research delegates deduction to external theorem provers
    by using LLMs as semantic parsers~\citep{olausson2023linc, PanLogicLM23}.
    While effective, these approaches do not directly improve the model's implicit deductive competence.
    Meanwhile, modern open-source models~\citep{qwen3technicalreport, magistal},
    driven by advancements in reinforcement learning, achieve strong deductive accuracy with CoT\@.
    Yet, their direct performance continues to lag behind in symbolic reasoning~\citep{sprague2025to},
    which we also observe (\refAppendix{sec:appendix_open_source_baselines}).
    Furthermore, as~\citep{wu2025on} showed, reinforcement learning primarily preserves rather than expands the base model's reasoning coverage,
    which we also observe (\refAppendix{sec:appendix_rl_failure}).

    Methodologically,
    we build on counterfactually-augmented data~\citep{kaushik2020learning} and factual–counterfactual balancing~\citep{xu2024counter},
    which is interpreted through the lens of information bottleneck
    that argues models achieve robustness by compressing their hidden representations to filter out spurious mutual information.
    Instead, we show that by decorrelating such features from labels,
    these near-collisions start competing with standard regularization.

    We differentiate between causal~\citep{Radford2018ImprovingLU, Radford2019LanguageMA} and non-causal decoders~\citep{2020t5, devlin2019bert}.
    It has been shown that the latter, which introduces a bidirectional prefix mask,
    outperforms the former after masked language modeling and subsequent multitask fine-tuning~\citep{wang2022architecturepretraining}.
    However, we find that in our setting, an autoregressive objective is sufficient for a non-causal decoder to outperform a causal decoder.
    Traditionally,
    multitask setting is a common method for jointly training implicit and explicit reasoning capabilities~\citep{narang2020wt5, hsieh2023distilling}.
    Inspired by recent advances in structured attention mechanisms designed to isolate concurrent reasoning pathways~\citep{zheng2025parallel},
    we propose a single-sequence formulation, which unifies both reasoning modes.

    To understand the internal dynamics of language models,
    various probing techniques have been developed~\citep{alain2018probe, tenney2019nlp}.
    However, a primary limitation of standard probes is that their accuracy degrades across network depth;
    a probe trained on representations at layer $l$ often fails to generalize to layer $l+1$ due to continuous transformations of the hidden state~\citep{belrose2025lens}.
    To address this, we develop an alignment method using Procrustes~\citep{wang2008procrustes, razzhigaev2024secretly}
    that accounts for the orthogonal transformation of the representation space.
    By doing so, we find that linearly separable information remains topologically consistent and decodable across all layers in our setting,
    allowing us to rotate intermediate representations into both the input and output spaces to track the evolution of our models' logical state.

    \vspace{-0.5em}
    \section{Methodology}
    \label{sec:methodology}

    We formulate deductive reasoning as a problem in propositional logic restricted to definite clauses (Horn clauses).
    Following the framework of~\cite{2023paradox}, an instance is defined as a tuple consisting of facts, rules, a query, and a boolean label.
    Let $\mathcal{P}$ be a set of predicates, which are propositional variables that evaluate to True or False.
    A \textbf{rule} is a definite clause of the form $A_1 \land A_2 \land \dots \land A_n \rightarrow B$,
    where the premises $A_i \in \mathcal{P}$ and the conclusion $B \in \mathcal{P}$.
    A \textbf{fact} is a predicate explicitly given as True, functionally equivalent to a rule with an empty body.
    Given a propositional theory $T$ comprising a set of facts and rules,
    a \textbf{query} predicate $Q \in \mathcal{P}$ is logically provable if it is either a known fact in $T$,
    or if there exists a rule $A_1 \land \dots \land A_n \rightarrow Q$ in $T$ where all premises $A_i$ are also provable.
    The \textbf{label} denotes whether the query is provable from $T$.

    \subsection{Model architecture}\label{subsec:model_architecture}
    We train a Llama 3-based~\citep{meta2025llama} decoder-only Transformer ($L=8$, $d_{\text{model}}=256$, full hyperparameters in \refAppendix{sec:appendix_hyperparameters}).
    We use a learned, compositional \textbf{type embedding} system to decouple logical roles from token identity,
    where the final input representation is constructed by summing the token embedding with active semantic role embeddings
    (e.g., tagging a token as both \texttt{rule} and \texttt{conclusion}, see \refAppendix{sec:appendix_type_embeddings}).
    This explicit grounding provides the model with a more direct handle on topology, reducing parsing overhead.
    During training, we find that applying weight decay to normalization parameters is essential for convergence in this domain (\refAppendix{sec:appendix_norm_ablation}).

    \subsection{Logic datasets}\label{subsec:datasets}
    \begin{minipage}[b]{0.50\textwidth}
        The dataset generators proposed by~\citep{2023paradox} allow us to control graph complexity and topology.
        The \textbf{Rule-Priority (RP)} algorithm generates problems with an entangled structure by uniformly sampling facts, rules, and a query.
        Ground truth is computed via \textbf{forward-chaining}, a process where rules are iteratively applied to facts to derive all reachable conclusions.
        Within this dependency graph, we define the \textbf{proof depth} ($\delta$) as the index of the forward breadth-first search (BFS) layer in which the query is first derived.
        In contrast, the \textbf{Label-Priority (LP)} algorithm constructs problems with a hierarchical backbone.
        It assigns truth values to predicates (propositional variables) in ordered levels and subsequently obscures this structure by injecting random distractor rules.
        The final third variant, \textbf{LP*}, modifies LP by increasing the number of cyclic dependencies.
    \end{minipage}
    \hfill
    \begin{minipage}[b]{0.50\textwidth}
        \vspace{-20pt}
        \centering
        \includegraphics[width=\textwidth]{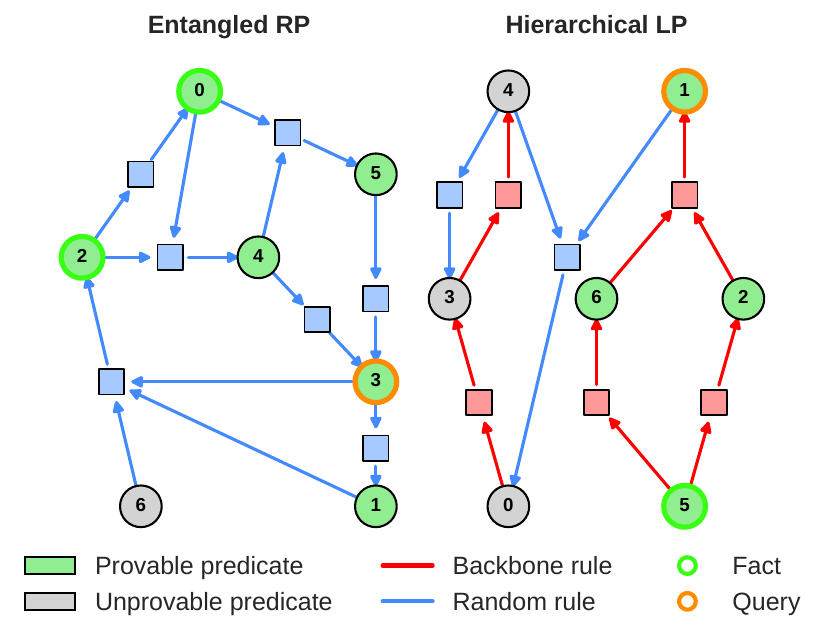}
        \captionof{figure}{Proof depth $\delta$ on RP and LP problems.}
        \label{fig:problem_visualization}
    \end{minipage}

    Figure~\ref{fig:problem_visualization} visualizes proof depth $\delta$,
    where circles represent predicates and squares represent rules,
    with incoming edges as premises and the outgoing edge as the conclusion.
    In the RP graph (left), the shortest derivation starts from facts $\textbf{0} \text{ and } \textbf{2}$, and follows
    $\textbf{0} \land \textbf{2} \to \textbf{4}$, and $\textbf{4}\to\textbf{3}$, resulting in a proof depth of $\delta=2$.
    Such forward-chaining depth can be calculated for all derived predicates, not just the query;
    for instance, deriving \textbf{1} requires continuing the chain from \textbf{3}, yielding a depth of 3.
    Conversely, in the LP graph (right), the fact \textbf{5} allows us to derive \textbf{6} and \textbf{2}, which together imply \textbf{1}.
    This results in a proof depth of $\delta=2$.
    For CoT training data, these trajectories are serialized into reasoning traces.
    When multiple rules apply simultaneously, the solver selects the first valid conclusion via a left-to-right scan.
    Traces for provable queries terminate once query is derived, whereas unprovable queries terminate once the graph reachability is exhausted.

    \subsection{Mitigating complexity shortcuts}
    \label{subsec:complexity_shortcuts}
    Random rule generation is biased toward shallow paths,
    as the frequency of paths decays exponentially relative to path length~\citep{albert2002statistical} (\refAppendix{sec:appendix_sampling_bias}).
    To mitigate this bias, and maintain balanced class distributions~\citep{abramov2025grokking}, we balance datasets by depth.

    \begin{minipage}[b]{0.59\textwidth}
        For provable samples, depth is directionally invariant;
        the shortest derivation length remains the same whether traversed forward from the facts or backward from the query.
        For \textbf{unprovable} samples, complexity is defined by the search effort required to verify non-existence, which is algorithm-dependent.
        Following prior work~\citep{2023paradox},
        these samples are typically balanced using backward-chaining depth-first search (DFS) depth,
        but that leaves the \textbf{forward BFS depth} heavily skewed toward shallow values on random graphs (Figure~\ref{fig:depth_imbalance}),
        which enables a shortcut where graph complexity correlates with provability.
    \end{minipage}
    \hfill
    \begin{minipage}[b]{0.38\textwidth}
        \centering
        \includegraphics[width=\linewidth]{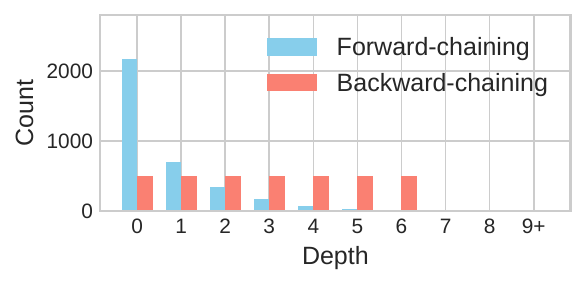}
        \vspace{-2em}
        \captionof{figure}{RP dataset balanced by backward-chaining depth.}
        \label{fig:depth_imbalance}
    \end{minipage}

    \subsection{Defining logical depth}
    To mitigate this shortcut, we balance unprovable samples by their maximum forward-chaining depth,
    incentivizing models to verify logical connectivity rather than using depth as a proxy for truth.
    Consequently, for evaluation, we use a unified \textbf{logical depth} metric $\delta$ representing the \textbf{required forward BFS horizon};
    for provable samples, $\delta$ is the layer where the query is found;
    for unprovable samples, $\delta$ is the layer where the reachable graph is exhausted.

    \subsection{Dataset specifications}\label{subsec:dataset_specifications}
    Our training set contains problems with $N_{\text{pred}} \le 30$ and $\delta \le 6$ over a vocabulary of 150 predicates,
    balanced via rejection sampling to 50k items per $(\delta,\text{label})$ bucket (500 for evaluation).
    To evaluate out-of-distribution generalization, we selectively relax these bounds,
    testing on problem spaces with $N_{\text{pred}} \le 60$ and logical depth up to $\delta \le 12$,
    where the average number of facts and rules scales proportionally with $N_{\text{pred}}$.
    Rules contain 1 to 3 premises; full generation details are in \refAppendix{sec:appendix_dataset_generation}.

    \section{Manifold alignment}
    \label{sec:manifold_alignment}

\begin{figure}[htbp]
        \centering
        \includegraphics[width=0.9\linewidth]{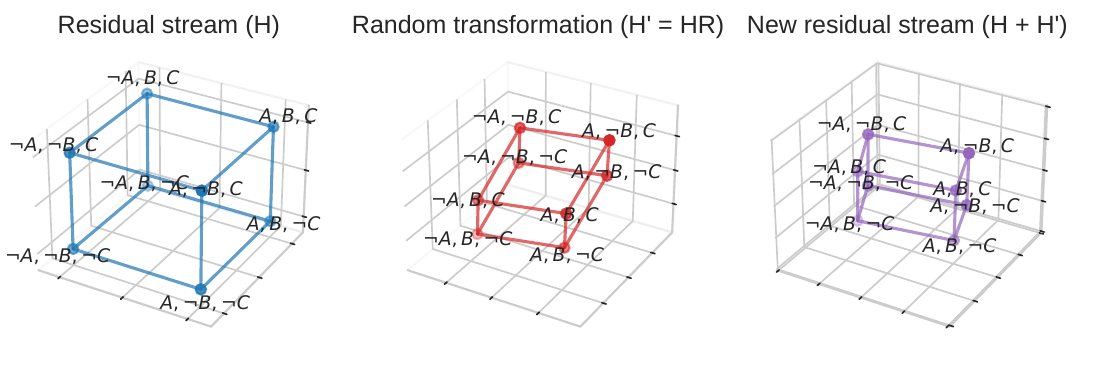}
        \vspace{-2em}
        \caption{LLM layer applies non-linear transformations to organize features, accompanied by orthogonal rotations of the representation space.
        Procrustes alignment enables factoring out such orthogonal transformations, so we can view $H$ from the basis of $H + H'$ and vice versa. (R is a random matrix)}
        \label{fig:orthogonal_transformation}
    \end{figure}

    To faithfully evaluate logical rules,
    the model must represent features necessary for causal deduction within the same $d$-dimensional hidden space.
    Prior work suggests this is achieved by packing features in linear combinations into superposition~\citep{prieto2026featuregeometry, elhage2022superposition}.
    We posit that to faithfully mirror underlying causal logic, necessary features must be linearly accessible.
    As logical deduction requires checking whether premises are met, these truth values cannot remain behind non-linearities;
    they must be linearly accessible.
    Furthermore, while it has been found that transformer decoders maintain near-perfect Procrustes similarity between sequential layers~\citep{razzhigaev2024secretly},
    the specific geometric arrangement of these features remains unclear.
    Our findings are indicative of non-linear transformations organizing features into separable subspaces,
    where orthogonal transformations rotate these representations across layers (Figure~\ref{fig:orthogonal_transformation}).

\begin{figure}[htbp]
        \centering
        \includegraphics[width=0.9\linewidth]{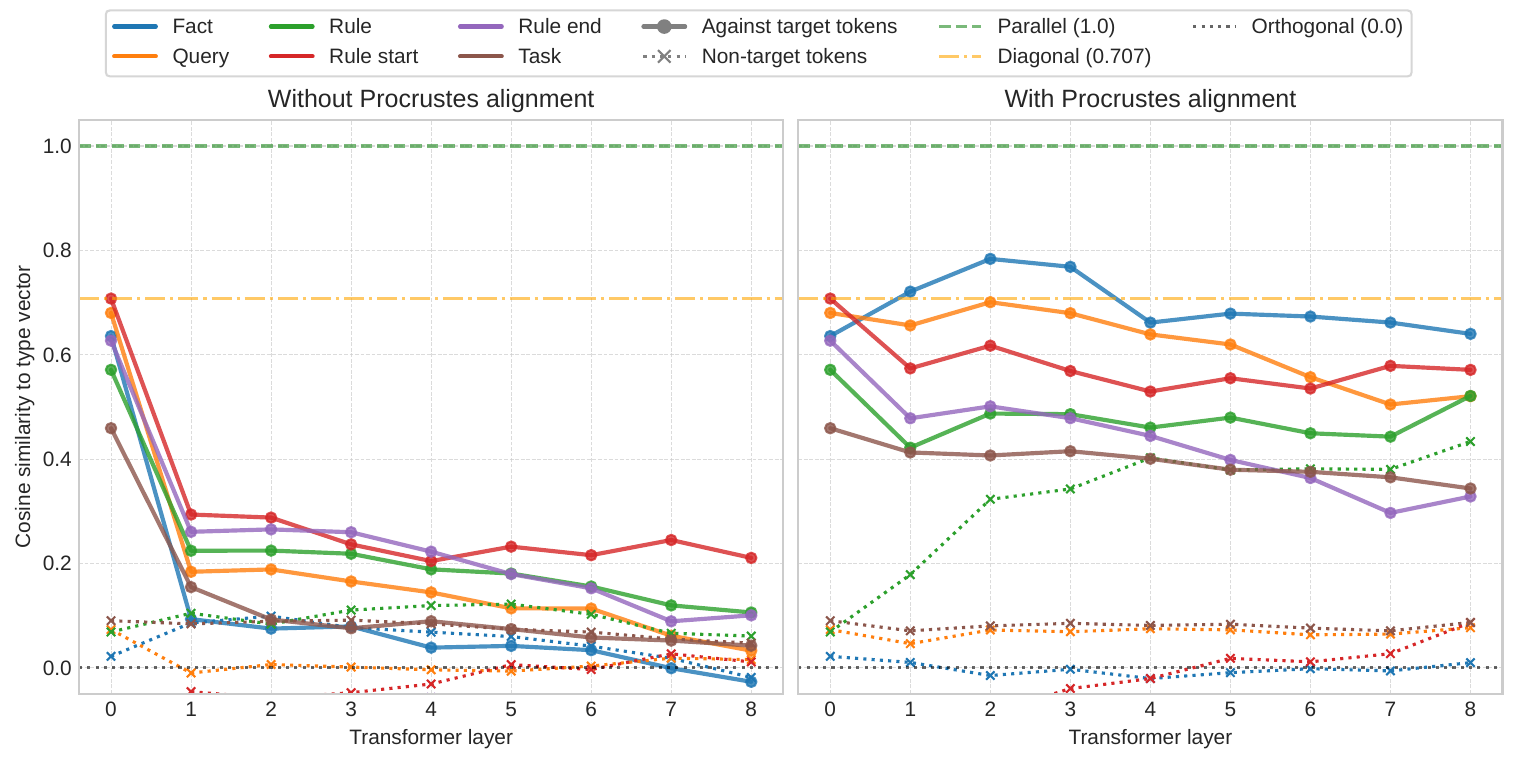}
        \caption{Cosine similarity of intermediate hidden states to type embeddings across transformer layers, consistent with an orthogonally transformed representation space.
        Similarities are averaged over 3 seeds with \texttt{corrective}, \texttt{bidirectional} and \texttt{r2} components, methodology in~\refAppendix{sec:appendix_procrustes_alignment}.}
        \label{fig:type_embedding_similarity}
    \end{figure}

    This enables us to develop a Procrustes alignment method to factor out these orthogonal transformations
    and recover linearly accessible information from these evolving subspaces.
    By aligning the intermediate states with a layer-specific transformations,
    we reveal a margin of separation between tokens with and without a semantic type present.
    This grounds our work in superposition hypothesis and by proxy in linear representation hypothesis~\citep{park2024linear, park2025the},
    where semantic types are linearly accessible even when not directly aligned with the input basis.
    As illustrated in Figure~\ref{fig:type_embedding_similarity}, raw hidden states exhibit a decrease in similarity across layers.
    However, when we apply Procrustes alignment~\citep{wang2008procrustes, razzhigaev2024secretly} to map the intermediate states back into the input space,
    much of the linear accessibility is recovered.

    \section{Architectural alignment}
    \label{sec:architectural_alignment}

    \subsection{Shortcut learning}
    \label{subsec:shortcut_learning}
    Reasoning shortcuts use concepts with unintended semantics~\citep{yang24ac}.
    While reasoning shortcut is the computational process of faulty reasoning,
    correlation between spurious features and the label is the input to that process~\citep{geirhos2020shortcut},
    which typically utilizes low-complexity circuits that dominate gradient flow during standard training~\citep{kalimeris2019sgd, perez2018deep}.
    We distinguish between two types of spurious features:
    \textbf{superficial features}, which are linearly accessible signals (token counts), and
    \textbf{structural features}, which are compositional properties that require non-linear computation to be recovered (graph topology).
    By decorrelating these features from provability,
    we substantially reduce the predictive power of spurious features.

    \subsection{The \texttt{r2} heuristic}
    \label{sec:r2_heuristic}
    Towards that end, we introduce the \texttt{r2} heuristic (\refAppendix{sec:appendix_r2_heuristic}),
    which constructs minimally different pairs $(C_1, C_2)$ with opposite labels yet nearly identical superficial statistics $\phi(C_1) \approx \phi(C_2)$.
    This mitigates reasoning shortcuts as linearly separating such near-collisions requires a large weight norm $\|w\|$,
    creating a conflict with standard regularization (\refAppendix{sec:appendix_r2_motivation}).
    This constraint increases pressure on the model to rely on \textbf{structural features} to resolve the ambiguity.

    In RP,
    to convert provable problems to unprovable,
    we employ the \textbf{greedy iterative modification} strategy,
    which iteratively removes rules or facts,
    greedily preserving logical depth,
    until the graph becomes unprovable.
    Removals are balanced by adding components that maintain complexity,
    such as rules concluding the same predicate but using the highest-depth available premises.
    Conversely, to transform unprovable problems to provable,
    we apply a \textbf{single-rule addition} that completes a proof path constrained to the problem's target logical depth
    and balance it by pruning a pre-existing rule that also concludes the query.

    As shown in Table~\ref{tab:correlation_analysis_brief},
    the original dataset exhibits strong label correlations for superficial and structural features,
    providing gradients for shortcut learning; full table in \refAppendix{sec:appendix_correlation_analysis}.
    Adversarial balancing via \texttt{r2} diminishes these aggregate correlations and weakens individual feature predictability,
    though non-linear models may still exploit higher-order features or artifacts.

    \begin{table}[h!]
        \centering
        \caption{
            Pearson correlation between features and the ground-truth label on RP dataset.
            The \texttt{r2}-augmented data significantly diminishes the correlations present in the final combined dataset.}
        \label{tab:correlation_analysis_brief}
        \begin{adjustbox}{max width=\linewidth}
            \begin{tabular}{@{}lrrr@{}}
                \toprule
                Feature                   & Original & \texttt{r2}-augmented & Combined \\
                \midrule
                num\_rules                & 0.277    & -0.247                & 0.019    \\
                query\_total\_occurrences & 0.431    & -0.298                & 0.078    \\
                branching\_factor         & 0.082    & -0.131                & -0.021   \\
                \bottomrule
            \end{tabular}
        \end{adjustbox}
    \end{table}

    \subsection{Bidirectional visibility}
    Causally masked models are brittle to premise ordering,
    as they assume physical order matches logical steps~\citep{chen2025premise}.
    We address this with a bidirectional prefix mask~\citep{narang2020wt5},
    enabling all-to-all attention within the problem statement to mitigate sequential bias.
    Our error analysis in \refAppendix{sec:appendix_error_breakdown} shows that this approach uniformly reduces the volume of both hallucinations and missed deductions,
    with hallucinations remaining dominant until the \texttt{r2} heuristic is introduced.
    Loss is calculated only on the output,
    as the randomly generated problem statement lacks learnable causal dependencies.

    \subsection{The corrective objective}
    By explicitly unrolling the computational graph,
    CoT expands the models' expressive power~\citep{merrill2024the},
    where it inadvertently bypasses many of the shortcuts that models typically exploit during direct generation~\citep{wang2024do}.
    Motivated by this, we explore whether direct prediction and CoT share reasoning primitives that can be aligned~\citep{he2025proto}.
    By imposing a step-by-step algorithmic bias on the direct prediction,
    the corrective objective promotes the learning of these shared reasoning primitives,
    more closely approximating the forward-chaining algorithm and outperforming the direct-only baseline,
    then intuitively, direct performance becomes limited by the quality of the signals coming from CoT traces.

    \begin{minipage}[b]{0.64\textwidth}
        Historically, single-task joint prediction uses a single sequence of $(x, y_{direct}, y_{cot})$,
        which first predicts the final answer and then the rationale, or vice versa~\citep{wiegreffe2021measuring}.
        A multi-task mixed dataset approach $\mathcal{D}_{mixed} = \{(x, \tau_{direct}, y_{direct})\} \cup \{(x, \tau_{cot}, y_{cot})\}$,
        where there are distinct training samples for direct and CoT traces, has been shown to outperform it~\citep{hsieh2023distilling}.
        However, that effectively doubles the number of training samples.
        Inspired by structured attention mechanisms designed to isolate concurrent reasoning pathways~\citep{zheng2025parallel},
        we introduce a single-sequence multi-task corrective format,
        which concatenates the targets into a single sequence $(x, \tau_{direct}, y_{direct}, \tau_{cot}, y_{cot})$ optimized via cross-entropy loss.
        To prevent information leakage, we apply an \textbf{isolated attention mask}
        that permits both task branches to attend to the bidirectional problem statement but prevents them from attending to each other
        (Figure~\ref{fig:corrective_mask}, where the \textcolor[HTML]{E67E22}{\textbf{problem statement}} is visible to all branches,
        and the \textcolor[HTML]{C0392B}{\textbf{direct branch}} is isolated from \textcolor[HTML]{27AE60}{\textbf{CoT branch}}).
        Both of these multi-task approaches perform comparably,
        while outperforming the single-task direct only baseline (\refAppendix{sec:appendix_multi_task_comparison}).
    \end{minipage}
    \hfill
    \begin{minipage}[b]{0.34\textwidth}
        \centering
        \includegraphics[width=\linewidth]{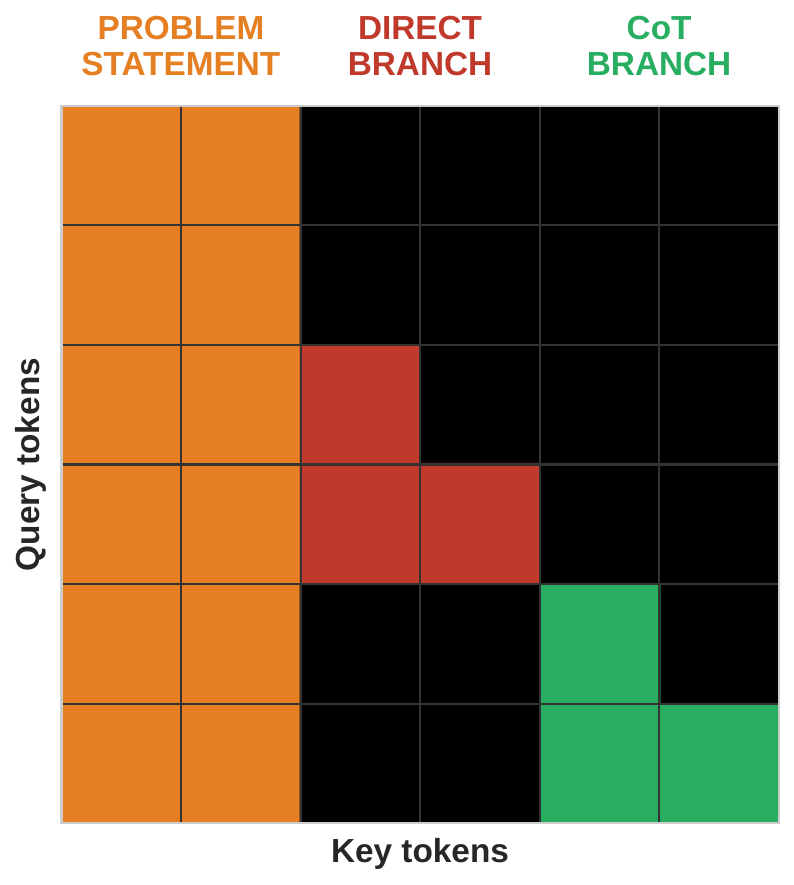}
        \captionof{figure}{Attention mask.}
        \label{fig:corrective_mask}
    \end{minipage}

    \newpage

    \section{Experimental results}
    \label{sec:experiments}

    We perform a factorial ablation across \texttt{corrective}, \texttt{bidirectional}, \texttt{r2}, and \texttt{ffn} components to identify their individual contribution.
    8-layer models are trained on the RP dataset ($N_{\text{pred}} \le 30, \delta \le 6$), and evaluated on the LP dataset;
    see~\refAppendix{sec:appendix_factorial_ablation} for other distributions.
    Table~\ref{tab:main_results_summary} reports the marginal contribution $\Delta(\text{component}) = avg_{\text{with}} - avg_{\text{without}}$ computed over all $2^4=16$ configurations per evaluation mode.

    \begin{table}[ht!]
        \centering
        \caption{Marginal contribution of components.
        Values denote the average percentage point difference attributable to each component.
        Evaluation across logical depths $\delta$, and predicate counts $N_{\text{pred}}$ on the LP dataset.
        \textcolor{green!70!black}{Green} indicates a statistically significant improvement ($p < 0.05$).}
        \label{tab:main_results_summary}
        \begin{adjustbox}{max width=\linewidth}
            \begin{tabular}{lcccc}
                \toprule
                & \multicolumn{2}{c}{$N_{\text{pred}} \le 30$} & \multicolumn{2}{c}{$N_{\text{pred}} \le 60$} \\
                \cmidrule(lr){2-3} \cmidrule(lr){4-5}
                Component
                & \multicolumn{1}{c}{$\delta \le 6$} & \multicolumn{1}{c}{$6 < \delta \le 12$}
                & \multicolumn{1}{c}{$\delta \le 6$} & \multicolumn{1}{c}{$6 < \delta \le 12$} \\
                \midrule
                direct w. corrective          & $\textcolor{green!70!black}{18.9 \pm 8.4}$ & $\textcolor{green!70!black}{5.3 \pm 4.6}$ & $\textcolor{green!70!black}{15.4 \pm 7.7}$ & $0.6 \pm 1.1$ \\
                direct w. bidirectional       & $\textcolor{green!70!black}{8.4 \pm 7.1}$  & $\textcolor{green!70!black}{2.8 \pm 2.5}$ & $\textcolor{green!70!black}{7.1 \pm 6.7}$ & $\textcolor{green!70!black}{1.0 \pm 0.7}$ \\
                direct w. r2 (w/ corrective)  & $\textcolor{green!70!black}{7.1 \pm 3.0}$  & $\textcolor{green!70!black}{9.2 \pm 2.3}$ & $\textcolor{green!70!black}{7.9 \pm 2.6}$ & $\textcolor{green!70!black}{7.8 \pm 0.9}$   \\
                direct w. r2 (w/o corrective) & $-0.6 \pm 7.8$                             & $2.2 \pm 2.0$                             & $-0.0 \pm 8.0$                             & $\textcolor{green!70!black}{7.8 \pm 1.3}$ \\
                direct w. ffn                 & $-2.4 \pm 5.7$                             & $-0.9 \pm 2.4$                            & $-0.5 \pm 5.4$                             & $-0.4 \pm 1.3$                            \\
                \midrule
                cot w. corrective             & $-0.8 \pm 6.3$                             & $-2.1 \pm 9.4$                            & $-1.2 \pm 5.2$                             & $-0.6 \pm 3.4$                            \\
                cot w. bidirectional          & $\textcolor{green!70!black}{6.0 \pm 4.9}$  & $5.7 \pm 7.9$                             & $\textcolor{green!70!black}{6.0 \pm 4.9}$ & $-0.0 \pm 3.3$  \\
                cot w. r2                     & $\textcolor{green!70!black}{3.7 \pm 4.3}$  & $\textcolor{green!70!black}{7.1 \pm 7.0}$ & $\textcolor{green!70!black}{4.3 \pm 4.7}$ & $\textcolor{green!70!black}{5.1 \pm 3.5}$ \\
                cot w. ffn                    & $-0.8 \pm 6.2$                             & $0.6 \pm 10.0$                            & $1.7 \pm 4.8$                              & $0.7 \pm 3.9$                             \\
                \bottomrule
            \end{tabular}
        \end{adjustbox}
    \end{table}

    In direct mode, the \texttt{corrective} objective is the most effective component,
    suggesting that aligning representations with algorithmic targets is important for simulating faithful forward-chaining.
    However, this benefit is contingent on the model possessing sufficient architectural depth ($L \ge \delta$, \refAppendix{sec:appendix_corrective_shallow}).
    More generally, computational capacity of the model must scale with the hardness of the task (\refAppendix{sec:appendix_high_level_analogoies}).
    Across modes, \texttt{bidirectional} mask is robust contributor via global visibility.
    When the necessary preconditions for embedding provability in intermediate steps only appear later in the sequence, which are not visible,
    then there is no way for models to respect the causality of logic when enforcing that the final answer must be provable.
    While removing spurious features using the \texttt{r2} generally improves generalization,
    as verified when controlling for training budget (\refAppendix{sec:appendix_sample_matched_r2_ablation}),
    it imposes a penalty on direct learning in complex scenarios (without \texttt{corrective}).
    Intuitively, if a model's capacity to minimize loss relies entirely on spurious correlations,
    eliminating these shortcuts leaves it without a viable learning signal.
    Conversely, \texttt{ffn} provides no systematic benefit,
    implying that increased point-wise capacity is difficult to utilize.

    \subsection{Recursive bias through Universal Transformer}
    \label{subsec:universal_transformer}

    To mirror logical recursion,
    we employ a Universal Transformer with a single weight-tied layer applied for $K=8$ iterations~\citep{dehghani2018universal},
    biasing the model toward learning a \textbf{next-step operator} homomorphic to graph traversal~\citep{giannou2023looped}.
    While point-wise capacity is difficult to utilize using the standard baseline architecture,
    this recursive bias facilitates the utility of \texttt{ffn} (Table~\ref{tab:universal_summary}, \refAppendix{sec:appendix_recursive_nature}).

    \begin{table}[h]
        \centering
        \caption{Universal and baseline direct performance on $\delta \le 6$ problems across predicate counts $N_{\text{pred}}$, ablated with \texttt{ffn} component.
        \textcolor{green!70!black}{Green} indicates a statistically significant improvement ($p < 0.05$).}
        \label{tab:universal_summary}
        \begin{adjustbox}{max width=\linewidth}
            \begin{tabular}{lcccccc}
                \toprule
                & \multicolumn{2}{c}{LP} & \multicolumn{2}{c}{LP*} & \multicolumn{2}{c}{RP (train)} \\
                \cmidrule(lr){2-3} \cmidrule(lr){4-5} \cmidrule(lr){6-7}
                Model $\quad \backslash \quad N_{\text{pred}}$  & $\le 30$                                  & $\le 60$                                  & $\le 30$                                  & $\le 60$                                  & $\le 30$       & $\le 60$       \\
                $\Delta$ (universal w. ffn - baseline w. ffn)   & $\textcolor{green!70!black}{6.8 \pm 3.5}$ & $\textcolor{green!70!black}{6.5 \pm 3.4}$  & $\textcolor{green!70!black}{1.6 \pm 0.8}$ & $\textcolor{green!70!black}{1.9 \pm 1.1}$ & $0.7 \pm 0.5$ & $-0.0 \pm 0.7$ \\
                $\Delta$ (universal w.o ffn - baseline w.o ffn) & $0.8 \pm 3.6$                             & $1.3 \pm 2.4$                             & $-0.2 \pm 1.7$ & $\textcolor{green!70!black}{1.6 \pm 0.8}$ & $-0.4 \pm 1.2$ & $-3.8 \pm 3.1$ \\
                \bottomrule
            \end{tabular}
        \end{adjustbox}
    \end{table}

    \begin{minipage}[b]{0.59\textwidth}
        \subsection{Closing the implicit--explicit gap}\label{subsec:scaling_trends}
        \vspace{1em}
        More generally, under standard complexity assumptions~\citep{merrill2022saturated},
        solving P-complete problems with a depth-bounded Transformer requires super-polynomial width growth in the worst case,
        making width-scaling an inefficient solution~\citep{feng2023towards, merrill2024a}~(\refAppendix{sec:appendix_scaling_trends}).
        We find the reliability of implicit reasoning in our setting scales primarily with the \textbf{number of layers ($L$)}.
        Even though the dataset size remained fixed, it was sufficient for convergence in increasingly deep models,
        demonstrating that corrective objective makes this optimization tractable (\refAppendix{sec:appendix_direct_tractability}).
        Figure~\ref{fig:scaling_performance_lp} compares direct and CoT evaluation modes on the LP dataset ($N_{\text{pred}} \le 30$).
        By scaling models from $L=8$ to $L=128$,
        we successfully close the gap between implicit and explicit reasoning within the training horizon ($\delta \le 6$)
        across graph topologies and problem widths ($p < 0.05$ for sustained non-inferiority; see \refAppendix{sec:appendix_scaling_trends}).
    \end{minipage}
    \hfill
    \begin{minipage}[b]{0.38\textwidth}
        \centering
        \includegraphics[width=\linewidth]{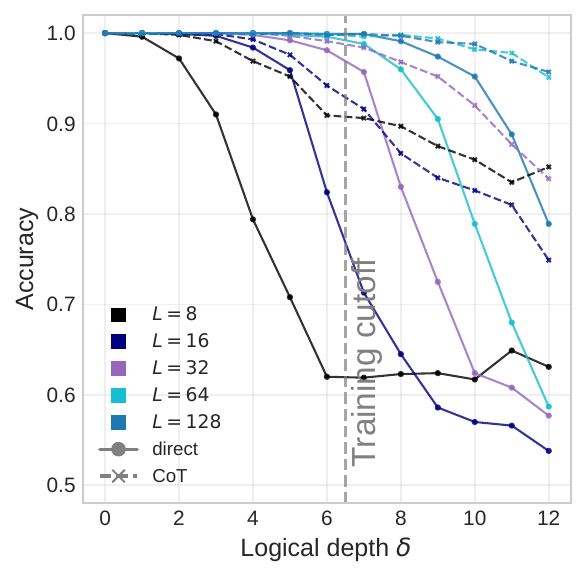}
        \captionof{figure}{Scaling model depth closes the gap within training horizon on LP.}
        \label{fig:scaling_performance_lp}
    \end{minipage}

    \begin{minipage}[t]{0.59\textwidth}
        \vspace{0pt}

        \subsection{Probing for traces of the successful approximation}\label{subsec:bfs_like_reasoning}
        \vspace{1em}
        To identify the learned approximation,
        we utilize Procrustes alignment to project intermediate states into the output space
        and apply a training-free linear probe to evaluate at which layer provability becomes decodable (\refAppendix{sec:appendix_decision_trace}).
        Hypothetically, a memory-efficient forward-chaining approximation distributes the state of proved predicates across the problem statement.
        We apply this probe to the hidden states of all tokens, evaluating the predictions against the ground-truth provability determined by a logic solver.
        Within the training horizon ($\delta \le 6$), these traces reveal a property of scale;
        as the number of layers increases, the provability of intermediate steps becomes increasingly apparent across the entire problem statement,
        reflected by a rising probe F1 score (Figure~\ref{fig:provability_probing_lp}, distinction by predicate counts $N_{\text{pred}}$).
    \end{minipage}
    \hfill
    \begin{minipage}[t]{0.38\textwidth}
        \vspace{-10pt}
        \centering
        \includegraphics[width=\linewidth]{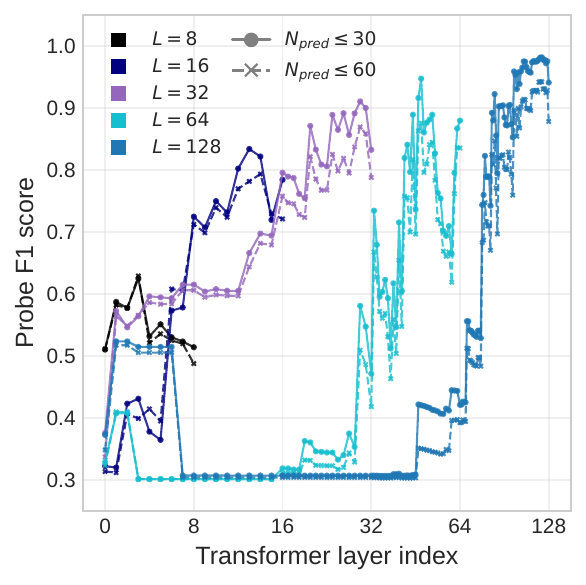}
        \captionof{figure}{Probing provability on LP reveals traces of forward-chaining.}
        \label{fig:provability_probing_lp}
    \end{minipage}

    \vspace{-20pt}

    \section{Discussion}
    \label{sec:discussion}
    These findings align with recent works showing that restructuring representations into a one-dimensionally linear geometry supports generalization to out-of-distribution compositional reasoning tasks~\citep{vompa2026beyond}.
    While linearly represented features are not always linearly accessible~\citep{pacela2026stop, garg2026featureslanguagemodelstore},
    Transformer layers can overcome the interference penalty of linear retrieval by leveraging the computational depth of the model ($L \gg \delta$)
    through several concurrent mechanisms:
    (1) the implementation of LISTA-style algorithms for approximating non-linear decoding~\citep{liu2025on};
    (2) the aggregation of sequence-wise distributed representations to cancel out interference noise~\citep{vompa2026beyond}; and
    (3) the utilization of attention's softmax mechanism to retrieve well-separated patterns in a single update step~\citep{ramsauer2021hopfield},
    thereby recomposing superposed features into a more linearly accessible format.
    Taken together, these mechanisms of geometric unwinding draw a parallel to the representation geometry of World Models,
    which force non-linear observations into a linear transform of the world's latent structure~\citep{klindt2026does}.
    With this redundancy, the direct method resembles CoT by distributing a single deduction across layers.

    More broadly, improving faithfulness and robustness of deductive reasoning could reduce shortcut‑driven failures and overconfident errors.
    But on the other hand, stronger implicit reasoning capabilities may also enable more capable automated systems whose internal decision processes become even more difficult to audit.
    While our experiments are restricted to synthetic tasks,
    care is needed when extrapolating these techniques to real‑world domains.

    \subsection{Limitations}
    \label{subsec:limitations}

    Our findings are established using a toy model on synthetic data modeling short contexts in a controlled Horn clause deduction setting.
    We make use of type embeddings to ground the representations in superposition hypothesis,
    and to separate the concerns of learning representations and learning circuits on these representations.
    The full benefit of the corrective objective is contingent on sufficient architectural depth ($L \ge \delta$).
    The \texttt{r2} heuristic is generally difficult to implement,
    does not fully decorrelate higher-order structural features from provability,
    and makes convergence more difficult.
    Furthermore, bidirectional masking remains difficult to adapt to multi-turn conversational contexts.
    Finally, while scaling depth with these components closes the implicit--explicit gap within the training horizon,
    CoT still exhibits superior depth extrapolation.

    \subsection{Future work}
    \label{subsec:future_work}

    There are several promising directions for future work:
    \begin{itemize}[noitemsep, topsep=2pt, parsep=2pt, partopsep=0px]
            \item[-] Formalizing the theoretical framework for implicit deduction,
            \item[-] and extending it to bound the limits of CoT prompting.
            \item[-] A finer characterization of reasoning primitives,
            \item[-] and developing optimization biases which better promote the learning of those primitives.
            \item[-] The expansion of these principles to richer logics, theorem-proving, or natural language.
            \item[-] The application of these tools to architectures with unbounded state or recurrence.
    \end{itemize}

    While we make an initial attempt at formalizing this framework and investigate potentially learned strategies in \refAppendix{sec:appendix_high_level_analogoies},
    our conclusions currently remain interpretations based on empirical evidence.
    Causal tracing of these complex representations with more powerful,
    non-linear probing methods runs into the non-linear representation dilemma~\citep{sutter2025nonlinear},
    where any neural network can be perfectly mapped to any algorithm,
    even when the model is entirely incapable of solving the actual task.
    Findings in \refAppendix{subsec:probe_visualized} further complicate the causal probing as model does not necessarily only learn a single algorithm.

    \subsection{Conclusion}\label{subsec:conclusion}
    In this work, we investigate the scaling properties of implicit deductive reasoning over Horn clauses in depth-bounded Transformers.
    To study these properties, we systematically mitigate shortcut-inducing biases through the \texttt{r2} heuristic,
    bidirectional prefix masking, and a corrective objective.
    We find increasing model depth allows implicit reasoning to approach the performance of explicit CoT across graph topologies and problem widths.

    \subsection*{Acknowledgment}
    This research was supported by the European Union and the Estonian Research Council through project TEM-TA141,
    and the Estonian Centre of Excellence in Artificial Intelligence (EXAI) project TK213U8,
    funded by the Estonian Ministry of Education and Research.

    \newpage
    \bibliographystyle{tmlr}
    \bibliography{references}

    \newpage
    \appendix

    \setcounter{table}{0}
    \setcounter{figure}{0}
    \renewcommand{\thetable}{A.\arabic{table}}
    \renewcommand{\thefigure}{A.\arabic{figure}}


    \section{High-level analogies capable of explaining empirical trends}
    \label{sec:appendix_high_level_analogoies}

    Deductive reasoning is traditionally categorized by the direction of inference;
    propagating provability from facts toward the query, or splitting queries into subqueries,
    the latter of which can be effectively parallelized~\citep{wolfson1988distributed}.
    More specifically, when modeling these processes as neural operations,
    we define a \textbf{reasoning step} as a composition of underlying reasoning primitives:
    \begin{itemize}[noitemsep, topsep=2pt, parsep=2pt, partopsep=0pt]
        \item[-] Rule synthesis, rule retrieval followed by \textit{synthesis} (to derive new rules).
        \item[-] Forward-chaining, fact retrieval followed by \textit{derivation} (to derive new facts).
    \end{itemize}

    Practically speaking, the pointwise space in FFNs is typically $1 \times Cd_{\text{model}}$.
    If that space contains only a conclusion token with its premises fulfilled,
    all that is needed is to store provability at that position so subsequent layers see that token as already proven,
    a derivation step FFNs can theoretically perform~\citep{meng2023locating, feng2023towards}.
    Here, attention can check across the sequence beforehand to determine whether premises are fulfilled elsewhere.
    Because the visible area in attention grows for every token,
    it can theoretically retrieve information dynamically as sequence length increases.
    However, the pointwise FFN lacks such dynamic scaling with respect to the input.
    To check for the fulfillment of premises on its own,
    it would have to store the entire state in that small space.
    This suggests that the complete reasoning step need not be localized,
    but can potentially be distributed across modules.

    The model has $L$ layers indexed by $\ell$.
    We define \textbf{depthwise complexity}, $\lambda(\delta)$,
    as the minimum layer count required to solve a problem of logical depth $\delta$.
    To contextualize this, we first need to define time in this scope: time is counted in layers.
    When we assume one layer can execute one cross-sequence scan and apply non-linear operations in parallel to make use of that seen information,
    we can start determining the minimal amount of layers required given the amount of times each token sees the full state of the problem.
    This metric reflects the architectural reality that Transformers process the sequence in parallel.
    Consequently, increasing the input size scales the computational width but does not add depth-wise sequential processing capacity.

    \subsection{Depth limits of forward-chaining}\label{subsec:forward_chaining}
    Deduction over Horn clauses is solvable in linear time~\citep{1984linear}.
    While considered sequential in the worst-case scenario,
    its P-completeness implies it is highly difficult (though not impossible) to parallelize efficiently~\citep{1974Complete, greenlaw1995limits}.
    Consequently, assuming attention is limited to a single sequential retrieval step per layer,
    depth-bounded models cannot consistently solve instances when $\delta > L$.
    To faithfully verify a chain of depth $\delta$,
    any approximation therefore hypothetically scales as $\lambda(\delta) = \Omega(\delta)$ layers to propagate the derivation from facts to the query.
    For instance, given fact $A$ and rules $A \rightarrow B$ and $A \land B \rightarrow C$,
    at the first layer, only $A \rightarrow B$ can apply its derivation because only that rule has its premises fulfilled after the initial cross-sequence scan.
    Thus, propagating provability step-by-step requires the number of layers to grow linearly with the depth of the reasoning chain.

    \subsection{Depth limits of rule synthesis}
    \label{subsec:rule_synthesis}

    \begin{minipage}[b]{0.58\textwidth}
        Rule synthesis can be interpreted through the lens of parallel computing;
        reachability in path-like single-premise chains ($A \to B \to C \to \cdots$)
        can be computed in $\Theta(\log \delta)$ time in parallel models via pointer-jumping/doubling~\citep{an1982connectivity}.
        More generally, directed graph reachability (which captures single-premise deduction) admits polylog-depth parallel algorithms,
        with similar algorithms extending to multiple-premise deduction graphs in restricted cases~\citep{bodlaender1998parallel}.
        This suggests that any faithful parallel approximation would hypothetically require at least $\lambda(\delta)=\Omega(\log \delta)$ layers~\citep{merrill2024a}.
        Rule synthesis is particularly plausible in the LP dataset due to its backbone,
        where dependencies flow layer to layer in the same direction (Figure~\ref{fig:synthesis_visualization}).
    \end{minipage}
    \hfill
    \begin{minipage}[b]{0.40\textwidth}
        \vspace{-20pt}
        \centering
        \includegraphics[width=\linewidth]{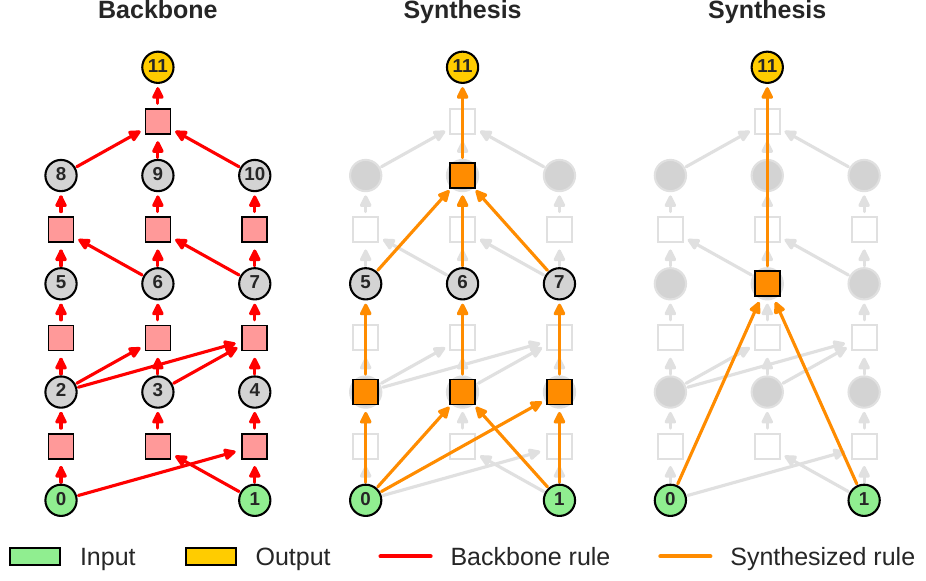}
        \captionof{figure}{Rule synthesis on LP backbone.}
        \label{fig:synthesis_visualization}
    \end{minipage}

    A simple example: given the a fact $A$, rules $A \rightarrow B$, $B \rightarrow C$, $C \rightarrow D$, and $D \rightarrow E$, and a query $E$.
    A single cross-sequence scan and consequent non-linear operations, best case scenario,
    yield the synthesized rules $A \rightarrow C$, $B \rightarrow D$, and $C \rightarrow E$.
    A second scan combines the first and last synthesized rule to produce $A \rightarrow E$.
    Applying fact $A$ to this synthesized rule proves the query.

    \subsection{Intractability of rule synthesis}
    \label{subsec:intractable_search}

    Rule synthesis introduces redundancy, however,
    such as generating the unnecessary rule $B \rightarrow E$ during the second scan of the previous example.
    When extending this heuristic to rules with multiple premises (e.g., $C \land D \rightarrow E$),
    the number of such redundant synthesized rules can grow exponentially.

    The process of combining premises to derive new rules closely mirrors classical query evaluation.
    In classical AI, such an efficient evaluation strategy relies on planning,
    which can manifest as an optimal variable elimination ordering ($d$) or a hypertree decomposition of a query ($Q$)~\citep{rina1994directional, gottlob1999hypertree}.
    From this perspective,
    the algorithmic complexity is bounded by the induced width ($w^*(d)$),
    which corresponds to the maximum number of variables in any synthesized rule,
    and hypertree width ($hw(Q)$),
    which corresponds to the maximum number of relations joined in intermediate computations, respectively.
    Intuitively, these widths reflect how many variables or relations need to be tracked simultaneously.
    However, as such planning is NP-hard~\citep{dechter1999bucket, gottlob1999hypertree},
    meaning, as these widths grow, the strategy becomes intractable in practice.

    Viewing rule synthesis as a circuit operation assumes rules partition cleanly.
    This framework applies to rule synthesis by generalizing operators over logical circuits,
    which encode Boolean functions as computational graphs with AND and OR gates~\citep{adnan2002knowledge}.
    This requires decomposability, meaning, product unit inputs depend on disjoint sets of variables.
    These units encode local factorizations, which, together with smoothness, enable tractable integration over sets of variables.
    Even when both input circuits are structured-decomposable and deterministic,
    computing their product as a decomposable circuit is \#P-hard
    when their variable decompositions are not guaranteed to be compatible~\citep{vergari2021a},
    and therefore at least NP-hard under standard reductions~\citep{complexity2009arora}.

    \subsection{Sparse-feature retrieval as a bottleneck}
    \label{subsec:memory_bandwidth}

    Selecting $s$ active features (premises) from a manifold of $n$ possible features requires encoding the $\binom{n}{s}$ combinatorial space.
    While non-linear decoding algorithms establish an information-theoretic lower bound of
    $d = \Omega(s \log(n/s))$ for recovery~\citep{elhage2022superposition, ba2011lower, wainwright2007information},
    giving an upper bound for storage; not all linearly stored (represented) features are linearly accessible, however~\citep{pacela2026stop}.
    Linear accessibility demands low pairwise interference among the $k$ active premises (analogous to finding cliques in interference graphs),
    resulting in a worst-case linear decoding lower bound of $d = \Omega_{\epsilon}(\frac{s^{2}}{\log s}\log(n/s))$
    with a nearly-matching upper bound confirming exponential feature superposition~\citep{garg2026featureslanguagemodelstore}.

    While distributing retrieval across layers mitigates this quadratic penalty,
    the following applies when doing retrieval within a single attention head.
    If a model faithfully approximates forward-chaining,
    the number of simultaneously active features theoretically scales with the number of premises per rule.
    For rule synthesis, however, the relevant bandwidth depends on the representational basis the model learns to plan with.
    If its latent features track individual variables, the feature space should scale with the number of latent predicates,
    and the simultaneously active features scale with the induced width ($w^*(d)$);
    if it tracks latent rules, the bottleneck shifts to the hypertree width ($hw(Q)$).

    \subsection{Empirical validation of theoretical scaling properties}
    \label{subsec:theoretic_scaling_empirical}

\begin{figure}[htbp]
        \centering
        \includegraphics[width=\linewidth]{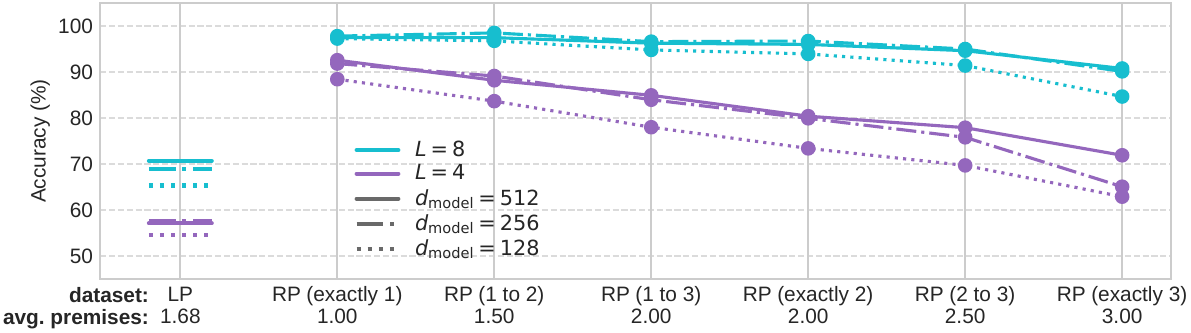}
        \caption{Evaluation across topologies on logical depth $\delta = 5$ problems.
        The number of attention heads is fixed to $H=4$.
        We vary model depth ($L$), dimension ($d_{\text{model}}$), and rule premise counts to assess the robustness of learned approximations.
        Statistically significant improvements from scaling dimensionality ($p < 0.05$) were observed only for $L=4$ models (\refAppendix{subsec:appendix_theory_validation}).}
        \label{fig:theory_visualized}
    \end{figure}

    To evaluate the properties of forward-chaining,
    we analyze models with sufficient depth to execute it,
    satisfying the condition $L \ge \lambda(\delta) = \Omega(\delta)$.
    For problems with logical depth $\delta = 5$, the $L = 8$ models achieved high, stable accuracy across rule premise counts (Figure~\ref{fig:theory_visualized}).
    This is consistent with a sequential strategy that, unlike unbounded search,
    only requires recovering a number of simultaneously active features linearly scaling off the rule's premises.
    Consequently, once these features are recoverable, further scaling the attention head dimensionality yields no meaningful performance improvements.

    \vspace{-1em}
    When models must rely on rule synthesis, requiring at least $\lambda(\delta) = \Omega(\log \delta)$ layers,
    increasing the number of premises in the original rules intuitively yields synthesized rules with even more premises,
    driving an intractable expansion of the search space.
    While scaling the attention head dimensionality to accommodate more simultaneously active features did yield significant improvements,
    particularly when rules had many premises, this only delays a fundamental bottleneck.
    As unbounded search necessitates worst-case exponential growth in complexity in random graphs as the number of premises increases (regardless of planning effectiveness),
    maintaining log-time execution requires \textbf{the number} of attention heads \textbf{and their dimensionality} to also grow exponentially to approximate a faithful algorithm.

    \textbf{Limitations.}
    However, there remains a gap between these experimental results and scaling trends observed in the main paper.
    Forward-chaining can theoretically do deduction in linear number of layers,
    but why was essentially an exponential number of layers required empirically (Section~\ref{subsec:scaling_trends})?
    The rule synthesis approach could explain the exponential amount of compute,
    but why must the compute need be depth-wise (Section~\ref{sec:appendix_scaling_trends}),
    and why do the intermediate traces become provable (Section~\ref{subsec:bfs_like_reasoning})?
    When compressing $A \rightarrow B$ and $B \rightarrow C$ to $A \rightarrow C$, B does not become provable.
    Findings in \refAppendix{subsec:probe_visualized} give some insight,
    that model must not necessarily learn to approximate only a single classical algorithm,
    so truth may lie somewhere inbetween forward-chaining and rule synthesis.

    \newpage

    \subsection{Procrustes probe visualized}\label{subsec:probe_visualized}

    Procrustes aligned linear probing method (\refAppendix{sec:appendix_decision_trace}) allows us to visualize the convergence dynamics,
    when we aggregate these scores calculated for the query token across the evaluation set.
    The bold lines represent the median confidence probability at each layer, with shaded regions indicating the interquartile range (25th--75th percentiles).
    Alongside the probabilities, we also plot the underlying logit magnitudes for both the correct and incorrect tokens to illustrate how the model's internal evidence accumulation drives the final softmax distribution.
    We evaluate our 8-layer RP trained model with \texttt{bidir} + \texttt{corrective} + \texttt{r2} components across logical depths $\delta \in \{0, \dots, 6\}$.
    Evaluation on datasets with 1-premise and 3-premise rules only allows us to directly contrast learned approximations.

    \begin{figure}[h!]
        \centering
        \begin{subfigure}{0.24\linewidth}
            \includegraphics[width=\linewidth]{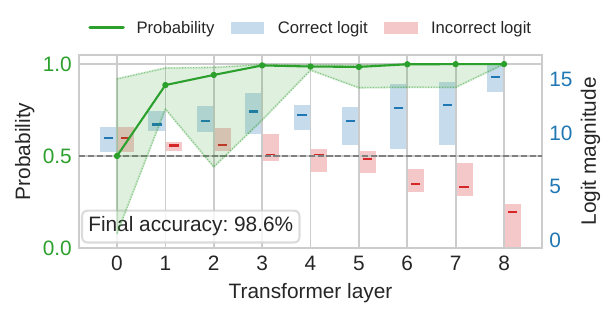}
            \caption{$\delta = 0$}
        \end{subfigure}\hfill
        \begin{subfigure}{0.24\linewidth}
            \includegraphics[width=\linewidth]{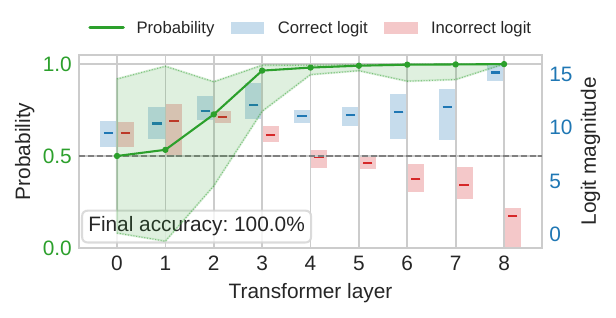}
            \caption{$\delta = 1$}
        \end{subfigure}\hfill
        \begin{subfigure}{0.24\linewidth}
            \includegraphics[width=\linewidth]{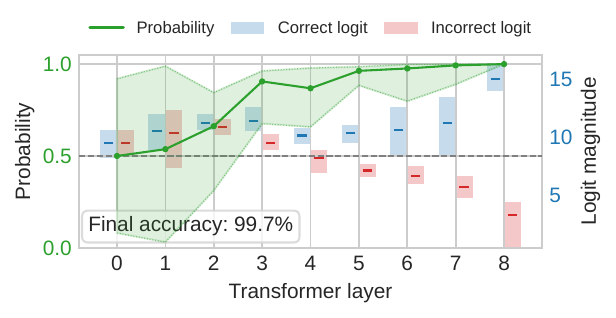}
            \caption{$\delta = 2$}
        \end{subfigure}\hfill
        \begin{subfigure}{0.24\linewidth}
            \includegraphics[width=\linewidth]{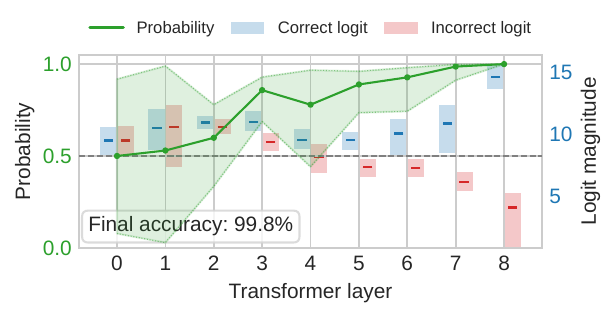}
            \caption{$\delta = 3$}
        \end{subfigure}

        \vspace{0em}

        \begin{subfigure}{0.32\linewidth}
            \includegraphics[width=\linewidth]{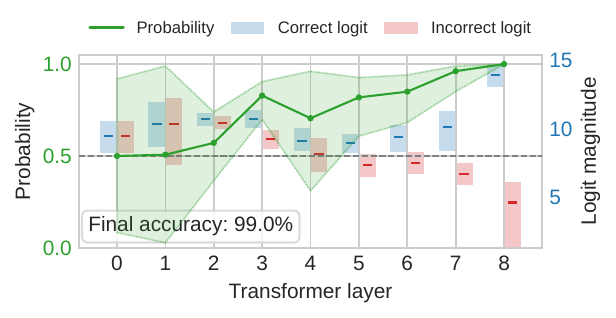}
            \caption{$\delta = 4$}
        \end{subfigure}\hfill
        \begin{subfigure}{0.32\linewidth}
            \includegraphics[width=\linewidth]{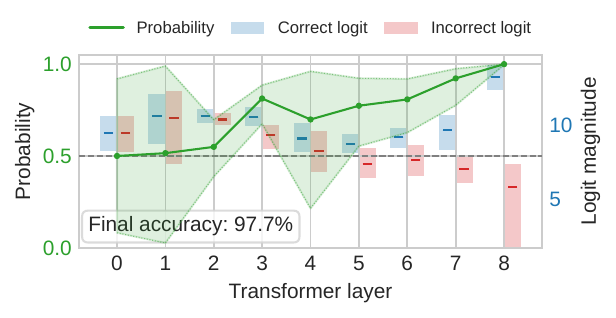}
            \caption{$\delta = 5$}
        \end{subfigure}\hfill
        \begin{subfigure}{0.32\linewidth}
            \includegraphics[width=\linewidth]{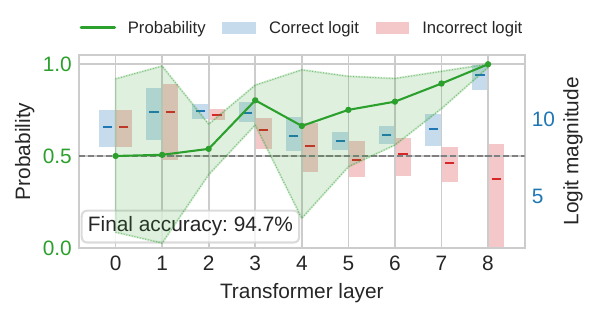}
            \caption{$\delta = 6$}
        \end{subfigure}
        \caption{Under 1-premise rules the lower bound of the IQR converges at $\ell \ge \lceil \log_2 \delta \rceil$, supporting rule synthesis.}
        \label{fig:trace_1_premise_depths}
    \end{figure}

    \begin{figure}[h!]
        \centering
        \begin{subfigure}{0.24\linewidth}
            \includegraphics[width=\linewidth]{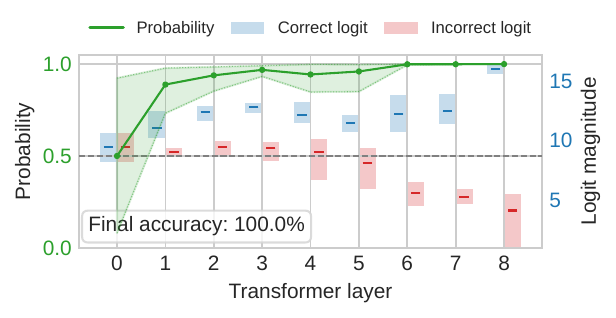}
            \caption{$\delta = 0$}
        \end{subfigure}\hfill
        \begin{subfigure}{0.24\linewidth}
            \includegraphics[width=\linewidth]{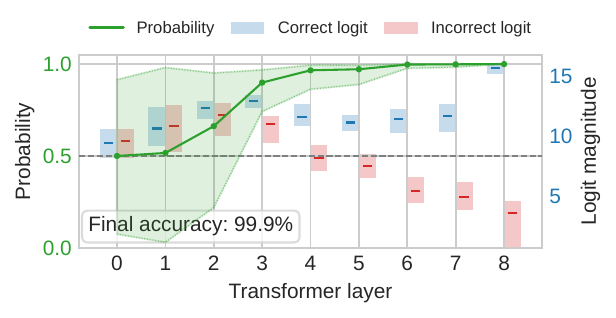}
            \caption{$\delta = 1$}
        \end{subfigure}\hfill
        \begin{subfigure}{0.24\linewidth}
            \includegraphics[width=\linewidth]{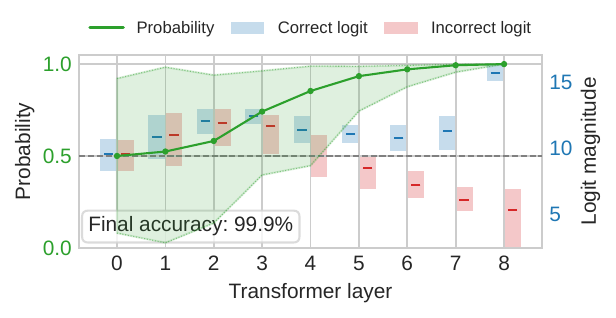}
            \caption{$\delta = 2$}
        \end{subfigure}\hfill
        \begin{subfigure}{0.24\linewidth}
            \includegraphics[width=\linewidth]{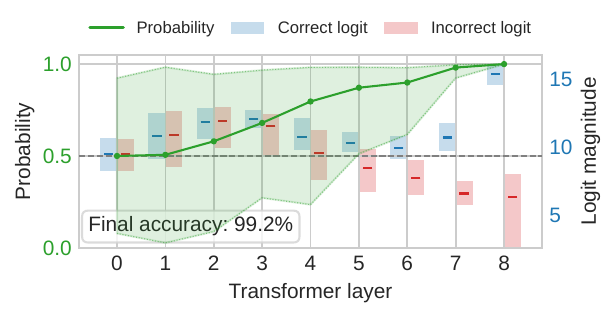}
            \caption{$\delta = 3$}
        \end{subfigure}

        \vspace{0em}

        \begin{subfigure}{0.32\linewidth}
            \includegraphics[width=\linewidth]{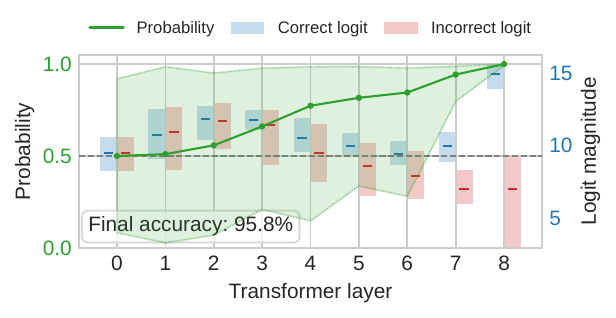}
            \caption{$\delta = 4$}
        \end{subfigure}\hfill
        \begin{subfigure}{0.32\linewidth}
            \includegraphics[width=\linewidth]{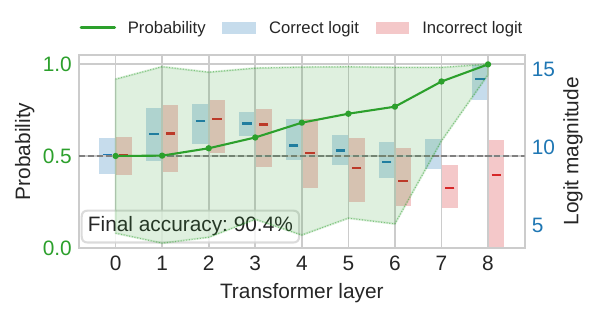}
            \caption{$\delta = 5$}
        \end{subfigure}\hfill
        \begin{subfigure}{0.32\linewidth}
            \includegraphics[width=\linewidth]{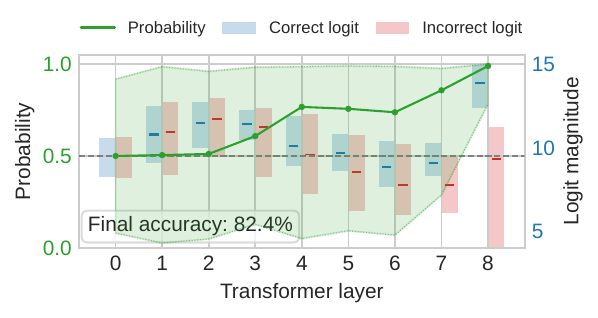}
            \caption{$\delta = 6$}
        \end{subfigure}
        \caption{Under 3-premise rules the convergence is delayed until the layer depth matches the logical depth ($\ell \ge \delta$), supporting forward-chaining.}
        \label{fig:trace_3_premise_depths}
    \end{figure}

    \newpage

    \subsection{Extended empirical validation of theoretical scaling properties}
    \label{subsec:appendix_theory_validation}

    Training was conducted on RP problems containing a maximum $N_{\text{pred}} \le 30$ with a logical depth of $\delta \le 6$,
    with rules containing 1 to 3 premises.
    Table~\ref{tab:corrective_shallow} details performance across a grid of model depths ($L \in \{2, 4, 8\}$) and model dimensions ($d_{\text{model}} \in \{128, 256, 512\}$).
    The number of attention heads is fixed to $H = 4$.

    \begin{table}[ht!]
        \centering
        \caption{$L$-layer $d$-dimensional model direct performance with \texttt{r2},
            \texttt{bidirectional} and \texttt{corrective} components on logical depth $\delta = 5$ and $6$ problems.
            Number of attention heads is fixed to $H=4$.
            we evaluate performance across varying premise counts, which alters the feasibility of rule synthesis.
            \textcolor{green!70!black}{Green} indicates a statistically significant improvement ($p < 0.05$).
            A one-tailed paired sample t-test was used to determine if wider models outperform the $d_{\text{model}}=128$ baseline at a given layer depth,
            where the t-threshold was calculated using $k-1$ degrees of freedom to account for the small sample size of 3 seeds.}
        \label{tab:corrective_shallow}
        \resizebox{\textwidth}{!}{
            \begin{tabular}{lcccccccc}
                \toprule
                Eval $\delta$
                & Model parameters              & LP  (1 to 3)                                  & RP (exactly 1)                              & RP (1 to 2)                                 & RP  (1 to 3)                                  & RP (exactly 2)                              & RP (2 to 3)                                 & RP (exactly 3)                              \\
                \midrule
                \multirow{9}{*}{$\delta = 5$}
                & $L=2$, $d_{\text{model}}=128$ & $0.55 \pm 0.01$                             & $0.83 \pm 0.03$                             & $0.79 \pm 0.01$                             & $0.67 \pm 0.03$  & $0.67 \pm 0.06$  & $0.63 \pm 0.04$  & $0.59 \pm 0.03$  \\
                & $L=2$, $d_{\text{model}}=256$ & $0.54 \pm 0.01$                             & $0.84 \pm 0.01$                             & $0.78 \pm 0.01$                             & $0.67 \pm 0.02$  & $0.70 \pm 0.03$  & $0.65 \pm 0.03$  & $0.60 \pm 0.02$  \\
                & $L=2$, $d_{\text{model}}=512$ & $0.53 \pm 0.02$                             & $0.73 \pm 0.18$                             & $0.71 \pm 0.14$                             & $0.62 \pm 0.08$  & $0.65 \pm 0.07$  & $0.60 \pm 0.02$  & $0.54 \pm 0.04$  \\
                \cmidrule(lr){2-9}
                & $L=4$, $d_{\text{model}}=128$ & $0.55 \pm 0.02$                             & $0.88 \pm 0.01$                             & $0.84 \pm 0.01$                             & $0.78 \pm 0.01$  & $0.73 \pm 0.02$  & $0.70 \pm 0.02$  & $0.63 \pm 0.03$  \\
                & $L=4$, $d_{\text{model}}=256$ & $0.58 \pm 0.03$                             & \textcolor{green!70!black}{$0.92 \pm 0.01$} & \textcolor{green!70!black}{$0.89 \pm 0.01$} & \textcolor{green!70!black}{$0.84 \pm 0.01$} & \textcolor{green!70!black}{$0.80 \pm 0.02$} & \textcolor{green!70!black}{$0.76 \pm 0.03$} & $0.65 \pm 0.06$  \\
                & $L=4$, $d_{\text{model}}=512$ & \textcolor{green!70!black}{$0.57 \pm 0.03$} & $0.93 \pm 0.03$  & \textcolor{green!70!black}{$0.88 \pm 0.02$} & \textcolor{green!70!black}{$0.85 \pm 0.02$} & \textcolor{green!70!black}{$0.80 \pm 0.02$} & \textcolor{green!70!black}{$0.78 \pm 0.04$} & \textcolor{green!70!black}{$0.72 \pm 0.03$} \\
                \cmidrule(lr){2-9}
                & $L=8$, $d_{\text{model}}=128$ & $0.65 \pm 0.07$                             & $0.97 \pm 0.01$                             & $0.97 \pm 0.01$                             & $0.95 \pm 0.03$  & $0.94 \pm 0.03$  & $0.91 \pm 0.04$  & $0.85 \pm 0.07$  \\
                & $L=8$, $d_{\text{model}}=256$ & $0.69 \pm 0.02$                             & $0.98 \pm 0.01$                             & \textcolor{green!70!black}{$0.98 \pm 0.01$} & $0.97 \pm 0.01$  & $0.97 \pm 0.00$  & $0.95 \pm 0.01$  & $0.90 \pm 0.01$  \\
                & $L=8$, $d_{\text{model}}=512$ & $0.71 \pm 0.07$                             & $0.97 \pm 0.03$                             & $0.97 \pm 0.03$                             & $0.96 \pm 0.03$  & $0.96 \pm 0.03$  & $0.95 \pm 0.03$  & $0.91 \pm 0.04$  \\
                \midrule
                \multirow{9}{*}{$\delta = 6$}
                & $L=2$, $d_{\text{model}}=128$ & $0.54 \pm 0.01$                             & $0.85 \pm 0.02$                             & $0.82 \pm 0.01$                             & $0.67 \pm 0.03$  & $0.67 \pm 0.02$  & $0.60 \pm 0.03$  & $0.58 \pm 0.04$  \\
                & $L=2$, $d_{\text{model}}=256$ & $0.53 \pm 0.01$                             & $0.85 \pm 0.01$                             & $0.81 \pm 0.01$                             & $0.66 \pm 0.01$  & $0.68 \pm 0.03$  & $0.62 \pm 0.01$  & $0.59 \pm 0.02$  \\
                & $L=2$, $d_{\text{model}}=512$ & $0.52 \pm 0.02$                             & $0.73 \pm 0.20$                             & $0.72 \pm 0.17$                             & $0.62 \pm 0.07$  & $0.63 \pm 0.09$  & $0.56 \pm 0.03$  & $0.53 \pm 0.02$  \\
                \cmidrule(lr){2-9}
                & $L=4$, $d_{\text{model}}=128$ & $0.56 \pm 0.00$                             & $0.88 \pm 0.01$                             & $0.82 \pm 0.03$                             & $0.75 \pm 0.01$  & $0.71 \pm 0.03$  & $0.63 \pm 0.02$  & $0.60 \pm 0.03$  \\
                & $L=4$, $d_{\text{model}}=256$ & $0.57 \pm 0.02$                             & \textcolor{green!70!black}{$0.92 \pm 0.01$} & \textcolor{green!70!black}{$0.87 \pm 0.02$} & \textcolor{green!70!black}{$0.81 \pm 0.01$} & \textcolor{green!70!black}{$0.77 \pm 0.02$} & \textcolor{green!70!black}{$0.69 \pm 0.03$} & $0.60 \pm 0.07$  \\
                & $L=4$, $d_{\text{model}}=512$ & $0.57 \pm 0.01$                             & $0.92 \pm 0.04$                             & $0.87 \pm 0.03$                             & $0.79 \pm 0.03$  & $0.76 \pm 0.03$  & \textcolor{green!70!black}{$0.72 \pm 0.02$} & \textcolor{green!70!black}{$0.67 \pm 0.01$} \\
                \cmidrule(lr){2-9}
                & $L=8$, $d_{\text{model}}=128$ & $0.59 \pm 0.05$                             & $0.95 \pm 0.02$                             & $0.95 \pm 0.02$                             & $0.92 \pm 0.03$  & $0.89 \pm 0.04$  & $0.85 \pm 0.05$  & $0.78 \pm 0.10$  \\
                & $L=8$, $d_{\text{model}}=256$ & $0.62 \pm 0.01$                             & $0.95 \pm 0.02$                             & $0.96 \pm 0.01$                             & $0.95 \pm 0.01$  & $0.93 \pm 0.01$  & $0.90 \pm 0.01$  & $0.81 \pm 0.02$  \\
                & $L=8$, $d_{\text{model}}=512$ & $0.63 \pm 0.03$                             & $0.96 \pm 0.05$                             & $0.96 \pm 0.03$                             & $0.94 \pm 0.04$  & $0.92 \pm 0.04$  & $0.89 \pm 0.04$  & $0.83 \pm 0.07$  \\
                \midrule
                & Avg. premises per rule        & 1.68                                        & 1.00                                        & 1.50                                        & 2.00                                        & 2.00                                        & 2.50                                        & 3.00                                        \\
                \bottomrule
            \end{tabular}
        }
    \end{table}

    On problems with depths $\delta = 5$ and $\delta = 6$,
    only models with $L = 4$ layers saw significant performance improvements when scaling attention head dimensionality,
    aligning with properties stemming from unbounded search.

    \newpage

    \section{Dataset generation methodologies}
    \label{sec:appendix_dataset_generation}

    We use three sampling algorithms proposed by~\citep{2023paradox},
    which produce datasets with different underlying structures, designed to test model's robustness under distribution shift.
    For every example, we first sample the number of predicates $N_{\text{pred}} \sim \mathrm{UnifInt}(N_{\min}, N_{\max})$,
    and then sample predicates themselves from vocabulary of 150.
    Across all generators, rules are constrained to have 1 to 3 premises.
    To prevent positional bias, after generation, we randomly shuffle the order of premises within each rule, the list of rules, and the list of facts.

    \textbf{Rule-Priority (RP)} generates random graphs.
    Given the predicate set, we sample the number of rules $N_{\text{rules}} \sim \mathrm{UnifInt}(0, 4N_{\text{pred}})$
    and the number of facts $N_{\text{facts}} \sim \mathrm{UnifInt}(0, N_{\text{pred}})$.
    Each rule is formed by sampling $k{+}1$ distinct predicates (where $k \in \{1,2,3\}$),
    using the first $k$ as premises and the last as the conclusion.
    We enforce uniqueness by rejecting duplicate rules (treating the premise set as unordered).
    Finally, facts and the query are sampled uniformly from the predicate set.

    \textbf{Label-Priority (LP)} generates hierarchical graphs with noise.
    We first sample an underlying backbone depth $D \sim \mathrm{UnifInt}(1, \lfloor N_{\text{pred}}/2 \rfloor)$ and partition predicates into $D$ ordered levels.
    We allocate $\lfloor N_{\text{pred}}/D \rfloor$ predicates per level and distribute the remaining $N_{\text{pred}} \bmod D$ predicates by adding at most one extra predicate to levels $l=1,2,\dots$ in order.
    Each predicate is assigned an initial random binary label.
    To ensure non-degeneracy, we enforce at least one False and one True predicate per level
    (implemented by overwriting the first two labels in each level).
    The generation proceeds in two phases:
    \begin{itemize}[noitemsep, topsep=2pt, parsep=2pt, partopsep=0px]
        \item[-] Backbone construction.
        For each predicate in level $l{+}1$, we sample a rule whose premises are drawn exclusively from level $l$,
        matching the premise labels to the conclusion label to ensure consistency.
        \item[-] Noise injection.
        We inject distractor rules with a target count $N_{\text{noise}} \sim \mathrm{UnifInt}(0, 3N_{\text{pred}})$.
        Heads and tails are sampled from all levels, rejecting any rule that would make a false-labeled predicate derivable from all-true premises.
        Whenever an all-true noise rule is sampled, we additionally inject an all-false rule.
    \end{itemize}
    The fact set is defined as the true-labeled predicates in the first level, and the query is sampled uniformly.

    \textbf{LP*} is similar to LP but with slight modifications to increase cyclic dependencies.
    In backbone generation, premises for false-labeled conclusions are drawn from \textbf{all} predicates in the previous level, with all-true premises being rejected.
    In noise rule generation (when $D>1$), each rule concludes in a true-labeled predicate chosen from an intermediate level $l \in \{0,\dots,D-2\}$,
    and its 1--3 premises are sampled from other predicates in levels $\{l,\dots,D-1\}$.
    As in LP, facts are the true-labeled predicates in the first level and the query is sampled uniformly.

    To give a sense of dataset complexity, Table~\ref{tab:rule_count_stats} reports the expected and maximum number of rules per example.
    Note that because we balance the training set by logical depth via rejection sampling,
    our actual instances lean toward the upper rule bound rather than the sampler mean,
    with all sequences capped at 1024 tokens.

    \begin{table}[h!]
        \centering
        \begin{adjustbox}{max width=\linewidth}
            \begin{tabular}{lcccc}
                \toprule
                & \multicolumn{2}{c}{$N_{\text{pred}}\in[5,30]$} & \multicolumn{2}{c}{$N_{\text{pred}}\in[5,60]$} \\
                Generator & Expected mean $N_{\text{rules}}$ & Max $N_{\text{rules}}$ & Expected mean $N_{\text{rules}}$ & Max $N_{\text{rules}}$ \\
                \midrule
                RP        & $35$                             & $120$                  & $65$                             & $240$                  \\
                LP        & $\approx 38.4$                   & $118$                  & $\approx 74.8$                   & $238$                  \\
                \bottomrule
            \end{tabular}
        \end{adjustbox}
        \caption{Expected and maximum number of rules per example under each generator.}
        \label{tab:rule_count_stats}
    \end{table}

    \textbf{How the means are computed.}
    RP sampler draws $N_{\text{rules}}\mid N_{\text{pred}} \sim \mathrm{UnifInt}(0,4N_{\text{pred}})$, so
    $\mathbb{E}[N_{\text{rules}}\mid N_{\text{pred}}]=2N_{\text{pred}}$ and therefore $\mathbb{E}[N_{\text{rules}}]=2\mathbb{E}[N_{\text{pred}}]=N_{\min}+N_{\max}$,
    with $\max N_{\text{rules}}=4N_{\max}$.

    For LP, the total rule count is the sum of backbone and noise rules.
    The backbone contributes exactly one rule for each predicate above level~0, i.e.,
    $N_{\text{bb}}(N_{\text{pred}},D)=N_{\text{pred}}-\lfloor N_{\text{pred}}/D\rfloor$ where $D\mid N_{\text{pred}}\sim \mathrm{UnifInt}(1,\lfloor N_{\text{pred}}/2\rfloor)$.
    Noise rules are drawn as $N_{\text{noise}}\mid N_{\text{pred}} \sim \mathrm{UnifInt}(0,3N_{\text{pred}})$, giving $\mathbb{E}[N_{\text{noise}}\mid N_{\text{pred}}]=1.5N_{\text{pred}}$.
    Thus $\mathbb{E}[N_{\text{rules}}]=\mathbb{E}_{N_{\text{pred}}}\!\left[1.5N_{\text{pred}}+\mathbb{E}_{D\mid N_{\text{pred}}}\!\left(N_{\text{pred}}-\lfloor N_{\text{pred}}/D\rfloor\right)\right]$.

    \newpage

    \section{Type embeddings for symbolic reasoning}
    \label{sec:appendix_type_embeddings}

    We employ \textbf{type embeddings} to give the model an explicit understanding of the problem's symbolic structure.
    Unlike standard language models that treat input as a flat sequence of token IDs,
    our approach augments the input representation by assigning a set of semantic roles to each token.
    This inductive bias allows the model to distinguish between logical components, such as facts, rule premises, and rule conclusions,
    independent of their specific token identity.

    \subsection{Embedding integration strategy}

    Formally, we define a learnable type embedding matrix $W_{\text{type}} \in \mathbb{R}^{N_{\text{type}} \times d_{\text{model}}}$,
    where $N_{\text{type}}$ is the vocabulary size of semantic types and $d_{\text{model}}$ is the model's hidden dimension.
    Our framework allows for a compositional representations where a single token $x_t$ at position $t$ is associated with a set of active semantic types $\mathcal{T}_t$.
    The final input representation $E_{\text{final}}^{(t)}$ is computed by summing the standard vocabulary embedding $E_{\text{token}}(x_t)$ with the sum of all associated type embeddings:
    \begin{equation}
        E_{\text{final}}^{(t)} = E_{\text{token}}(x_t) + \sum_{i \in \mathcal{T}_t} W_{\text{type}}[i].
    \end{equation}

    This summation operation enables the model to encode hierarchical information
    (e.g., a token is both part of a \textit{rule} and specifically the \textit{premise} of a rule)
    into the dense vector space before the first Transformer layer processes the sequence.

    \subsection{Semantic vocabulary and tokenization}

    We utilize a compact vocabulary of fixed semantic types detailed in Table~\ref{tab:semantic_types}.
    These types describe the functional role of each token within the logical graph.

    \begin{table}[h!]
        \centering
        \caption{The vocabulary of semantic types used to annotate input tokens.}
        \label{tab:semantic_types}
        \begin{adjustbox}{max width=\linewidth}
            \begin{tabular}{@{}lcl@{}}
                \toprule
                Type label            & Index & Logical role                                             \\
                \midrule
                \texttt{fact\_e}      & 1     & Denotes a known \textbf{fact}.                           \\
                \texttt{query\_e}     & 2     & Denotes the target \textbf{query} to be proven.          \\
                \texttt{rule\_e}      & 3     & General type for any token belonging to a \textbf{rule}. \\
                \texttt{rulestart\_e} & 4     & Specific marker for a rule's \textbf{premise}.           \\
                \texttt{ruleend\_e}   & 5     & Specific marker for a rule's \textbf{conclusion}.        \\
                \texttt{task\_e}      & 8     & Special token delimiting the reasoning \textbf{task}.    \\
                \bottomrule
            \end{tabular}
        \end{adjustbox}
    \end{table}

    To illustrate the tokenization and typing process, consider a minimal logical problem consisting of the fact set $\mathcal{F}=\{0, 1\}$,
    a single rule $1 \to 2$, and a query for predicate $3$.
    The serialization pipeline constructs a context sequence by concatenating facts and rules, followed by the task token and query.
    Table~\ref{tab:tokenization_visual} visualizes the resulting input sequence.

    \begin{table}[h!]
        \centering
        \caption{Visualization of the input sequence and associated type sets for the example problem.}
        \label{tab:tokenization_visual}
        \begin{adjustbox}{max width=\linewidth}
            \begin{tabular}{@{}lcccccc@{}}
                \toprule
                Step                           & Fact 0     & Fact 1     & Rule premise   & Rule conclusion   & Task delimiter & Query      \\
                \midrule
                Token ID                       & \texttt{0} & \texttt{1} & \texttt{1}     & \texttt{2}        & \texttt{200}   & \texttt{3} \\
                \cmidrule(lr){1-7}
                Type indices ($\mathcal{T}_t$) & $\{1\}$    & $\{1\}$    & $\{3, 4\}$     & $\{3, 5\}$        & $\{8\}$        & $\{2\}$    \\
                Active embeddings              & fact       & fact       & rule + premise & rule + conclusion & task           & query      \\
                \bottomrule
            \end{tabular}
        \end{adjustbox}
    \end{table}

    \newpage

    \section{Open source baselines}
    \label{sec:appendix_open_source_baselines}

\begin{figure}[h!]
        \centering
        \begin{subfigure}{0.48\textwidth}
            \includegraphics[width=\linewidth]{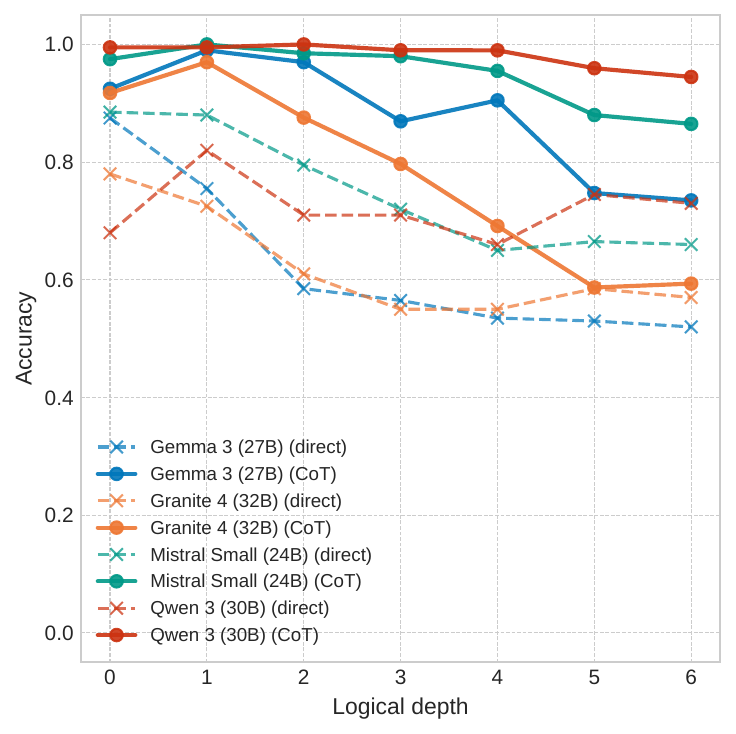}
            \caption{Evaluation on RP reasoning graphs.}
            \label{fig:benchmark_rp}
        \end{subfigure}
        \hfill
        \begin{subfigure}{0.48\textwidth}
            \includegraphics[width=\linewidth]{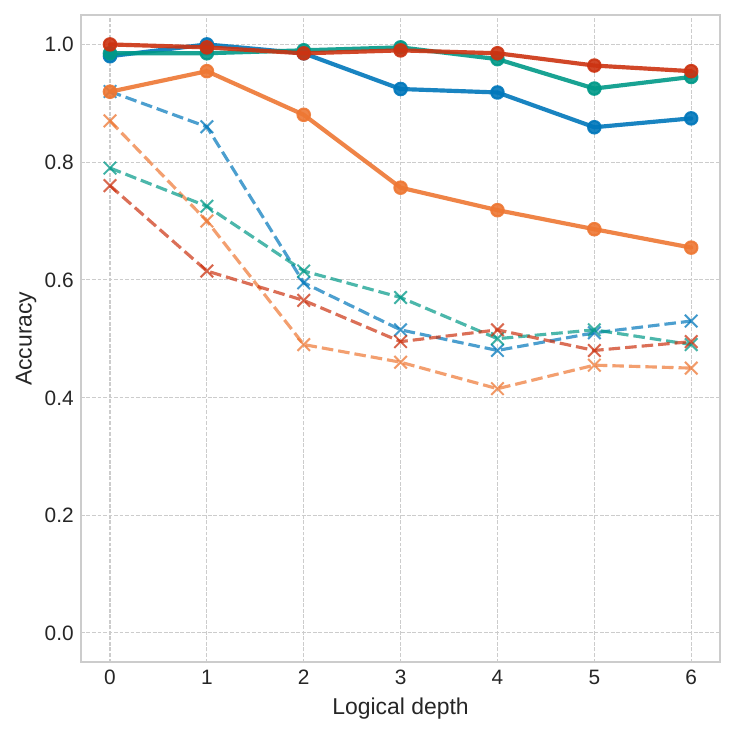}
            \caption{Evaluation on LP reasoning graphs.}
            \label{fig:benchmark_lp}
        \end{subfigure}
        \caption{
            Baseline performance of open-source models ($\approx$30B) under direct (dashed) and CoT (solid) evaluation.
            While CoT reasoning is robust in models like Qwen, direct evaluation degrades rapidly as logical depth increases.
        }
        \label{fig:sota_benchmarks}
    \end{figure}

    We evaluated state-of-the-art open-source models ($\approx$ 30B parameters) on LP and RP distributions to benchmark our findings.
    Experiments were constrained to problems with logical depth $\delta \le 6$ using a one-shot prompt format.
    We enforced a generation budget of 10,000 tokens per sample.
    \footnote{While the Granite model exceeded this limit in $>5\%$ of cases (which were subsequently ignored), its performance remained poor, so this attrition did not skew the comparative rankings.}
    The results, illustrated in Figure~\ref{fig:sota_benchmarks},
    highlight a fundamental gap between implicit and explicit computational capabilities in current architectures.

    \textbf{The limitation of direct evaluation.}
    Without specific fine-tuning, current models severely struggle to simulate the forward-chaining algorithm directly.
    On the LP dataset, accuracy collapses to random chance ($\approx 50\%$) for logical depths $\ge 2$.
    Similarly, on the RP dataset, performance degrades below 70\% beyond depth 2.
    While larger models may offer marginal improvements,
    this trend suggests that standard pre-training objectives do not naturally induce the state transitions required for implicit deduction.

    \textbf{The sufficiency of CoT.}
    In contrast, the capability for deductive logic is clearly present when the computation is unrolled explicitly.
    Notably, Qwen 3 (30B) achieves consistent $>90\%$ accuracy across both distributions using CoT.
    This suggests the barrier is less about logical knowledge and more about the inability to compress that reasoning into a single forward pass.

    \newpage

    \subsection{Prompting methodology}
    \label{subsec:prompt_methodology}

    We employ a prompting strategy constructed by concatenating three segments: a fixed system context,
    the dynamic problem statement, and a mode-specific evaluation trigger.
    Variables enclosed in \texttt{\{braces\}} are substituted dynamically during evaluation.

    \noindent\textbf{1. System context} (fixed)
    \begin{tcolorbox}[colback=gray!5, colframe=gray!40, boxrule=0.8pt, sharp corners, left=4pt, right=4pt, top=4pt, bottom=4pt]
        \small\ttfamily
        You are a logical reasoner. Your task is to determine if the query is provable or unprovable based on the given facts and rules.\\

        Instructions:\\
        Reason by building a forward-chaining proof. Start with your known facts.
        Apply rules to derive new facts. Repeat until you either derive the query or can no longer apply any new rules.
        A rule can ONLY be applied if ALL of its premises are currently present in the known facts.
    \end{tcolorbox}

    \noindent\textbf{2. Problem statement} (dynamic)
    \begin{tcolorbox}[colback=white, colframe=gray!40, boxrule=0.8pt, sharp corners, left=4pt, right=4pt, top=4pt, bottom=4pt]
        \small\ttfamily
        Facts (known to be true): \{comma\_separated\_facts\}\\

        Rules:\\
        R1: IF \{premises\} THEN \{conclusion\}\\
        R2: IF \{premises\} THEN \{conclusion\}\\
        ...\\

        Query:\\
        Is the proposition '\{query\}' true based on the above facts and rules?
    \end{tcolorbox}

    \noindent\textbf{3. Evaluation triggers} (condition-dependent)
    \begin{tcolorbox}[colback=gray!5, colframe=gray!40, boxrule=0.8pt, sharp corners, left=4pt, right=4pt, top=4pt, bottom=4pt]
        \small\ttfamily
        \textbf{[Option A: CoT evaluation]}\\

        Show your steps clearly.\\

        Example step:\\
        - Known facts: 1, 2\\
        - Apply R1 (IF 1 AND 2 THEN 3): Derived new fact 3. Known facts: 1, 2, 3\\

        Reason step-by-step to reach your conclusion.
        If the query is derived, stop and output "Conclusion: provable".
        If no more rules apply and query is not derived, output "Conclusion: unprovable".

        \vspace{1em}
        \hrule
        \vspace{1em}

        \textbf{[Option B: direct evaluation]}\\

        Do not provide any intermediate steps or explanations.
        Based on the facts and rules, output ONLY the final answer in the format "Conclusion: [provable/unprovable]".
    \end{tcolorbox}

    \newpage

    \section{Bias towards shorter paths when sampling randomly}
    \label{sec:appendix_sampling_bias}

\begin{figure}[h!]
        \centering
        \begin{subfigure}{0.48\textwidth}
            \includegraphics[width=\linewidth]{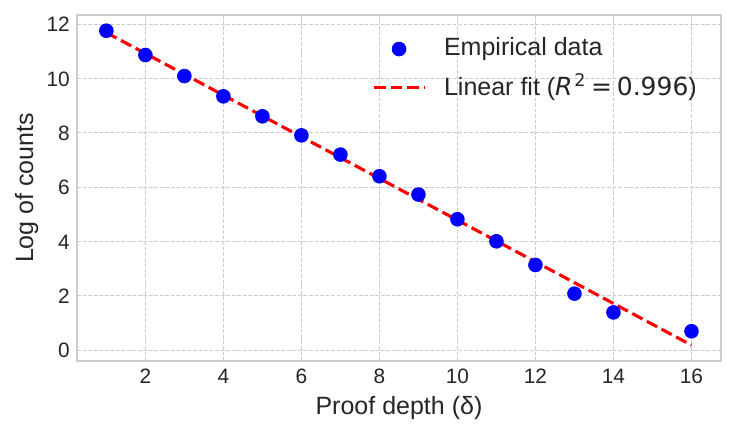}
            \caption{Variable-size RP generator.}
            \label{fig:log_linear_plot}
        \end{subfigure}
        \hfill
        \begin{subfigure}{0.48\textwidth}
            \includegraphics[width=\linewidth]{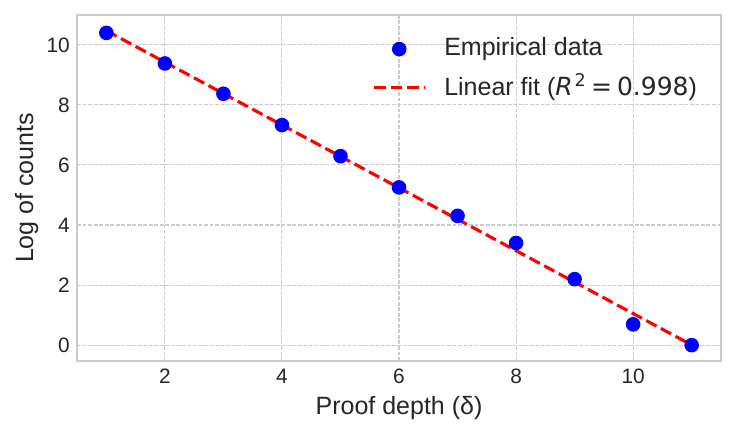}
            \caption{Fixed-size RP generator.}
            \label{fig:log_linear_plot_det}
        \end{subfigure}
        \caption{Semi-log plots of counts per proof depth ($\delta \ge 1$).
        Both the variable (left) and deterministic setting (right) exhibit linear behavior on a log scale, indicating exponential decay.}
        \label{fig:exponential_decay}
    \end{figure}

    Log-linear analysis on the baseline variable-size generator (Figure~\ref{fig:log_linear_plot})
    reveals a linear relationship between proof depth and the logarithm of the counts ($R^2 > 0.99$),
    confirming an exponential decay distribution.
    To verify this distribution is not an artifact of randomized hyperparameters,
    we fixed the predicate and rule counts to 30 and the number of facts to 3 for all samples, removing variance in problem size.
    Figure~\ref{fig:log_linear_plot_det} illustrates the results of this fixed-parameter setting.
    The distribution retains its exponential decay characteristics.
    This confirms that the scarcity of deep proofs is intrinsic to the topology of randomly generated implication graphs,
    rather than a side effect of the distribution of problem sizes.
    Because queries are selected randomly from the set of all predicates,
    this means this bias permeates the entire graph structure.
    Consequently, deep reasoning chains are statistical outliers within the graph topology itself.

    \newpage

    \section{Separable subspaces}
    \label{sec:appendix_separable_subspaces}

    \begin{minipage}[t]{0.58\textwidth}
        The distribution of RMSNorm weights suggests a potential mechanism for how the model manages these subspace projections to enhance expressivity.
        Specifically, the dimension-wise scaling and inversion applied by the learnable weights of the RMSNorm could serve to reweight feature dimensions,
        which is crucial for the expressivity of the multi-head attention layer that follows it~\citep{brody2023expressivity} (Figure~\ref{fig:rmsnorm_weight_distribution}).
        This hypothesis is consistent with our observation that weight decay on these parameters was essential for convergence (\refAppendix{sec:appendix_norm_ablation}).
        By suppressing specific dimensions near zero,
        this structure theoretically helps mitigate inference noise from irrelevant superimposed features~\citep{elhage2022superposition}.
    \end{minipage}
    \hfill
    \begin{minipage}[t]{0.40\textwidth}
        \vspace{-20px}
        \centering
        \includegraphics[width=\textwidth]{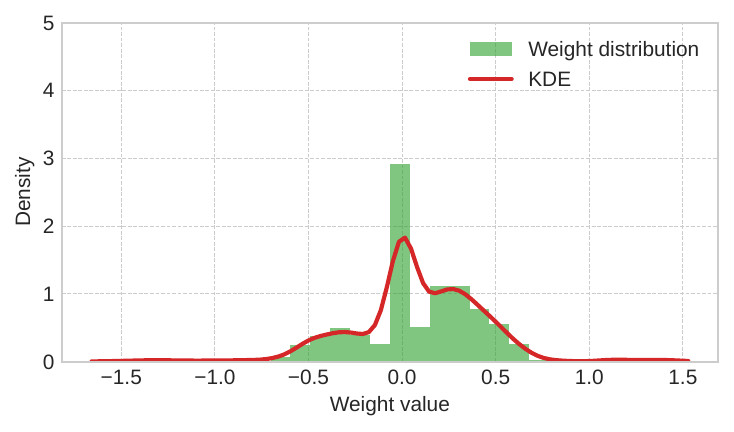}
        \captionof{figure}{A density distribution of \mbox{RMSNorm} weights across all modules.}
        \label{fig:rmsnorm_weight_distribution}
    \end{minipage}

    Consequently, this theoretically aids in isolating the lower-dimensional
    subspace required to make target features separable for subsequent attention operations,
    conceptually aligning with recent findings that effective reasoning processes reduce intrinsic dimensionality~\citep{prasad2026effective}.
    We note that this remains a correlational observation of the network's weight distribution within a largely unexplored territory.

    \section{Procrustes alignment}
    \label{sec:appendix_procrustes_alignment}

    Intuitively, the model's layer-wise computation resembles solving a Rubik's cube.
    The network applies linear transformations --- rotating, scaling (and reflecting) the representation space to gain a better \textit{view} ---
    while executing non-linear operations, the actual \textit{turns}, to progress toward the solution.
    Isolating these operations via orthogonal Procrustes analysis assumes the following:
    \begin{itemize}[noitemsep, topsep=2pt, parsep=2pt, partopsep=0px]
        \item[-] Logical properties (such as facts or provability) are linearly decodable.
        \item[-] The logic state evolution (deduction) is non-linear.
        \item[-] The logic state remains topologically consistent (i.e., facts remain facts, provable statements remain provable).
    \end{itemize}

    To isolate the non-linear transformations corresponding to logical state changes,
    we first align the intermediate hidden states to the target layer's representational space
    using an orthogonal transformation derived via Procrustes analysis on a downsampled calibration subset of the training dataset.
    Specifically, we find the orthogonal matrix $R^{(\ell)}$
    that minimizes the Frobenius norm $\|H^{(\ell)} R^{(\ell)} - H^{(\ell^*)}\|_F$,
    where $H^{(\ell)}$ and $H^{(\ell^*)}$ are the concatenated hidden states at layer $\ell$ and the target layer $\ell^*$,
    respectively.
    We then compute the aligned hidden states for evaluation input sequences $H^{(\ell)}$ as $\tilde{H}^{(\ell)} = H^{(\ell)} R^{(\ell)}$ across the validation dataset.

    \section{Tracing the decision-making process}
    \label{sec:appendix_decision_trace}

    For a token with hidden state $\vec{h}^{(\ell)}$ at layer $\ell$,
    we compute a hypothetical confidence score by projecting the state into the output space of the final layer using Procrustes alignment (\refAppendix{sec:appendix_procrustes_alignment}).
    We then calculate the logits for the provable and unprovable classes using the normalized state and apply a softmax:
    \begin{equation*}
        P(\text{correct} | \vec{h}^{(\ell)}) = \left[ \text{softmax}\left( \begin{bmatrix}
                                                                               \text{RMSNorm}(\tilde{\vec{h}}^{(\ell)}) \cdot \vec{w}_{\text{correct}} \\ \text{RMSNorm}(\tilde{\vec{h}}^{(\ell)}) \cdot \vec{w}_{\text{incorrect}}
        \end{bmatrix} \right) \right]_0
    \end{equation*}

    Basically, this is how the model calculates its final output token during direct generation, where we can apply this "probe" for all tokens and layers.


    \newpage

    \section{The \texttt{r2} heuristic}
    \label{sec:appendix_r2_heuristic}
    To create a training set that actively discourages reliance on shortcuts, we employ the \texttt{r2} heuristic,
    a data augmentation strategy that generates minimally-different problem pairs with opposite labels.
    Its goal is to alter the query's provability while preserving its superficial feature distributions.
    The quantitative breakdown of the heuristic's application in Table~\ref{tab:heuristic_stats} highlights how its strategy adapts to the underlying data structure.

    \begin{table}[h!]
        \centering
        \caption{Distribution of applied \texttt{r2} heuristic strategies per dataset.}
        \label{tab:heuristic_stats}
        \begin{adjustbox}{max width=\linewidth}
            \begin{tabular}{@{}cllr@{}}
                \toprule
                Dataset
                & Label                      & Heuristic strategy               & Percentage \\
                \midrule
                \multirow{3}{*}{LP}
                & \multirow{2}{*}{1 $\to$ 0} & {single-rule prune and addition} & 80.3\%     \\
                &                            & {query alteration}               & 19.5\%     \\
                \cmidrule(lr){2-4}
                & \multirow{1}{*}{0 $\to$ 1} & {single-rule addition and prune} & 100\%      \\
                \midrule
                \multirow{2}{*}{RP}
                & \multirow{1}{*}{1 $\to$ 0} & {greedy iterative modification}  & 97.9\%     \\
                \cmidrule(lr){2-4}
                & \multirow{1}{*}{0 $\to$ 1} & {single-rule addition and prune} & 100\%      \\
                \bottomrule
            \end{tabular}
        \end{adjustbox}
    \end{table}

    When transforming a provable problem to unprovable for LP problems,
    the heuristic's primary strategy is a \textbf{single-rule prune}.
    This is a two-step process designed to break the proof while camouflaging the change.
    First, it randomly searches for and removes one rule whose removal renders the query unprovable.
    Second, to balance this change and maintain superficial feature similarity,
    it applies a \textbf{single-rule addition} of a plausible but logically irrelevant distractor rule.
    This distractor is generated using transitive closure logic ($A \to B, B \to C \vdash A \to C$) and is specifically chosen to conclude the query using existing premises, but without restoring the valid proof path.
    For LP problems where no single rule removal can falsify the query,
    the heuristic employs a fallback strategy: \textbf{query alteration},
    where it searches for an alternative predicate in the graph that is not provable and sets it as the new query.

    For the RP problems,
    which often cannot be broken by a single-rule removal,
    the heuristic uses a \textbf{greedy iterative modification} strategy.
    This function iteratively removes rules or facts (up to 100 iterations),
    at every step greedily prioritizing the removal that maximizes the logical depth of the remaining graph structure.
    Crucially, every removal is paired with a compensatory addition:
    \begin{itemize}[noitemsep, topsep=2pt, parsep=2pt, partopsep=0px]
        \item[-] If a rule is removed, a new \textbf{balancing rule} is added (80\% probability).
        To preserve superficial statistics,
        this rule implies the same conclusion as the removed rule but uses disjoint, high-depth premises that do not satisfy the proof.
        \item[-] If a fact is removed, a new \textbf{balancing fact} is added (90\% probability),
        selected to maximize the logical depth of the remaining graph.
    \end{itemize}
    These stochastic thresholds are set below 100\% because some structures simply cannot be made unprovable (e.g., consider a 5-predicate problem with 5 facts).
    In cases where the heuristic attempts but fails to find a valid balancing component with different label (after 100 iterations),
    no new component is added while the unaltered initial problem instance is retained.

    When transforming an unprovable problem to provable for both LP and RP,
    the heuristic's goal is to introduce a valid proof path while maintaining the problem's complexity by replacing, rather than simply adding, rules.
    This is achieved by first applying a \textbf{single-rule addition} that completes a proof path for the query.
    For RP problems, this addition is specifically constrained to use premises that match the problem's logical depth $\delta$.
    To balance this, the heuristic applies a \textbf{single-rule prune}, where it removes one pre-existing rule that also concludes the query,
    effectively swapping one path for another.

    \newpage

    \section{Mitigating superficial and structural features}
    \label{sec:appendix_correlation_analysis}

    We analyzed how the \texttt{r2} heuristic affects statistical features that a model might exploit as shortcuts.
    We distinguish between \textbf{superficial features} (linear counts accessible via simple summation)
    and \textbf{structural features} (compositional properties requiring non-linear calculation).
    Despite their difference in computational complexity,
    both can exhibit strong correlations with the label, offering a gradient for shortcut learning.

    Table~\ref{tab:correlation_analysis} summarizes these correlations.
    The \textbf{original} datasets exhibit strong correlations across both categories.
    Simple superficial features like \texttt{num\_rules} show high correlation ($0.416$),
    but so do structural features like \texttt{ratio\_rules\_facts} ($0.437$).
    This suggests that the original distribution possesses non-causal predictors not just through token frequency,
    but through the shape of the logical graph itself.
    The \textbf{r2-augmented} column reveals the heuristic's underlying strategy.
    It generates counterfactuals with inverse correlations,
    actively punishing any model that learns a monotonic relationship on these features.
    The final outcome is shown in the \textbf{combined} column.

    \begin{table}[h!]
        \centering
        \caption{
            Pearson correlation between statistical features (superficial and structural) and the ground-truth label.
            The \texttt{r2}-augmented data significantly diminishes the correlations present in the final combined dataset.}
        \label{tab:correlation_analysis}
        \begin{adjustbox}{max width=\linewidth}
            \begin{tabular}{@{}lrrrrrr@{}}
                \toprule
                & \multicolumn{3}{c}{LP dataset} & \multicolumn{3}{c}{RP dataset} \\
                \cmidrule(lr){2-4} \cmidrule(lr){5-7}
                Feature                            & Original & r2-augmented & Combined & Original & r2-augmented & Combined \\
                \midrule
                num\_rules                         & 0.416    & -0.410       & 0.004    & 0.277    & -0.247       & 0.019    \\
                num\_facts                         & -0.171   & 0.174        & 0.001    & 0.123    & -0.070       & 0.030    \\
                num\_distinct\_predicates\_rules   & 0.344    & -0.344       & 0.001    & -0.032   & 0.040        & 0.004    \\
                num\_distinct\_predicates\_total   & 0.339    & -0.342       & -0.001   & -0.043   & 0.049        & 0.003    \\
                query\_total\_occurrences          & -0.068   & 0.015        & -0.026   & 0.431    & -0.298       & 0.078    \\
                query\_as\_rule\_conclusion\_count & 0.090    & -0.125       & -0.022   & 0.486    & -0.270       & 0.131    \\
                query\_in\_rule\_premises\_count   & -0.166   & 0.092        & -0.034   & 0.259    & -0.254       & 0.003    \\
                avg\_rule\_premises                & -0.094   & 0.104        & -0.001   & 0.044    & -0.109       & -0.030   \\
                ratio\_rules\_facts                & 0.437    & -0.434       & 0.003    & -0.139   & -0.024       & -0.100   \\
                branching\_factor                  & 0.102    & -0.115       & -0.003   & 0.082    & -0.131       & -0.021   \\
                \bottomrule
            \end{tabular}
        \end{adjustbox}
    \end{table}

    While a low Pearson correlation suggests that a simple linear model cannot exploit these features,
    a powerful non-linear model like a Transformer could still potentially learn compositional patterns from the same features.
    For example, the model might learn to distinguish between original and augmented samples (a non-linear discrimination)
    and then conditionally apply the original statistics to the relevant subset, effectively bypassing the aggregate mitigation.

    \newpage

    \section{Why \texttt{r2} stresses linear models over superficial features}
    \label{sec:appendix_r2_motivation}

    Let $\phi(x)$ denote superficial features, and consider a linear classifier $w^\top \phi(x)$.
    The \texttt{r2} heuristic constructs counterfactual pairs $(C_1,C_2)$ with opposite labels ($y_1\neq y_2$) while matching superficial statistics,
    so $\Delta \phi \coloneqq \phi(C_1)-\phi(C_2)$ is small.
    To separate such a pair with margin $\epsilon$, a linear model must satisfy
    $|w^\top \Delta\phi| \ge \epsilon$.
    By Cauchy--Schwarz,
    \begin{equation}
        \|w\| \;\ge\; \frac{\epsilon}{\|\Delta\phi\|}.
        \label{eq:r2_norm_lb}
    \end{equation}
    Thus, as \texttt{r2} drives $\|\Delta\phi\|$ downward, achieving a fixed margin requires large weight norms.
    Under standard norm control (explicit $\ell_2$ regularization, implicit bias of optimization),
    this yields a margin--generalization tension: either (1) the classifier fails to realize margin on \texttt{r2} near-collisions,
    or (2) it increases $\|w\|$ and becomes sensitive to incidental variations in $\phi$ when problems have different superficial features.

    Consequently, separating \texttt{r2} pairs using only superficial features results in brittle decision boundaries.
    This motivates learning representations that move counterfactual pairs apart by encoding the underlying structural differences
    (i.e., features not linearly accessible from $\phi$), after which a simple linear head suffices.

    \section{Hyperparameters}\label{sec:appendix_hyperparameters}
    Our model employs the \textbf{Llama 3} architecture~\citep{meta2025llama},
    implemented as a decoder-only Transformer in PyTorch and modified to support type embeddings and custom masking.
    \footnote{Since we do not model long contexts in this work, we scaled down the RoPE $\theta$ parameter to the value in original paper~\citep{su2024roformer}.}

    \begin{table}[h]
        \centering
        \begin{minipage}[t]{0.48\textwidth}
            \centering
            \caption{Transformer architecture.}
            \label{tab:arch_specs}
            \begin{adjustbox}{max width=\linewidth}
                \begin{tabular}{@{}ll@{}}
                    \toprule
                    Parameter                        & Value                                           \\
                    \midrule
                    Hidden dim. ($d_{\text{model}}$) & 256                                             \\
                    FFN hidden dim.                  & 768                                             \\
                    Layers ($L$)                     & 8                                               \\
                    Attn. heads ($H$)                & 4 ($d_{\text{model}}/64$)                       \\
                    Head dim. ($d_{\text{head}}$)    & 64                                              \\
                    Vocab size                       & 256 (symbolic)                                  \\
                    \midrule
                    Activation                       & SiLU                                            \\
                    Normalization                    & RMSNorm ($\epsilon_{\text{rms}} \!=\! 10^{-5}$) \\
                    Positional Emb.                  & RoPE ($\theta \!=\! 10^4$)                      \\
                    Dropout                          & 0.0                                             \\
                    \bottomrule
                \end{tabular}
            \end{adjustbox}
        \end{minipage}%
        \hfill
        \begin{minipage}[t]{0.48\textwidth}
            \centering
            \caption{Optimization hyperparameters.}
            \label{tab:opt_specs}
            \begin{adjustbox}{max width=\linewidth}
                \begin{tabular}{@{}ll@{}}
                    \toprule
                    Parameter                  & Value              \\
                    \midrule
                    Batch size                 & 500                \\
                    Epochs                     & 15                 \\
                    Precision                  & float32            \\
                    Grad. clipping             & 3.0 (max norm)     \\
                    \midrule
                    Scheduler                  & Cosine annealing   \\
                    LR start                   & $1 \times 10^{-2}$ \\
                    LR end                     & $1 \times 10^{-4}$ \\
                    Optimizer                  & AdamW              \\
                    Weight decay               & 0.1                \\
                    Betas ($\beta_1, \beta_2$) & $(0.9, 0.99)$      \\
                    \bottomrule
                \end{tabular}
            \end{adjustbox}
        \end{minipage}
    \end{table}

    Best architecture and hyperparameters were found after extensive grid searches.
    A notable deviation from standard practice is the application of weight decay to RMSNorm parameters,
    which we found empirically crucial for this task (\refAppendix{sec:appendix_norm_ablation}).
    Training was performed on a single NVIDIA A100-80GB GPU\@.
    We standardized our approach to use 50k samples per $(\delta,\text{label})$ bucket (total dataset size being 50k $\times$ 2 labels $\times$ 7 depths);
    the application of the \texttt{r2} heuristic doubles this effective size.
    The datasets are pre-generated, the heuristics for every epoch are pre-generated (every epoch has different heuristics) and for training, same data is used.
    Furthermore, the seeds of random number generators are the same across all training runs, so model initialization is the same across runs,
    unless we aggregate accuracies across a range of seeds to get a measure of robustness under different initializations.

    Given the compact model size ($\approx$ 2.2M parameters without FFN) and dataset scale (1.4M samples per epoch),
    full convergence (15 epochs) is achieved in approximately 12 hours.
    When \texttt{r2}-heuristic augmented datasets for every epoch are computed from scratch,
    then using 20 CPU cores it takes roughly another 12 hours.
    When scaling model size, then training takes longer with proportion to scaled amount.
    E.g., 128 layer model training took somewhere around 2 weeks (gradient accumulation and checkpointing slows it down a bit).
    Our experiments were carried out in the HPC Centre of TalTech~\citep{herrmann.btt.m.2025}.

    \newpage

    \section{Normalization ablation}
    \label{sec:appendix_norm_ablation}

    This ablation compares the performance of RMSNorm versus LayerNorm.
    Normalization was applied prior to the attention, and FFN blocks when the FFN component was enabled (indicated by \texttt{+ ffn}).
    Models were trained on RP and evaluated on both in-distribution (RP) and out-of-distribution (LP) problems where $N_{\text{pred}} \le 30$ and logical depth $\delta \le 6$.

    \vspace{1em}
    As shown in Table~\ref{tab:rmsnorm_vs_layernorm}, both normalization methods perform comparably,
    with RMSNorm showing a slight advantage in out-of-distribution direct evaluation when FFN component was enabled.
    In these runs, weight decay ($0.1$) was applied to all model parameters.
    More importantly, Table~\ref{tab:no_wd_on_norm} demonstrates that
    \textbf{disabling weight decay on normalization parameters prevents convergence},
    with accuracies hovering at random chance ($\approx 50\%$) across all configurations.
    All models are with \texttt{r2 + corrective + bidirectional} components.

    \noindent
    \begin{minipage}[t]{0.49\textwidth}

        \vspace{0px}

        \captionof{table}{RMSNorm vs LayerNorm performance \textbf{with} weight decay on normalization parameters.}
        \label{tab:rmsnorm_vs_layernorm}
        \begin{adjustbox}{max width=\linewidth}
            \begin{tabular}{lrrr}
                \toprule
                Model                     & LP (\%)       & RP (train, \%) \\
                \midrule
                direct w. LayerNorm + ffn & 79.9          & 97.4           \\
                direct w. RMSNorm + ffn   & \textbf{92.5} & 99.7           \\
                direct w. LayerNorm       & 88.0          & 98.8           \\
                direct w. RMSNorm         & 87.6          & 98.4           \\
                \midrule
                cot w. LayerNorm + ffn    & 96.2          & 99.8           \\
                cot w. RMSNorm + ffn      & 96.5          & 99.9           \\
                cot w. LayerNorm          & 96.9          & 99.8           \\
                cot w. RMSNorm            & 96.8          & 99.8           \\
                \bottomrule
            \end{tabular}
        \end{adjustbox}

    \end{minipage}
    \hfill
    \begin{minipage}[t]{0.49\textwidth}

        \vspace{0px}

        \captionof{table}{RMSNorm vs LayerNorm performance \textbf{without} weight decay on normalization parameters.}
        \label{tab:no_wd_on_norm}
        \begin{adjustbox}{max width=\linewidth}
            \begin{tabular}{lrrr}
                \toprule
                Model                     & LP (\%) & RP (train, \%) \\
                \midrule
                direct w. LayerNorm + ffn & 49.0    & 51.6           \\
                direct w. RMSNorm + ffn   & 51.5    & 49.4           \\
                direct w. LayerNorm       & 51.5    & 51.7           \\
                direct w. RMSNorm         & 50.1    & 54.2           \\
                \midrule
                cot w. LayerNorm + ffn    & 50.9    & 53.2           \\
                cot w. RMSNorm + ffn      & 53.5    & 54.1           \\
                cot w. LayerNorm          & 49.3    & 49.9           \\
                cot w. RMSNorm            & 52.4    & 54.8           \\
                \bottomrule
            \end{tabular}
        \end{adjustbox}
    \end{minipage}

    \newpage

    \section{Comparison between multi-task approaches}
    \label{sec:appendix_multi_task_comparison}

    To unify direct prediction and step-by-step reasoning, a standard alternative is a \textbf{mixed} curriculum.
    Let $x$ be the logical context and $y$ be the target output (either a binary label or a proof sequence).
    We define task identifiers $\tau_{direct}$ for direct prediction and $\tau_{cot}$ for CoT.
    The mixed curriculum treats these as independent, disjoint samples:
    $ \mathcal{D}_{mixed} = \{(x, \tau_{direct}, y_{direct})\} \cup \{(x, \tau_{cot}, y_{cot})\} $
    This contrasts with our \textbf{corrective} format,
    which concatenates them into a single sequence $(x, \tau_{direct}, y_{direct}, \tau_{cot}, y_{cot})$ mediated by attention masking.
    We implement this concatenated sequence using continuous position IDs,
    without resetting the RoPE for the parallel branches,
    as the primary order-dependence in our tasks lies between the rule premises and the correct conclusion,
    rendering the exact relative distance to the broader context a negligible factor during generation.
    Furthermore, despite the token length imbalance between the direct and CoT targets,
    we found that standard unweighted token-averaging during the cross-entropy loss calculation is sufficient for stable convergence without branch-specific weighting.
    As shown in Table~\ref{tab:baseline_mixed_comparison}, both approaches perform comparably.

    \begin{table}[h!]
        \centering
        \caption{Performance comparison of corrective and mixed data formats.
        All models are trained with \texttt{bidirectional} and \texttt{r2} components on RP problems with logical depth $\delta \le 6$.
        Bold values indicate the highest accuracy within each column.}
        \label{tab:baseline_mixed_comparison}
        \begin{adjustbox}{max width=\linewidth}
            \begin{tabular}{lcccccccccccc}
                \toprule
                & \multicolumn{4}{c}{LP} & \multicolumn{4}{c}{LP*} & \multicolumn{4}{c}{RP (train)} \\
                \cmidrule(lr){2-5} \cmidrule(lr){6-9} \cmidrule(lr){10-13}
                & \multicolumn{2}{c}{$N_{\text{pred}} \le 30$} & \multicolumn{2}{c}{$N_{\text{pred}} \le 60$} & \multicolumn{2}{c}{$N_{\text{pred}} \le 30$} & \multicolumn{2}{c}{$N_{\text{pred}} \le 60$} & \multicolumn{2}{c}{$N_{\text{pred}} \le 30$} & \multicolumn{2}{c}{$N_{\text{pred}} \le 60$} \\
                \cmidrule(lr){2-3} \cmidrule(lr){4-5} \cmidrule(lr){6-7} \cmidrule(lr){8-9} \cmidrule(lr){10-11} \cmidrule(lr){12-13}
                Model
                & \multicolumn{1}{c}{$\delta \le 6$} & \multicolumn{1}{c}{$6 < \delta \le 12$}
                & \multicolumn{1}{c}{$\delta \le 6$} & \multicolumn{1}{c}{$6 < \delta \le 12$}
                & \multicolumn{1}{c}{$\delta \le 6$} & \multicolumn{1}{c}{$6 < \delta \le 12$}
                & \multicolumn{1}{c}{$\delta \le 6$} & \multicolumn{1}{c}{$6 < \delta \le 12$}
                & \multicolumn{1}{c}{$\delta \le 6$} & \multicolumn{1}{c}{$6 < \delta \le 12$}
                & \multicolumn{1}{c}{$\delta \le 6$} & \multicolumn{1}{c}{$6 < \delta \le 12$} \\
                \midrule
                direct w. corrective & \textbf{85.7} & \textbf{62.7} & \textbf{80.9} & \textbf{55.4} & 96.9 & 82.8 & 91.8 & \textbf{73.5} & 98.4 & 72.9 & 96.9 & 74.9 \\
                direct w. mixed      & 85.3          & 58.0          & 80.0          & 51.2          & \textbf{97.6} & \textbf{83.5} & \textbf{92.1} & 70.4 & \textbf{98.5} & \textbf{76.2} & \textbf{97.4} & \textbf{76.6} \\
                direct               & 65.6          & 53.9          & 63.4          & 54.3          & 73.5          & 55.1          & 70.1          & 52.2          & 83.3          & 72.9          & 83.2          & 65.5          \\
                \midrule
                cot w. corrective    & \textbf{97.4} & \textbf{87.1} & \textbf{88.2} & \textbf{60.8} & \textbf{99.7} & \textbf{98.8} & \textbf{94.7} & \textbf{79.7} & \textbf{99.9} & \textbf{96.8} & \textbf{95.4} & \textbf{75.2} \\
                cot w. mixed         & 94.5          & 72.9          & 85.3          & 54.9          & 99.5          & 96.7          & 93.3          & 73.5          & 99.6          & 92.6          & 95.1          & 70.2          \\
                cot                  & 93.2          & 76.2          & 84.1          & 55.9          & 99.4          & 96.4          & 92.5          & 76.2          & 99.3          & 91.0          & 94.9          & 67.9          \\
                \bottomrule
            \end{tabular}
        \end{adjustbox}
    \end{table}

    Because the isolated attention mask prevents cross-branch visibility,
    the forward pass computations for each target branch are identical to processing them as independent samples.
    Furthermore, because the loss is calculated exclusively on the output branches
    (as the randomly generated problem statement lacks learnable causal dependencies),
    the computed gradients are effectively equivalent across both approaches.
    We note two minor optimization differences to keep in mind:
    (1) due to dataset shuffling,
    the corrective format always processes the direct and CoT pairs simultaneously within the same gradient step,
    whereas the mixed method may process them asynchronously;
    and (2) because we use the standard cross-entropy function provided by PyTorch for both of these approaches,
    which computes a global mean over all active tokens in a batch,
    long CoT traces will dominate the loss denominator,
    placing a proportionally smaller relative weight on the individual direct targets compared to processing them in independent,
    direct-only batches.
    Meaning, when training on long contexts,
    weighted token-averaging during the cross-entropy loss calculation might be necessary.

    \newpage

    \section{Factorial ablation}
    \label{sec:appendix_factorial_ablation}

    Training was conducted on RP problems containing a maximum $N_{\text{pred}} \le 30$ with a logical depth of $\delta \le 6$.
    We evaluate these models on expanded problem spaces involving up to $N_{\text{pred}} \le 60$ and logical depths up to 12.
    Table~\ref{tab:marginal_impact} isolates the marginal contribution of specific components
    by calculating the average percentage point difference in accuracy between models with and without each component ($avg_{with} - avg_{without} \pm std_{diff}$).

    \begin{table}[ht!]
        \centering
        \caption{\textbf{Marginal contribution of components.}
        Values denote the average percentage point difference attributable to each component.
        \textcolor{green!70!black}{Green} indicates a statistically significant improvement, while \textcolor{red}{red} indicates a significant degradation ($p < 0.05$).
        A two-tailed paired sample t-test was used to determine if the performance of the model with the component significantly differs from the model without it.
        The t-threshold was calculated using $k-1$ degrees of freedom, where $k$ is the number of configuration pairs.
        }
        \label{tab:marginal_impact}
        \setlength{\tabcolsep}{2pt}
        \begin{adjustbox}{max width=\linewidth}
            \begin{tabular}{lrrrrrrrrrrrr}
                \toprule
                & \multicolumn{4}{c}{LP} & \multicolumn{4}{c}{LP*} & \multicolumn{4}{c}{RP (train)} \\
                \cmidrule(lr){2-5} \cmidrule(lr){6-9} \cmidrule(lr){10-13}
                & \multicolumn{2}{c}{$N_{\text{pred}} \le 30$} & \multicolumn{2}{c}{$N_{\text{pred}} \le 60$}
                & \multicolumn{2}{c}{$N_{\text{pred}} \le 30$} & \multicolumn{2}{c}{$N_{\text{pred}} \le 60$}
                & \multicolumn{2}{c}{$N_{\text{pred}} \le 30$} & \multicolumn{2}{c}{$N_{\text{pred}} \le 60$} \\
                \cmidrule(lr){2-3} \cmidrule(lr){4-5} \cmidrule(lr){6-7} \cmidrule(lr){8-9} \cmidrule(lr){10-11} \cmidrule(lr){12-13}
                Model
                & \multicolumn{1}{c}{$\delta \le 6$} & \multicolumn{1}{c}{$6 < \delta \le 12$}
                & \multicolumn{1}{c}{$\delta \le 6$} & \multicolumn{1}{c}{$6 < \delta \le 12$}
                & \multicolumn{1}{c}{$\delta \le 6$} & \multicolumn{1}{c}{$6 < \delta \le 12$}
                & \multicolumn{1}{c}{$\delta \le 6$} & \multicolumn{1}{c}{$6 < \delta \le 12$}
                & \multicolumn{1}{c}{$\delta \le 6$} & \multicolumn{1}{c}{$6 < \delta \le 12$}
                & \multicolumn{1}{c}{$\delta \le 6$} & \multicolumn{1}{c}{$6 < \delta \le 12$} \\
                \midrule
                direct w. corrective          & $\textcolor{green!70!black}{18.9 \pm 8.4}$ & $\textcolor{green!70!black}{5.3 \pm 4.6}$ & $\textcolor{green!70!black}{15.4 \pm 7.7}$ & $0.6 \pm 1.1$ & $\textcolor{green!70!black}{25.5 \pm 10.7}$ & $\textcolor{green!70!black}{17.6 \pm 9.5}$ & $\textcolor{green!70!black}{21.4 \pm 10.2}$ & $\textcolor{green!70!black}{9.0 \pm 8.0}$ & $\textcolor{green!70!black}{19.6 \pm 16.8}$ & $6.9 \pm 13.6$ & $\textcolor{green!70!black}{18.6 \pm 18.2}$ & $6.7 \pm 15.1$ \\
                direct w. bidirectional       & $\textcolor{green!70!black}{8.4 \pm 7.1}$  & $\textcolor{green!70!black}{2.8 \pm 2.5}$ & $\textcolor{green!70!black}{7.1 \pm 6.7}$ & $\textcolor{green!70!black}{1.0 \pm 0.7}$ & $\textcolor{green!70!black}{9.6 \pm 7.6}$ & $\textcolor{green!70!black}{7.0 \pm 5.9}$ & $\textcolor{green!70!black}{8.9 \pm 6.8}$ & $\textcolor{green!70!black}{4.6 \pm 3.4}$ & $\textcolor{green!70!black}{7.7 \pm 8.7}$ & $3.4 \pm 9.5$ & $\textcolor{green!70!black}{8.8 \pm 10.4}$ & $4.7 \pm 9.3$ \\
                direct w. r2 (w/ corrective)  & $\textcolor{green!70!black}{7.1 \pm 3.0}$  & $\textcolor{green!70!black}{9.2 \pm 2.3}$ & $\textcolor{green!70!black}{7.9 \pm 2.6}$ & $\textcolor{green!70!black}{7.8 \pm 0.9}$ & $2.5 \pm 3.0$ & $\textcolor{green!70!black}{8.1 \pm 3.4}$ & $\textcolor{green!70!black}{5.4 \pm 2.7}$ & $\textcolor{green!70!black}{8.2 \pm 4.7}$ & $-0.5 \pm 1.2$ & $\textcolor{red}{-2.1 \pm 1.1}$ & $0.2 \pm 1.0$ & $2.7 \pm 2.4$ \\
                direct w. r2 (w/o corrective) & $-0.6 \pm 7.8$                             & $2.2 \pm 2.0$                             & $-0.0 \pm 8.0$                             & $\textcolor{green!70!black}{7.8 \pm 1.3}$ & $-10.9 \pm 14.0$ & $-5.7 \pm 4.7$ & $-8.6 \pm 12.8$ & $-3.8 \pm 3.6$ & $\textcolor{red}{-27.5 \pm 14.6}$ & $\textcolor{red}{-23.0 \pm 11.9}$ & $\textcolor{red}{-29.1 \pm 15.8}$ & $\textcolor{red}{-23.6 \pm 12.8}$ \\
                direct w. ffn                 & $-2.4 \pm 5.7$                             & $-0.9 \pm 2.4$                            & $-0.5 \pm 5.4$                             & $-0.4 \pm 1.3$                            & $-3.0 \pm 9.6$ & $0.1 \pm 3.9$ & $-1.3 \pm 8.7$ & $2.0 \pm 5.4$ & $-4.3 \pm 10.4$ & $-3.0 \pm 8.0$ & $-3.2 \pm 11.4$ & $-1.7 \pm 9.9$ \\
                \midrule
                cot w. corrective             & $-0.8 \pm 6.3$                             & $-2.1 \pm 9.4$                            & $-1.2 \pm 5.2$                             & $-0.6 \pm 3.4$                            & $-1.5 \pm 4.2$ & $-2.2 \pm 7.5$ & $-1.6 \pm 4.1$ & $-3.2 \pm 5.2$ & $-0.7 \pm 2.4$ & $-0.5 \pm 4.9$ & $-0.9 \pm 2.1$ & $-1.5 \pm 4.8$ \\
                cot w. bidirectional          & $\textcolor{green!70!black}{6.0 \pm 4.9}$  & $5.7 \pm 7.9$                             & $\textcolor{green!70!black}{6.0 \pm 4.9}$ & $-0.0 \pm 3.3$ & $\textcolor{green!70!black}{4.0 \pm 3.6}$ & $\textcolor{green!70!black}{9.8 \pm 6.4}$ & $\textcolor{green!70!black}{5.7 \pm 3.4}$ & $\textcolor{green!70!black}{6.3 \pm 4.3}$ & $\textcolor{green!70!black}{2.7 \pm 2.1}$ & $\textcolor{green!70!black}{10.7 \pm 3.5}$ & $\textcolor{green!70!black}{4.5 \pm 1.9}$ & $\textcolor{green!70!black}{5.9 \pm 4.4}$ \\
                cot w. r2                     & $\textcolor{green!70!black}{3.7 \pm 4.3}$  & $\textcolor{green!70!black}{7.1 \pm 7.0}$ & $\textcolor{green!70!black}{4.3 \pm 4.7}$ & $\textcolor{green!70!black}{5.1 \pm 3.5}$ & $1.0 \pm 2.3$ & $3.9 \pm 4.8$ & $\textcolor{green!70!black}{2.9 \pm 3.1}$ & $\textcolor{green!70!black}{6.4 \pm 4.2}$ & $0.5 \pm 1.4$ & $\textcolor{green!70!black}{3.8 \pm 3.3}$ & $1.5 \pm 2.2$ & $2.4 \pm 4.0$ \\
                cot w. ffn                    & $-0.8 \pm 6.2$                             & $0.6 \pm 10.0$                            & $1.7 \pm 4.8$                              & $0.7 \pm 3.9$                             & $-1.2 \pm 3.9$ & $-1.0 \pm 7.3$ & $1.1 \pm 3.1$ & $2.7 \pm 4.4$ & $-0.6 \pm 2.2$ & $0.2 \pm 3.8$ & $1.7 \pm 2.2$ & $\textcolor{green!70!black}{3.9 \pm 3.6}$ \\
                \bottomrule
            \end{tabular}
        \end{adjustbox}
    \end{table}

    For completeness, Table~\ref{tab:main_cot_rp} provides the absolute accuracy metrics across all evaluated runs.

    \begin{table}[ht!]
        \centering
        \caption{\textbf{Full results.} Absolute accuracy metrics across all evaluated model configurations. Bold values indicate the highest accuracy within each column.}
        \label{tab:main_cot_rp}
        \begin{adjustbox}{max width=\linewidth}
            \begin{tabular}{lcccccccccccc}
                \toprule
                & \multicolumn{4}{c}{LP} & \multicolumn{4}{c}{LP*} & \multicolumn{4}{c}{RP (train)} \\
                \cmidrule(lr){2-5} \cmidrule(lr){6-9} \cmidrule(lr){10-13}
                & \multicolumn{2}{c}{$N_{\text{pred}} \le 30$} & \multicolumn{2}{c}{$N_{\text{pred}} \le 60$} & \multicolumn{2}{c}{$N_{\text{pred}} \le 30$} & \multicolumn{2}{c}{$N_{\text{pred}} \le 60$} & \multicolumn{2}{c}{$N_{\text{pred}} \le 30$} & \multicolumn{2}{c}{$N_{\text{pred}} \le 60$} \\
                \cmidrule(lr){2-3} \cmidrule(lr){4-5} \cmidrule(lr){6-7} \cmidrule(lr){8-9} \cmidrule(lr){10-11} \cmidrule(lr){12-13}
                Model                                     & $\delta \le 6$ & $6 < \delta \le 12$ & $\delta \le 6$ & $6 < \delta \le 12$ & $\delta \le 6$       & $6 < \delta \le 12$      & $\delta \le 6$       & $6 < \delta \le 12$      & $\delta \le 6$ & $6 < \delta \le 12$ & $\delta \le 6$ & $6 < \delta \le 12$ \\
                \midrule
                direct                                    & 56.8           & 48.2                & 55.3           & 45.2                & 67.1           & 55.4                & 63.4           & 53.0                & 87.9           & 77.1                & 87.6           & 72.3                \\
                direct w. bidirectional                   & 56.8           & 49.3                & 55.0           & 45.2                & 66.6           & 54.6                & 62.8           & 52.4                & 89.2           & 78.1                & 88.9           & 71.7                \\
                direct w. corrective                      & 66.3           & 48.5                & 61.4           & 46.0                & 82.6           & 60.7                & 74.2           & 56.0                & 92.6           & 76.2                & 88.1           & 64.0                \\
                direct w. ffn                             & 56.8           & 49.4                & 55.3           & 45.7                & 67.0           & 57.9                & 64.1           & 55.9                & 87.6           & 77.0                & 87.0           & 70.4                \\
                direct w. r2                              & 58.2           & 50.6                & 56.6           & 53.8                & 56.6           & 50.3                & 55.6           & 50.2                & 55.7           & 46.4                & 50.4           & 42.2                \\
                direct w. bidirectional+corrective        & 81.0           & 55.4                & 74.6           & 47.8                & 95.6           & 74.3                & 87.0           & 61.2                & 98.4           & 73.6                & 97.1           & 69.7                \\
                direct w. bidirectional+ffn               & 62.3           & 50.9                & 62.4           & 47.3                & 78.3           & 61.2                & 75.6           & 64.8                & 91.8           & \textbf{79.2}       & 91.9           & \textbf{78.7}       \\
                direct w. bidirectional+r2                & 65.6           & 53.9                & 63.4           & 54.3                & 73.5           & 55.1                & 70.1           & 52.2                & 83.3           & 72.9                & 83.2           & 65.5                \\
                direct w. corrective+ffn                  & 64.1           & 49.1                & 61.4           & 45.6                & 80.6           & 63.8                & 74.8           & 59.4                & 91.2           & 74.1                & 89.4           & 66.7                \\
                direct w. corrective+r2                   & 77.7           & 60.5                & 72.4           & 54.9                & 89.5           & 72.5                & 83.5           & 66.6                & 93.0           & 73.5                & 89.3           & 64.8                \\
                direct w. ffn+r2                          & 54.2           & 51.6                & 56.2           & 53.1                & 54.0           & 50.4                & 54.3           & 52.5                & 53.9           & 49.8                & 53.2           & 48.0                \\
                direct w. bidirectional+corrective+ffn    & 81.4           & 51.9                & 76.3           & 46.7                & 95.4           & 73.3                & 87.8           & 61.4                & \textbf{98.7}  & 73.4                & 96.6           & 66.7 \\
                direct w. bidirectional+corrective+r2     & 85.7           & \textbf{62.7}       & 80.9           & \textbf{55.4}       & 96.9           & \textbf{82.8} & 91.8 & \textbf{73.5} & 98.4 & 72.9 & 96.9 & 74.9 \\
                direct w. bidirectional+ffn+r2            & 52.4           & 50.5                & 51.6           & 53.6                & 51.2           & 50.6                & 51.6           & 56.0                & 53.6           & 50.6            & 52.1           & 43.1            \\
                direct w. corrective+ffn+r2               & 71.2           & 56.5                & 70.3           & 52.4                & 80.8           & 67.2                & 77.7           & 61.1                & 88.9           & 71.0                & 88.3 & 71.1 \\
                direct w. bidirectional+corrective+ffn+r2 & \textbf{86.7}  & 62.0                & \textbf{81.8}  & 54.5                & \textbf{97.1} & 82.0 & \textbf{92.1} & 69.7 & 98.6 & 71.3 & \textbf{97.4} & 67.2 \\
                \midrule
                cot                                       & 81.1           & 64.9                & 71.6           & 53.1                & 93.2           & 80.3                & 82.4           & 62.6                & 95.4           & 77.5                & 88.3           & 62.0                \\
                cot w. bidirectional                      & 93.0           & 76.6                & 85.0           & 56.7                & 99.2           & 94.9                & 91.7           & 73.6                & 99.5           & 89.5                & 93.4           & 71.5                \\
                cot w. corrective                         & 90.4           & 73.3                & 79.2           & 54.3                & 96.9           & 88.2                & 86.5           & 65.3                & 98.1           & 82.7                & 89.2           & 64.8                \\
                cot w. ffn                                & 89.8           & 78.0                & 81.7           & 56.9                & 97.0           & 89.0                & 89.4           & 72.2                & 97.7           & 81.9                & 93.2           & 69.1                \\
                cot w. r2                                 & 94.5           & 84.0                & 86.3           & 61.4                & 99.0           & 94.4                & 92.2           & 75.4                & 99.1           & 87.9                & 94.7           & 70.2                \\
                cot w. bidirectional+corrective           & 94.2           & 75.2                & 83.9           & 52.5                & 99.4           & 95.6                & 91.8           & 69.7                & 99.7           & 92.8 & 95.9 & 71.1 \\
                cot w. bidirectional+ffn                  & 94.5           & 79.8                & 87.2           & 55.6                & 99.3           & 97.4                & 93.4           & 74.5                & 99.7           & 92.1                & 97.5           & 74.3                \\
                cot w. bidirectional+r2                   & 93.2           & 76.2                & 84.1           & 55.9                & 99.4           & 96.4                & 92.5           & 76.2                & 99.3           & 91.0                & 94.9           & 67.9                \\
                cot w. corrective+ffn                     & 83.4           & 68.0                & 77.1           & 54.5                & 91.5           & 80.2                & 84.7           & 67.0                & 94.8           & 78.5                & 90.7           & 68.0                \\
                cot w. corrective+r2                      & 94.0           & 83.2                & 83.6           & \textbf{62.8}       & 98.0           & 93.2                & 88.0           & 70.8                & 98.3           & 83.6                & 91.0 & 63.1 \\
                cot w. ffn+r2                             & 94.3           & 83.7                & 87.3           & 61.3                & 99.0           & 94.9                & 93.3           & 79.4                & 98.9           & 87.8                & 93.2           & 73.6                \\
                cot w. bidirectional+corrective+ffn       & 93.5           & 78.8                & 85.8           & 56.1                & 99.1           & 96.2                & 92.8           & 73.5                & 99.7 & 92.5 & 95.4 & 71.0 \\
                cot w. bidirectional+corrective+r2        & 97.4           & 87.1                & 88.2           & 60.8                & 99.7           & 98.8                & 94.7           & 79.7                & 99.9           & 96.8 & 95.4 & 75.2 \\
                cot w. bidirectional+ffn+r2               & \textbf{97.5}  & \textbf{88.5}       & \textbf{90.2}  & 61.9                & 99.8           & \textbf{99.0} & \textbf{95.5} & \textbf{82.9} & \textbf{99.9} & \textbf{97.0} & \textbf{97.8} & \textbf{78.5} \\
                cot w. corrective+ffn+r2                  & 83.0           & 65.3                & 77.9           & 55.9                & 89.2           & 78.2                & 84.6           & 66.2                & 93.7           & 80.2                & 91.4           & 67.0                \\
                cot w. bidirectional+corrective+ffn+r2    & 95.5           & 83.6                & 88.4           & 60.6                & \textbf{99.8}  & 98.4                & 94.7 & 79.0 & 99.7 & 93.6 & 97.1 & 75.1 \\
                \bottomrule
            \end{tabular}
        \end{adjustbox}
    \end{table}

    \newpage

    \section{Training budget matched \texttt{r2} ablation}\label{sec:appendix_sample_matched_r2_ablation}

    To separate the effect of the \texttt{r2} heuristic from the benefit of an increased training budget,
    we conducted an ablation with a matched number of samples.
    Our standard baseline (without \texttt{r2}) uses 50k samples per $(\delta, \text{label})$ bucket.
    Because the \texttt{r2} augmentation generates a counterfactual for each sample, it doubles the effective size to 100k samples per bucket.
    To control for this, we introduce an \texttt{r2}-half configuration,
    where we subsample the base dataset down to first 25k samples per bucket before applying the \texttt{r2} doubling.
    This brings the effective dataset size back to 50k samples per bucket, matching the sample count of the baseline.

    All datasets and epoch-specific heuristics were pre-generated.
    We also kept random seeds consistent across runs so that model initialization remains identical,
    leaving the \texttt{r2} balancing effect as the only varying factor.

    \begin{table}[ht!]
        \centering
        \caption{Training budget matched ablation of the \texttt{r2} heuristic.
        All models have \texttt{corrective} and \texttt{bidirectional} components present,
            ablated with the addition of \texttt{r2} component,
            where \texttt{r2}-half has same amount of samples as the baseline.
            \textcolor{green!70!black}{Green} indicates a statistically significant improvement ($p < 0.05$).
            A one-tailed paired sample t-test was used to determine if model with \texttt{r2} outperforms the model without \texttt{r2},
            where t-threshold was calculated using $k-1$ degrees of freedom to account for small sample size of 3 seeds.}
        \label{tab:r2_ablation_false}
        \resizebox{\textwidth}{!}{
            \begin{tabular}{lcccccccccccc}
                \toprule
                & \multicolumn{4}{c}{LP} & \multicolumn{4}{c}{LP*} & \multicolumn{4}{c}{RP (train)} \\
                \cmidrule(lr){2-5} \cmidrule(lr){6-9} \cmidrule(lr){10-13}
                & \multicolumn{2}{c}{$N_{\text{pred}} \le 30$} & \multicolumn{2}{c}{$N_{\text{pred}} \le 60$}
                & \multicolumn{2}{c}{$N_{\text{pred}} \le 30$} & \multicolumn{2}{c}{$N_{\text{pred}} \le 60$}
                & \multicolumn{2}{c}{$N_{\text{pred}} \le 30$} & \multicolumn{2}{c}{$N_{\text{pred}} \le 60$} \\
                \cmidrule(lr){2-3} \cmidrule(lr){4-5} \cmidrule(lr){6-7} \cmidrule(lr){8-9} \cmidrule(lr){10-11} \cmidrule(lr){12-13}
                Model                                  & $\delta \le 6$ & $6 < \delta \le 12$                        & $\delta \le 6$                            & $6 < \delta \le 12$                       & $\delta \le 6$                                     & $6 < \delta \le 12$                                   & $\delta \le 6$                                    & $6 < \delta \le 12$                                   & $\delta \le 6$ & $6 < \delta \le 12$ & $\delta \le 6$ & $6 < \delta \le 12$ \\
                \midrule
                direct baseline                        & $76.9 \pm 5.8$ & $54.5 \pm 0.8$                             & $72.3 \pm 3.1$                            & $48.5 \pm 0.9$                            & $94.5 \pm 2.9$ & $78.0 \pm 3.6$ & $86.7 \pm 3.1$ & $66.3 \pm 4.4$ & $97.5 \pm 1.5$ & $75.1 \pm 1.5$ & $96.0 \pm 1.6$ & $70.7 \pm 0.9$ \\
                direct w. \texttt{r2}-half             & $85.1 \pm 1.2$ & $61.9 \pm 2.2$                             & $79.1 \pm 2.2$                            & $54.1 \pm 2.3$                            & $95.9 \pm 0.3$ & $78.7 \pm 4.7$ & $89.3 \pm 1.6$ & $66.8 \pm 4.3$ & $97.8 \pm 0.3$ & $71.8 \pm 3.9$ & $95.3 \pm 2.0$ & $69.1 \pm 1.3$ \\
                direct w. \texttt{r2}                  & $85.2 \pm 0.7$ & $58.3 \pm 4.2$                             & $79.4 \pm 1.6$                            & $51.8 \pm 3.1$                            & $96.7 \pm 0.3$ & $79.4 \pm 3.2$ & $89.9 \pm 1.7$ & $67.1 \pm 5.8$ & $98.4 \pm 0.3$ & $72.4 \pm 0.6$ & $96.8 \pm 0.2$ & $69.8 \pm 4.9$ \\
                $\Delta$ (\texttt{r2}-half - baseline) & $8.2 \pm 6.4$  & $\textcolor{green!70!black}{7.4 \pm 2.7}$ & $6.8 \pm 4.9$ & $\textcolor{green!70!black}{5.6 \pm 1.5}$ & $1.5 \pm 3.1$ & $0.7 \pm 8.0$ & $2.6 \pm 4.7$ & $0.5 \pm 7.3$ & $0.3 \pm 1.3$ & $-3.3 \pm 4.4$ & $-0.8 \pm 3.2$ & $-1.6 \pm 1.3$ \\
                $\Delta$ (\texttt{r2} - baseline)      & $8.2 \pm 6.0$  & $3.9 \pm 3.7$                              & $\textcolor{green!70!black}{7.0 \pm 3.3}$ & $3.3 \pm 3.7$ & $2.3 \pm 3.1$ & $1.4 \pm 6.2$ & $3.3 \pm 3.3$ & $0.8 \pm 10.0$ & $0.9 \pm 1.8$ & $-2.7 \pm 1.8$ & $0.7 \pm 1.5$ & $-1.0 \pm 5.6$ \\
                \midrule
                cot baseline                           & $91.4 \pm 4.7$ & $71.8 \pm 5.5$                             & $81.8 \pm 3.8$                            & $52.5 \pm 1.4$                            & $98.5 \pm 1.5$ & $93.1 \pm 5.2$ & $91.0 \pm 2.6$ & $70.6 \pm 3.3$ & $99.2 \pm 0.9$ & $90.0 \pm 6.0$ & $95.0 \pm 1.3$ & $70.4 \pm 2.8$ \\
                cot w. \texttt{r2}-half                & $93.7 \pm 1.3$ & $77.9 \pm 4.1$                             & $85.0 \pm 0.7$                            & $57.7 \pm 2.1$                            & $99.3 \pm 0.2$ & $96.4 \pm 1.1$ & $92.4 \pm 0.9$ & $74.0 \pm 1.5$ & $99.5 \pm 0.1$ & $90.6 \pm 2.4$ & $94.7 \pm 0.5$ & $70.0 \pm 1.4$ \\
                cot w. \texttt{r2}                     & $96.4 \pm 1.0$ & $82.9 \pm 3.8$                             & $87.5 \pm 0.9$                            & $59.8 \pm 1.6$                            & $99.6 \pm 0.2$ & $98.1 \pm 0.6$ & $93.8 \pm 1.5$ & $76.6 \pm 2.8$ & $99.8 \pm 0.1$ & $95.2 \pm 1.7$ & $94.3 \pm 1.4$ & $72.0 \pm 3.5$ \\
                $\Delta$ (\texttt{r2}-half - baseline) & $2.4 \pm 4.8$  & $6.1 \pm 6.8$                              & $3.2 \pm 3.9$ & $\textcolor{green!70!black}{5.2 \pm 2.1}$ & $0.8 \pm 1.5$ & $3.3 \pm 5.2$ & $1.4 \pm 3.5$ & $3.4 \pm 4.2$ & $0.3 \pm 0.8$ & $0.7 \pm 5.6$ & $-0.4 \pm 0.9$ & $-0.4 \pm 3.7$ \\
                $\Delta$ (\texttt{r2} - baseline)      & $5.0 \pm 3.9$  & $\textcolor{green!70!black}{11.0 \pm 3.6}$ & $\textcolor{green!70!black}{5.7 \pm 2.9}$ & $\textcolor{green!70!black}{7.3 \pm 0.9}$ & $1.1 \pm 1.3$ & $5.0 \pm 4.7$ & $\textcolor{green!70!black}{2.8 \pm 1.3}$ & $6.1 \pm 4.1$ & $0.6 \pm 0.9$ & $5.2 \pm 4.7$ & $-0.7 \pm 0.2$ & $1.7 \pm 2.3$ \\
                \bottomrule
            \end{tabular}
        }
    \end{table}

    \newpage

    \section{Corrective objective on shallow models}
    \label{sec:appendix_corrective_shallow}

    Training was conducted on RP problems containing a maximum $N_{\text{pred}} \le 30$ with a logical depth of $\delta \le 6$,
    with rules containing 1 to 3 premises.
    Table~\ref{tab:corrective_vs_direct} measures the marginal contribution of corrective objective on shallow models ($L \in \{2, 4\}$).

    \begin{table}[ht!]
        \centering
        \caption{A marginal contribution of the \texttt{corrective} objective of shallow models
        with \texttt{r2} and \texttt{bidirectional} components on logical depth $\delta \in 2, 3, 4, 5, 6$ problems.
        Values denote the average accuracy across 3 seeds, alongside the difference ($\Delta$) attributable to the corrective objective.
        \textcolor{green!70!black}{Green} indicates a statistically significant improvement when using the corrective objective, while \textcolor{red}{red} indicates a significant degradation ($p < 0.05$).
        A two-tailed paired sample t-test was used to determine if the average difference between these matched setups is significantly different from zero.
        We evaluate performance across varying premise counts.}
        \label{tab:corrective_vs_direct}
        \resizebox{\textwidth}{!}{
            \begin{tabular}{lllccccccc}
                \toprule
                Eval $\delta$
                & Model parameters              & Metric               & LP   (1 to 3)                                    & RP (exactly 1)    & RP (1 to 2)       & RP  (1 to 3)        & RP (exactly 2)    & RP (2 to 3) & RP (exactly 3) \\
                \midrule
                \multirow{6}{*}{$\delta = 2$}
                &                               & Direct               & $0.533 \pm 0.025$                             & $0.805 \pm 0.111$ & $0.724 \pm 0.140$ & $0.663 \pm 0.114$ & $0.683 \pm 0.087$ & $0.648 \pm 0.107$ & $0.623 \pm 0.077$ \\
                & $L=2$, $d_{\text{model}}=256$ & Corrective           & $0.551 \pm 0.012$                             & $0.855 \pm 0.005$ & $0.792 \pm 0.013$ & $0.710 \pm 0.019$ & $0.718 \pm 0.009$ & $0.716 \pm 0.009$ & $0.684 \pm 0.033$ \\
                &                               & $\Delta$             & $+0.018$                                      & $+0.050$          & $+0.067$          & $+0.047$          & $+0.035$          & $+0.068$          & $+0.061$          \\
                \cmidrule(lr){2-10}
                &                               & Direct               & $0.560 \pm 0.035$                             & $0.787 \pm 0.221$ & $0.751 \pm 0.208$ & $0.717 \pm 0.195$ & $0.725 \pm 0.181$ & $0.734 \pm 0.171$ & $0.719 \pm 0.152$ \\
                & $L=4$, $d_{\text{model}}=256$ & Corrective           & \textcolor{green!70!black}{$0.711 \pm 0.010$} & $0.966 \pm 0.012$ & $0.951 \pm 0.003$ & $0.921 \pm 0.010$ & $0.926 \pm 0.008$ & $0.932 \pm 0.008$ & $0.890 \pm 0.015$ \\
                &                               & $\Delta$             & $+0.151$                                      & $+0.179$          & $+0.200$          & $+0.205$          & $+0.201$          & $+0.197$          & $+0.171$          \\
                \midrule
                \multirow{6}{*}{$\delta = 3$}
                &                               & Direct               & $0.554 \pm 0.015$                             & $0.769 \pm 0.163$ & $0.701 \pm 0.143$ & $0.648 \pm 0.091$ & $0.655 \pm 0.091$ & $0.628 \pm 0.080$ & $0.595 \pm 0.067$ \\
                & $L=2$, $d_{\text{model}}=256$ & Corrective           & $0.547 \pm 0.008$                             & $0.848 \pm 0.009$ & $0.774 \pm 0.012$ & $0.694 \pm 0.008$ & $0.702 \pm 0.016$ & $0.698 \pm 0.012$ & $0.654 \pm 0.025$ \\
                &                               & $\Delta$             & $-0.007$                                      & $+0.079$          & $+0.074$          & $+0.046$          & $+0.047$          & $+0.070$          & $+0.059$          \\
                \cmidrule(lr){2-10}
                &                               & Direct               & $0.554 \pm 0.025$                             & $0.774 \pm 0.239$ & $0.729 \pm 0.198$ & $0.710 \pm 0.153$ & $0.693 \pm 0.163$ & $0.697 \pm 0.149$ & $0.664 \pm 0.113$ \\
                & $L=4$, $d_{\text{model}}=256$ & Corrective           & \textcolor{green!70!black}{$0.646 \pm 0.034$} & $0.954 \pm 0.015$ & $0.911 \pm 0.008$ & $0.900 \pm 0.002$ & $0.889 \pm 0.004$ & $0.870 \pm 0.009$ & $0.778 \pm 0.053$ \\
                &                               & $\Delta$             & $+0.092$                                      & $+0.180$          & $+0.182$          & $+0.190$          & $+0.196$          & $+0.173$          & $+0.114$          \\
                \midrule
                \multirow{6}{*}{$\delta = 4$}
                &                               & direct               & $0.536 \pm 0.019$                             & $0.762 \pm 0.184$ & $0.697 \pm 0.142$ & $0.644 \pm 0.095$ & $0.641 \pm 0.085$ & $0.598 \pm 0.091$ & $0.585 \pm 0.070$ \\
                & $L=2$, $d_{\text{model}}=256$ & direct w. corrective & $0.544 \pm 0.015$                             & $0.859 \pm 0.006$ & $0.761 \pm 0.006$ & $0.685 \pm 0.014$ & $0.663 \pm 0.026$ & $0.648 \pm 0.012$ & $0.638 \pm 0.038$ \\
                &                               & $\Delta$             & $+0.009$                                      & $+0.098$          & $+0.064$          & $+0.041$          & $+0.023$          & $+0.049$          & $+0.052$          \\
                \cmidrule(lr){2-10}
                &                               & direct               & $0.539 \pm 0.019$                             & $0.768 \pm 0.236$ & $0.717 \pm 0.213$ & $0.693 \pm 0.162$ & $0.672 \pm 0.148$ & $0.655 \pm 0.140$ & $0.656 \pm 0.105$ \\
                & $L=4$, $d_{\text{model}}=256$ & direct w. corrective & \textcolor{green!70!black}{$0.600 \pm 0.011$} & $0.934 \pm 0.003$ & $0.900 \pm 0.014$ & $0.860 \pm 0.003$ & $0.845 \pm 0.008$ & $0.790 \pm 0.023$ & $0.703 \pm 0.071$ \\
                &                               & $\Delta$             & $+0.061$                                      & $+0.166$          & $+0.184$          & $+0.167$          & $+0.173$          & $+0.135$          & $+0.047$          \\
                \midrule
                \multirow{6}{*}{$\delta = 5$}
                &                               & direct               & $0.525 \pm 0.021$                             & $0.742 \pm 0.196$ & $0.691 \pm 0.174$ & $0.627 \pm 0.106$ & $0.657 \pm 0.100$ & $0.617 \pm 0.066$ & $0.562 \pm 0.072$ \\
                & $L=2$, $d_{\text{model}}=256$ & direct w. corrective & $0.536 \pm 0.015$                             & $0.839 \pm 0.006$ & $0.777 \pm 0.010$ & $0.670 \pm 0.018$ & $0.704 \pm 0.027$ & $0.647 \pm 0.027$ & $0.603 \pm 0.022$ \\
                &                               & $\Delta$             & $+0.011$                                      & $+0.097$          & $+0.086$          & $+0.042$          & $+0.047$          & $+0.031$          & $+0.040$          \\
                \cmidrule(lr){2-10}
                &                               & direct               & $0.541 \pm 0.017$                             & $0.766 \pm 0.231$ & $0.715 \pm 0.206$ & $0.691 \pm 0.158$ & $0.655 \pm 0.158$ & $0.652 \pm 0.113$ & $0.605 \pm 0.086$ \\
                & $L=4$, $d_{\text{model}}=256$ & direct w. corrective & $0.576 \pm 0.028$                             & $0.918 \pm 0.006$ & $0.891 \pm 0.012$ & $0.840 \pm 0.011$ & $0.799 \pm 0.015$ & $0.758 \pm 0.030$ & $0.651 \pm 0.058$ \\
                &                               & $\Delta$             & $+0.035$                                      & $+0.152$          & $+0.176$          & $+0.149$          & $+0.144$          & $+0.106$          & $+0.046$          \\
                \midrule
                \multirow{6}{*}{$\delta = 6$}
                &                               & direct               & $0.521 \pm 0.021$                             & $0.748 \pm 0.192$ & $0.714 \pm 0.184$ & $0.623 \pm 0.090$ & $0.657 \pm 0.078$ & $0.580 \pm 0.064$ & $0.549 \pm 0.048$ \\
                & $L=2$, $d_{\text{model}}=256$ & direct w. corrective & $0.532 \pm 0.008$                             & $0.855 \pm 0.007$ & $0.808 \pm 0.006$ & $0.664 \pm 0.013$ & $0.680 \pm 0.026$ & $0.619 \pm 0.014$ & $0.591 \pm 0.025$ \\
                &                               & $\Delta$             & $+0.011$                                      & $+0.107$          & $+0.094$          & $+0.041$          & $+0.024$          & $+0.039$          & $+0.043$          \\
                \cmidrule(lr){2-10}
                &                               & direct               & $0.539 \pm 0.026$                             & $0.767 \pm 0.235$ & $0.735 \pm 0.200$ & $0.668 \pm 0.147$ & $0.668 \pm 0.140$ & $0.622 \pm 0.106$ & $0.574 \pm 0.074$ \\
                & $L=4$, $d_{\text{model}}=256$ & direct w. corrective & $0.572 \pm 0.017$                             & $0.923 \pm 0.008$ & $0.872 \pm 0.018$ & $0.809 \pm 0.010$ & $0.767 \pm 0.015$ & $0.693 \pm 0.030$ & $0.603 \pm 0.068$ \\
                &                               & $\Delta$             & $+0.033$                                      & $+0.156$          & $+0.137$          & $+0.140$          & $+0.099$          & $+0.071$          & $+0.029$          \\

                \bottomrule
            \end{tabular}
        }
    \end{table}

    The performance of these objectives on logical depth $\delta = 5$ and $6$ problems,
    using a shallow model incapable of faithfully executing forward-chaining ($L \le 4$),
    is statistically indistinguishable.
    However, \texttt{corrective} objective provides a more stable convergence indicated by lower variance and higher average accuracy ($\Delta$),
    so we still use it even on shallow models.

    \newpage

    \section{Breakdown of reasoning errors}\label{sec:appendix_error_breakdown}

\begin{figure}[h!]
        \begin{subfigure}{0.48\textwidth}
            \includegraphics[width=\linewidth]{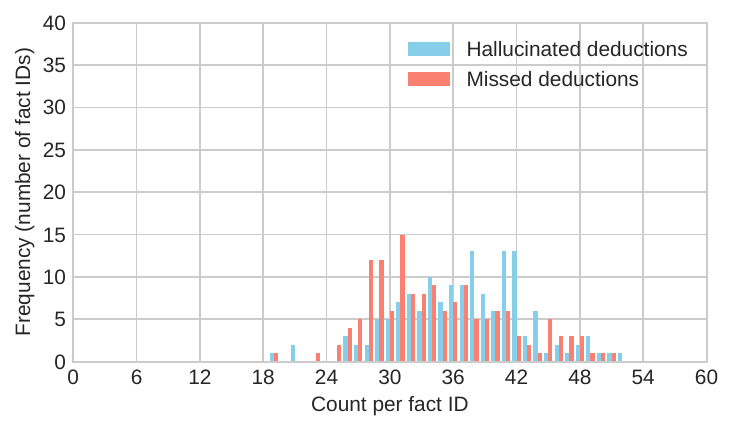}
            \caption{\textbf{Causal} baseline has a high volume of errors.}
            \label{fig:causal_rp_on_lp}
        \end{subfigure}
        \hfill
        \begin{subfigure}{0.48\textwidth}
            \includegraphics[width=\linewidth]{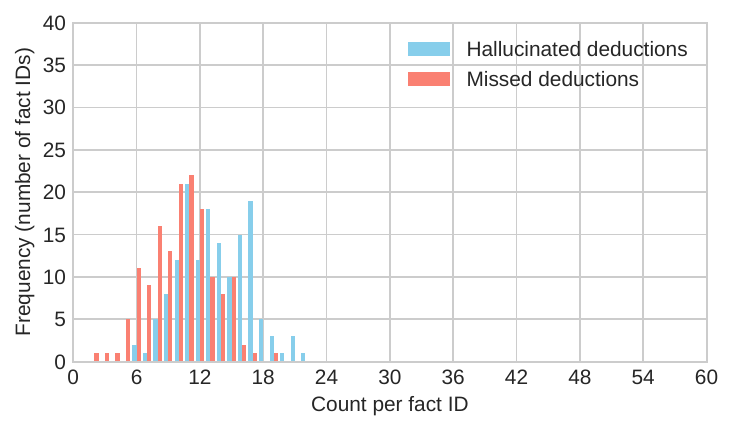}
            \caption{\textbf{Bidirectional} reduces volume, but hallucinations dominate.}
            \label{fig:bidir_rp_on_lp}
        \end{subfigure}

        \vspace{1em}

        \centering
        \begin{subfigure}{0.48\textwidth}
            \includegraphics[width=\linewidth]{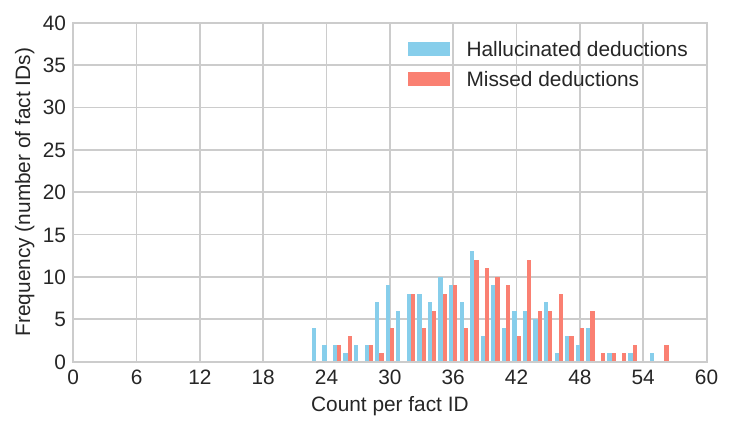}
            \caption{\textbf{r2 + causal} reduces hallucinations.}
            \label{fig:r2_causal_rp_on_lp}
        \end{subfigure}
        \hfill
        \begin{subfigure}{0.48\textwidth}
            \includegraphics[width=\linewidth]{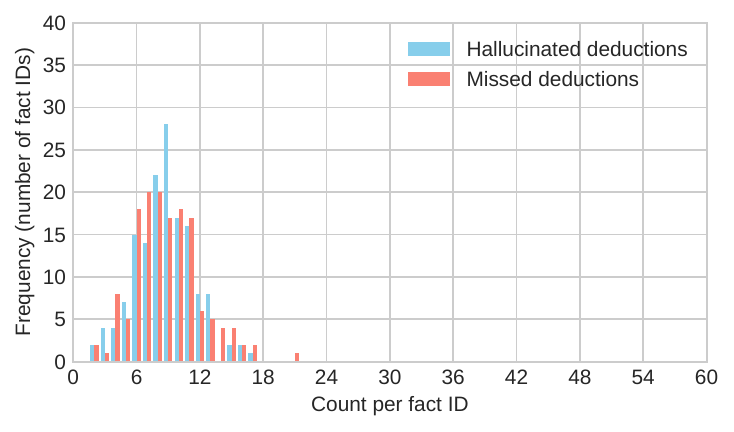}
            \caption{\textbf{r2 + bidirectional} balances the error distribution.}
            \label{fig:r2_bidir_rp_on_lp}
        \end{subfigure}
        \caption{Error breakdown under RP $\to$ LP distribution shift.
        The bidirectional mask (top row) acts on error magnitude, while the r2 heuristic (bottom row) balances the error distribution.}
        \label{fig:error_analysis_breakdown}
    \end{figure}

    To determine the contributions of bidirectional prefix mask and data augmentation,
    we analyze the distribution of CoT reasoning errors during out-of-distribution generalization (train on RP $\to$ eval on LP).
    We categorize errors into \textbf{hallucinated deductions}, where the model derives a fact not supported by the premises; and
    \textbf{missed deductions}, where the model fails to derive a fact that is logically provable.
    While the bidirectional mask uniformly reduces the total error volume,
    it does not mitigate the model's tendency to hallucinate.
    The \texttt{r2} heuristic corrects this imbalance,
    balancing the frequencies of hallucinations and missed deductions.
    Figure~\ref{fig:error_analysis_breakdown} compares these error distributions across mask types and inclusion of \texttt{r2} heuristic.

    \newpage

    \section{Reinforcement learning failure modes}
    \label{sec:appendix_rl_failure}

    To investigate whether reinforcement learning with verifiable rewards could further enhance our best-performing models by further training on the RP dataset,
    we conducted an analysis between \textbf{GRPO}~\citep{grpo} (reward maximization) and \textbf{FlowRL}~\citep{zhu2026flowrl} (distribution matching).
    We evaluated these functions across three distributions (LP, LP*, RP) using four reward configurations.
    RL models were initialized with the optimal supervised hyperparameters for CoT learning.
    Following standard protocols established in recent literature~\citep{grpo}, all models were trained for 3 epochs and
    the reference model $\pi_{ref}$ was updated after each epoch to match the current policy $\pi_\theta$.
    Our analysis explored the intersection of signal density and application granularity.
    We contrasted \textbf{sparse} rewards against \textbf{dense} rewards applied at either the \textbf{sequence level} or \textbf{token level}.
    Sparse rewards used binary $r \in \{0, 1\}$ based solely on final correctness, while dense rewards were augmented with $\pm 0.1$ step-wise verification.
    Sequence-level configurations aggregate the cumulative log-probability of the complete trajectory.
    Token-level configurations normalize this value by the generation length to enforce per-step quality independent of proof duration.

    \begin{figure}[h]
        \vspace{-5px}
        \centering
        \includegraphics[width=0.92\textwidth]{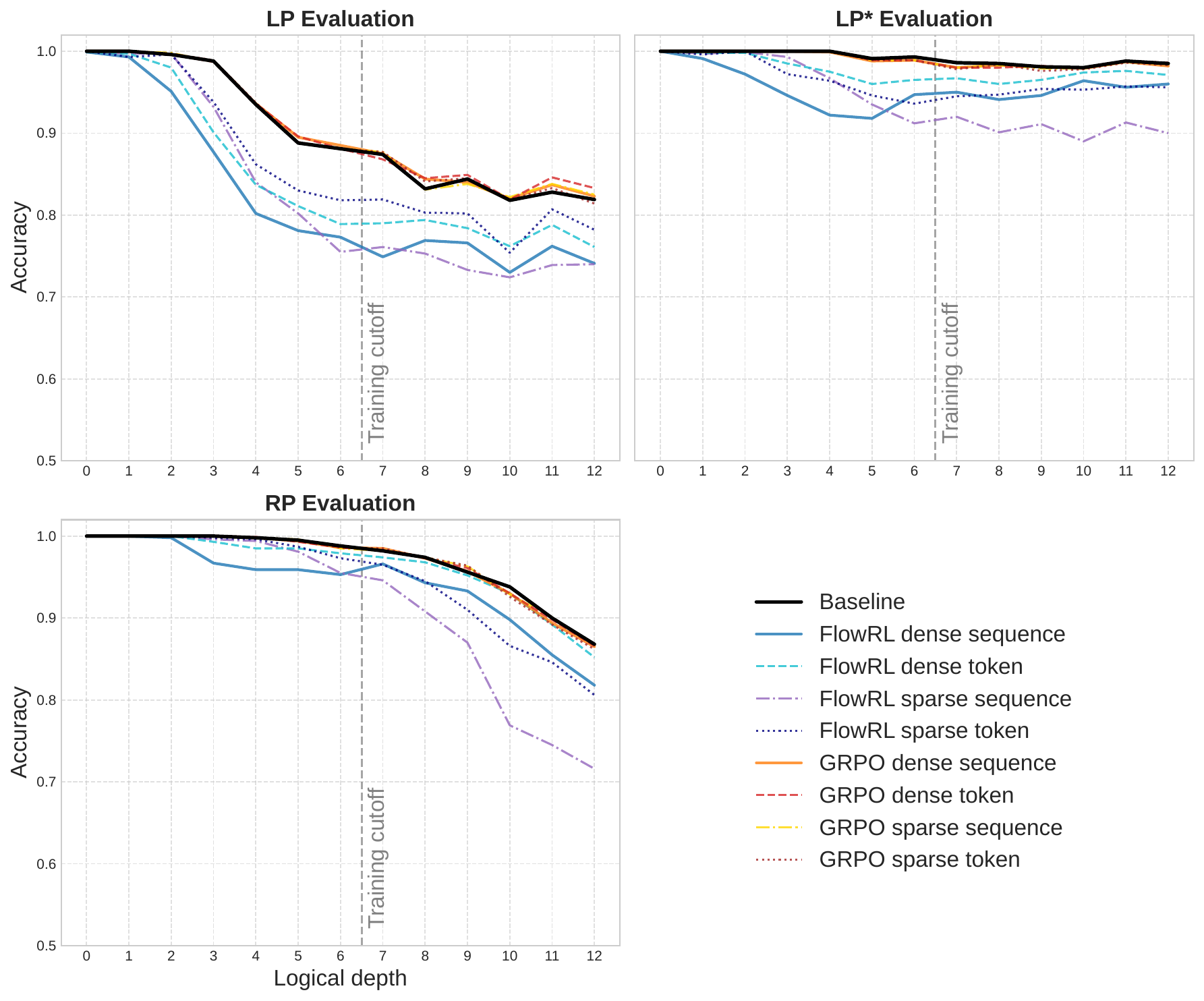}
        \caption{
            Accuracy across logical depth for \textbf{baseline}, \textcolor[HTML]{C0392B}{\textbf{GRPO}}- and \textcolor[HTML]{2980B9}{\textbf{FlowRL}}-trained models.
            The vertical line indicates the maximum depth seen during training.
            All models have \texttt{corrective}, \texttt{r2}, \texttt{bidirectional} and \texttt{ffn} components present.
        }
        \label{fig:rl_performance}
    \end{figure}

    \vspace{-10px}
    Our results reveal that \textbf{GRPO} matched the baseline and
    did not meaningfully improve out-of-distribution generalization to the underlying logic.
    This aligns with the \textbf{support preservation}~\citep{wu2025on},
    where the model also reinforces logically inconsistent computational pathways with successful trajectories.
    Conversely, \textbf{FlowRL} suffers from systematic degradation as logical depth increases.
    Error decomposition attributes this to \textbf{missed deductions},
    consistent with \textbf{empirical-support shrinkage}~\citep{wu2025on},
    where the model sacrifices reasoning coverage in favor of entropy-reduced convergence.

    \newpage

    This failure mode of FlowRL persists across all sparse, dense, token, and sequence-level reward settings,
    suggesting that the performance degradation is intrinsic to the FlowRL objective function rather than an artifact of the reward signal itself.
    To determine the reason behind this degradation,
    we decomposed the error modes into hallucinated and missed deductions,
    following the methodology of \refAppendix{sec:appendix_error_breakdown}.

    \begin{figure}[h]
        \centering
        \begin{minipage}{0.48\textwidth}
            \centering
            \includegraphics[width=\linewidth]{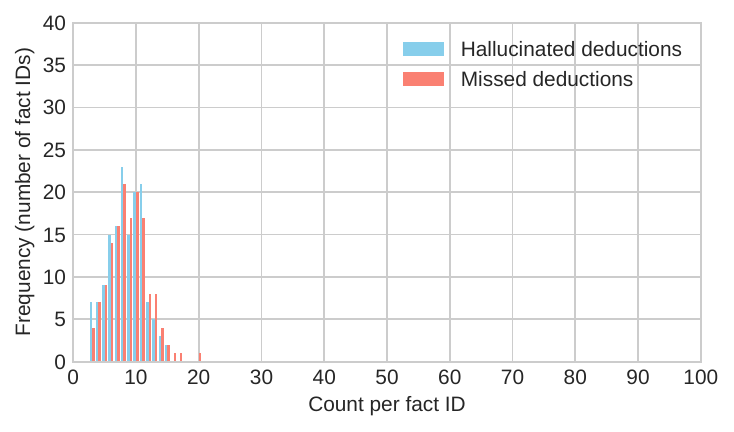}
            \caption{\textbf{GRPO} error profile on LP, maintaining a balanced distribution between hallucinated (blue) and missed deductions (red),
                mirroring the baseline model's error distribution.}
            \label{fig:grpo_errors}
        \end{minipage}\hfill
        \begin{minipage}{0.48\textwidth}
            \centering
            \includegraphics[width=\linewidth]{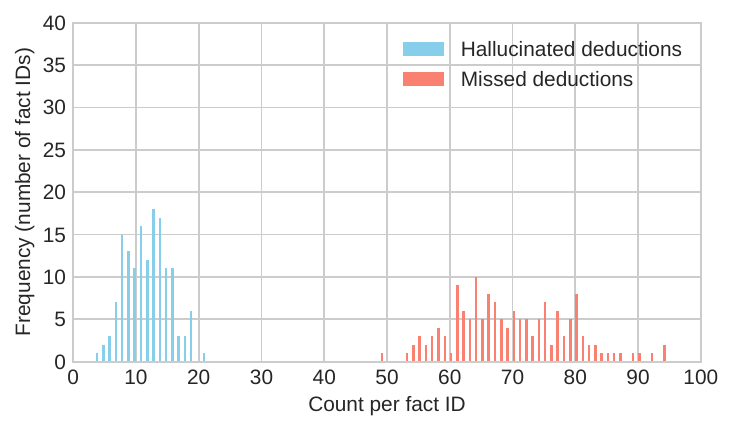}
            \caption{\textbf{FlowRL} error profile on LP, exhibiting a structural shift toward \textbf{missed deductions} (red),
                indicating a failure to notice necessary premises during generation.}
            \label{fig:flowrl_errors}
        \end{minipage}
        \label{fig:rl_errors}
    \end{figure}

    \subsection{Hyperparameters and configurations}
    \label{subsec:rl_reproducibility}

    \begin{minipage}[t]{0.58\textwidth}
        \vspace{0px}
        All models were initialized from the best supervised baseline and trained with a constant learning rate of $1 \times 10^{-6}$.
        Total batch size is 512 which consisted of 64 problems having 8 rollouts each utilizing nucleus sampling ($p=0.95$).

        \vspace{1em}
        \textbf{Reward formulation.}
        The sparse reward is binary, assigning $1.0$ solely for a correct final answer and $0.0$ otherwise.
        The dense reward augments this by verifying the reasoning trace.
        We add $+0.1$ for each unique proven fact that appears in the ground truth path, and penalize invalid facts with $-0.1$.
    \end{minipage}
    \hfill
    \begin{minipage}[t]{0.40\textwidth}
        \vspace{0px}
        \centering
        \captionof{table}{Algorithm-specific hyperparameters.}
        \label{tab:rl_hyperparams}
        \begin{adjustbox}{max width=\linewidth}
            \begin{tabular}{@{}llc@{}}
                \toprule
                Method
                & Parameter                       & Value       \\
                \midrule
                \multirow{2}{*}{GRPO}
                & Clip ratio ($\epsilon$)         & 0.2         \\
                & KL coeff. ($\beta_{\text{KL}}$) & 0.04        \\
                \midrule
                \multirow{4}{*}{FlowRL}
                & Reward scale ($\beta$)          & 15.0        \\
                & Partition $Z_\phi$              & 3-layer MLP \\
                & $Z_\phi ~ d_{\text{model}}$     & 256         \\
                & $Z_\phi$ activation fn.         & SiLU        \\
                \bottomrule
            \end{tabular}
        \end{adjustbox}
    \end{minipage}

    \newpage

    \section{Architecture as inductive bias}
    \label{sec:appendix_recursive_nature}
    In logical deduction, same set of rules is applied recursively until a conclusion is reached.
    To investigate whether aligning the model architecture with problem topology enhances reasoning,
    we experimented with a \textbf{Universal Transformer} configuration~\citep{dehghani2018universal}.
    Rather than stacking $L$ distinct layers, we employ a single Transformer layer applied recursively $K=8$ times.
    We index recurrent iterations by $k \in \{1,\dots,K\}$.
    This introduces a strong inductive bias by pushing the model to learn a single state-transition function that refines the representation towards a solution.
    We compare this recursive universal model against the baseline (8-layer stack) in Table~\ref{tab:appendix_universal_direct_rp}.

    \begin{table}[ht!]
        \centering
        \caption{Baseline (8 distinct layers) comparison against the universal model (1 recursive layer $\times$ 8 iterations).
        Both models have \texttt{corrective}, \texttt{bidirectional} and \texttt{r2} components present, ablated with and without the addition of \texttt{ffn} component.
        \textcolor{green!70!black}{Green} indicates a statistically significant improvement ($p < 0.05$).
        A one-tailed paired sample t-test was used to determine if model with FFN outperforms the model without FFN,
            and whether universal transformer outperforms the baseline transformer, where t-threshold was calculated using $k-1$ degrees of freedom to account for small sample size of 3 seeds.}
        \label{tab:appendix_universal_direct_rp}
        \begin{adjustbox}{max width=\linewidth}
            \begin{tabular}{lcccccccccccc}
                \toprule
                & \multicolumn{4}{c}{LP} & \multicolumn{4}{c}{LP*} & \multicolumn{4}{c}{RP (train)} \\
                \cmidrule(lr){2-5} \cmidrule(lr){6-9} \cmidrule(lr){10-13}
                & \multicolumn{2}{c}{$N_{\text{pred}} \le 30$} & \multicolumn{2}{c}{$N_{\text{pred}} \le 60$} & \multicolumn{2}{c}{$N_{\text{pred}} \le 30$} & \multicolumn{2}{c}{$N_{\text{pred}} \le 60$} & \multicolumn{2}{c}{$N_{\text{pred}} \le 30$} & \multicolumn{2}{c}{$N_{\text{pred}} \le 60$} \\
                \cmidrule(lr){2-3} \cmidrule(lr){4-5} \cmidrule(lr){6-7} \cmidrule(lr){8-9} \cmidrule(lr){10-11} \cmidrule(lr){12-13}
                Model
                & \multicolumn{1}{c}{$\delta \le 6$} & \multicolumn{1}{c}{$6 < \delta \le 12$}
                & \multicolumn{1}{c}{$\delta \le 6$} & \multicolumn{1}{c}{$6 < \delta \le 12$}
                & \multicolumn{1}{c}{$\delta \le 6$} & \multicolumn{1}{c}{$6 < \delta \le 12$}
                & \multicolumn{1}{c}{$\delta \le 6$} & \multicolumn{1}{c}{$6 < \delta \le 12$}
                & \multicolumn{1}{c}{$\delta \le 6$} & \multicolumn{1}{c}{$6 < \delta \le 12$}
                & \multicolumn{1}{c}{$\delta \le 6$} & \multicolumn{1}{c}{$6 < \delta \le 12$} \\
                \midrule
                direct (baseline w. ffn)                        & $89.1 \pm 2.2$                             & $63.4 \pm 3.7$ & $84.0 \pm 2.2$                            & $56.2 \pm 2.5$ & $97.8 \pm 0.7$ & $78.8 \pm 2.8$ & $92.4 \pm 1.1$ & $67.8 \pm 3.0$ & $99.2 \pm 0.5$ & $73.4 \pm 1.8$ & $97.9 \pm 0.6$ & $70.2 \pm 3.9$ \\
                direct (baseline w.o ffn)                       & $85.2 \pm 0.7$                             & $58.3 \pm 4.2$ & $79.4 \pm 1.6$                            & $51.8 \pm 3.1$ & $96.7 \pm 0.3$ & $79.4 \pm 3.2$ & $89.9 \pm 1.7$ & $67.1 \pm 5.8$ & $98.4 \pm 0.3$ & $72.4 \pm 0.6$ & $96.8 \pm 0.2$ & $69.8 \pm 4.9$ \\
                $\Delta$ (w. ffn - w.o ffn)                     & $3.9 \pm 2.9$                              & $5.1 \pm 5.3$  & $4.6 \pm 3.8$                             & $4.4 \pm 5.0$ & $1.1 \pm 0.9$ & $-0.6 \pm 1.2$ & $2.5 \pm 2.1$ & $0.7 \pm 4.0$ & $0.7 \pm 0.6$ & $1.0 \pm 2.2$ & $\textcolor{green!70!black}{1.1 \pm 0.7}$ & $0.4 \pm 7.2$ \\
                direct (universal w. ffn)                       & $95.9 \pm 1.5$                             & $63.4 \pm 3.3$ & $90.5 \pm 1.9$                            & $53.5 \pm 1.5$ & $99.4 \pm 0.1$ & $79.4 \pm 2.0$ & $94.3 \pm 0.1$ & $61.8 \pm 0.8$ & $99.9 \pm 0.0$ & $76.5 \pm 1.4$ & $97.9 \pm 0.6$ & $71.3 \pm 0.9$ \\
                direct (universal w.o ffn)                      & $85.9 \pm 3.6$                             & $60.9 \pm 1.7$ & $80.7 \pm 2.9$                            & $54.2 \pm 1.5$ & $96.6 \pm 1.6$ & $77.8 \pm 8.1$ & $91.5 \pm 1.0$ & $70.6 \pm 6.4$ & $98.1 \pm 1.0$ & $70.2 \pm 0.6$ & $93.0 \pm 3.2$ & $67.4 \pm 4.6$ \\
                $\Delta$ (w. ffn - w.o ffn)                     & $\textcolor{green!70!black}{10.0 \pm 2.3}$ & $2.4 \pm 5.0$ & $\textcolor{green!70!black}{9.8 \pm 1.2}$ & $-0.7 \pm 3.0$ & $\textcolor{green!70!black}{2.8 \pm 1.5}$ & $1.6 \pm 9.4$ & $\textcolor{green!70!black}{2.8 \pm 0.8}$ & $-8.8 \pm 7.2$ & $\textcolor{green!70!black}{1.8 \pm 0.9}$ & $\textcolor{green!70!black}{6.2 \pm 1.5}$ & $4.9 \pm 3.6$ & $3.9 \pm 5.3$ \\
                \midrule
                $\Delta$ (universal w. ffn - baseline w. ffn)   & $\textcolor{green!70!black}{6.8 \pm 3.5}$ & $-0.1 \pm 5.7$ & $\textcolor{green!70!black}{6.5 \pm 3.4}$ & $-2.7 \pm 3.5$ & $\textcolor{green!70!black}{1.6 \pm 0.8}$ & $0.6 \pm 1.5$ & $\textcolor{green!70!black}{1.9 \pm 1.1}$ & $-6.0 \pm 2.3$ & $0.7 \pm 0.5$ & $3.1 \pm 3.2$ & $-0.0 \pm 0.7$ & $1.1 \pm 3.0$ \\
                $\Delta$ (universal w.o ffn - baseline w.o ffn) & $0.8 \pm 3.6$                              & $2.6 \pm 5.5$  & $1.3 \pm 2.4$ & $2.4 \pm 4.6$ & $-0.2 \pm 1.7$ & $-1.6 \pm 10.6$ & $\textcolor{green!70!black}{1.6 \pm 0.8}$ & $3.5 \pm 12.2$ & $-0.4 \pm 1.2$ & $-2.2 \pm 0.3$ & $-3.8 \pm 3.1$ & $-2.4 \pm 2.9$ \\
                \bottomrule
            \end{tabular}
        \end{adjustbox}
    \end{table}

    Under direct evaluation, the universal model significantly outperforms the baseline only when \texttt{ffn} component is present.

\begin{figure}[h!]
        \centering
        \begin{subfigure}{0.48\textwidth}
            \centering
            \includegraphics[width=\linewidth]{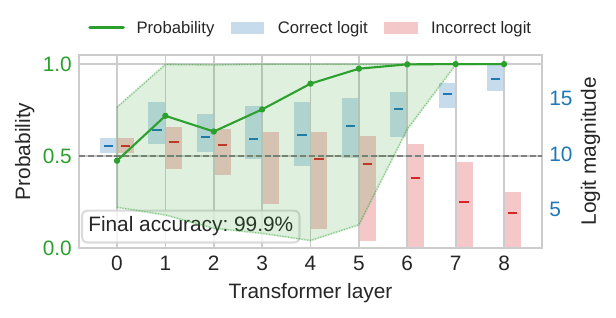}
            \caption{\textbf{Eval on RP.} Premature convergence $k < \delta$.}
            \label{fig:trace_baseline_appendix}
        \end{subfigure}
        \hfill
        \begin{subfigure}{0.48\textwidth}
            \centering
            \includegraphics[width=\linewidth]{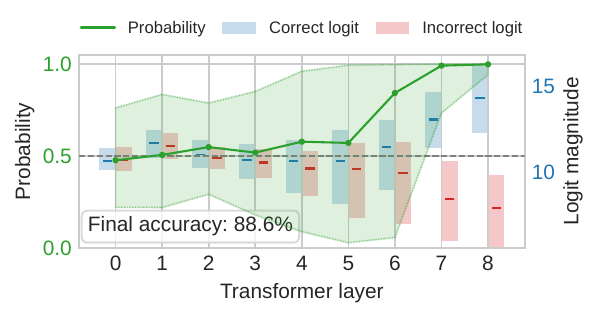}
            \caption{\textbf{Eval on LP.} Delayed convergence $k \ge \delta$.}
            \label{fig:trace_universal_appendix}
        \end{subfigure}
        \caption{Iteration-wise decision traces of the Universal Transformer model
        with \texttt{corrective}, \texttt{bidirectional}, \texttt{r2} and \texttt{ffn} components,
            trained on RP problems under direct evaluation with fixed logical depth $\delta=6$.}
        \label{fig:universal_vs_trace}
    \end{figure}

    Under in-distribution evaluation (Figure~\ref{fig:trace_baseline_appendix}),
    the Universal Transformer shows signs of early convergence ($k < \delta$),
    indicating that recursive bias does not prevent learning shortcuts.
    However, when evaluated on the out-of-distribution LP dataset (Figure~\ref{fig:trace_universal_appendix}),
    convergence is delayed until matching depth ($k \ge \delta$) supporting forward-chaining,
    supporting our theoretical grounding in~\refAppendix{sec:appendix_high_level_analogoies}.

    \newpage

    \section{Scaling trends and out-of-distribution generalization}
    \label{sec:appendix_scaling_trends}

\begin{figure}[htbp]
        \centering
        \begin{subfigure}{\textwidth}
            \centering
            \includegraphics[width=0.48\linewidth]{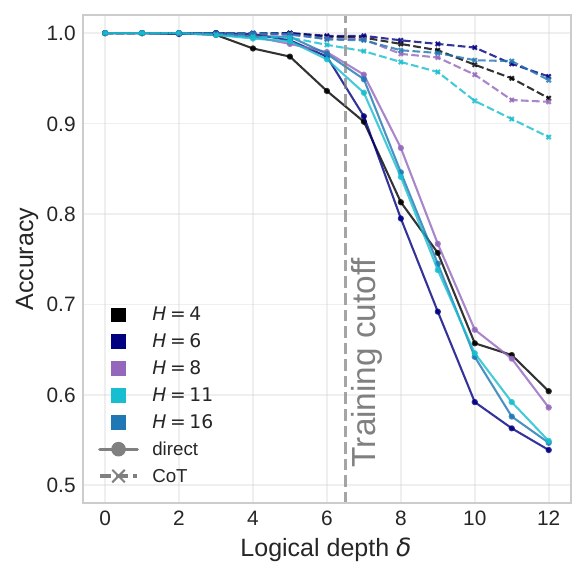}
            \hfill
            \includegraphics[width=0.48\linewidth]{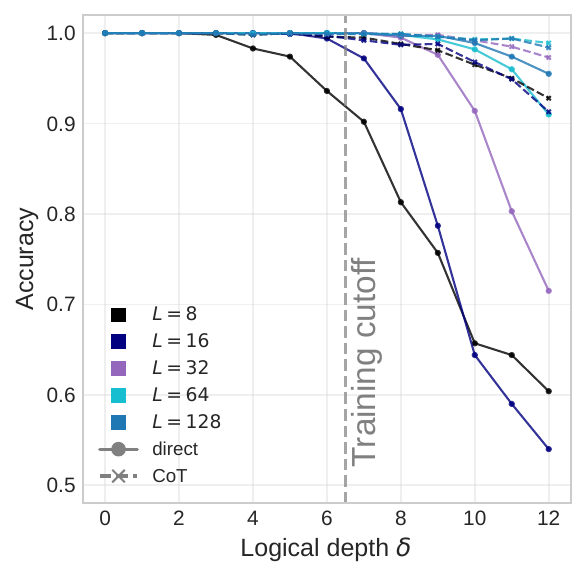}
            \vspace{-10px}
            \caption{Evaluation on RP}
            \label{fig:row1_rp}
        \end{subfigure}
        \vspace{1em}
        \begin{subfigure}{\textwidth}
            \centering
            \includegraphics[width=0.48\linewidth]{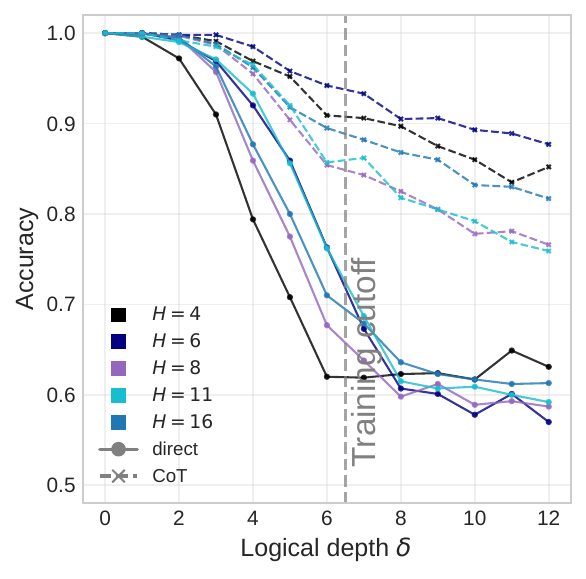}
            \hfill
            \includegraphics[width=0.48\linewidth]{images/final_models_scaling_curves_p30_layers_lp}
            \vspace{-10px}
            \caption{Evaluation on LP}
            \label{fig:row2_lp}
            \vspace{-10px}
        \end{subfigure}
        \caption{Black lines are the baseline model with $L = 8$ layers and $H = 4$ attention heads.
        All models are trained with \texttt{corrective}, \texttt{bidirectional} and \texttt{r2} components.
        Models with same colors across figures are comparable as they have roughly the same amount of parameters.}
        \label{fig:head_scale_master}
    \end{figure}

    Scaling the amount of attention heads ($H$) doesn't close the implicit-explicit gap within training horizon across distributions
    and scaling the number of layers ($L$) is needed (Figure~\ref{fig:head_scale_master}, Table~\ref{tab:gap_closure_summary}).

    \begin{table}[ht!]
        \centering
        \caption{Minimum layers required for direct performance to sustain non-inferiority (within a 1\% margin)
            relative to CoT at all individual logical depths, ensuring the gap closure is robust across problem complexities.
            The reported layer count indicates the threshold at which the reasoning gap closes and remains closed for all subsequent model depths.}
        \label{tab:gap_closure_summary}
        \begin{tabular}{lcccc}
            \toprule
            & \multicolumn{2}{c}{$N_{\text{pred}} \le 30$} & \multicolumn{2}{c}{$N_{\text{pred}} \le 60$} \\
            \cmidrule(lr){2-3} \cmidrule(lr){4-5}
            \textbf{Eval domain}
            & \multicolumn{1}{c}{$\delta \le 6$} & \multicolumn{1}{c}{$6 < \delta \le 12$}
            & \multicolumn{1}{c}{$\delta \le 6$} & \multicolumn{1}{c}{$6 < \delta \le 12$} \\
            \midrule
            LP     & 48 & — & 32 & —   \\
            LP$^*$ & 24 & — & 64 & —   \\
            RP     & 16 & — & 8  & 128 \\
            \bottomrule
        \end{tabular}
    \end{table}

    \newpage
    For reproducibility, Table~\ref{tab:main_direct_rp} reports numerically the accuracies of these scaling runs.
    \begin{table}[ht!]
        \centering
        \caption{Scaling trends under direct and CoT evaluation.
        All models trained with \texttt{corrective}, \texttt{bidirectional} and \texttt{r2} components (no \texttt{ffn} component) on RP samples restricted to $N_{\text{pred}} \le 30$ and a logical depth of $\delta \le 6$,
            and evaluated on problems with depths up to 12 and predicate counts up to 60.
            Column-wise best results are bolded.}
        \label{tab:main_direct_rp}
        \begin{adjustbox}{max width=\linewidth}
            \begin{tabular}{lcccccccccccc}
                \toprule
                & \multicolumn{4}{c}{LP} & \multicolumn{4}{c}{LP*} & \multicolumn{4}{c}{RP (train)} \\
                \cmidrule(lr){2-5} \cmidrule(lr){6-9} \cmidrule(lr){10-13}
                & \multicolumn{2}{c}{$N_{\text{pred}} \le 30$} & \multicolumn{2}{c}{$N_{\text{pred}} \le 60$} & \multicolumn{2}{c}{$N_{\text{pred}} \le 30$} & \multicolumn{2}{c}{$N_{\text{pred}} \le 60$} & \multicolumn{2}{c}{$N_{\text{pred}} \le 30$} & \multicolumn{2}{c}{$N_{\text{pred}} \le 60$} \\
                \cmidrule(lr){2-3} \cmidrule(lr){4-5} \cmidrule(lr){6-7} \cmidrule(lr){8-9} \cmidrule(lr){10-11} \cmidrule(lr){12-13}
                Model
                & \multicolumn{1}{c}{$\delta \le 6$} & \multicolumn{1}{c}{$6 < \delta \le 12$}
                & \multicolumn{1}{c}{$\delta \le 6$} & \multicolumn{1}{c}{$6 < \delta \le 12$}
                & \multicolumn{1}{c}{$\delta \le 6$} & \multicolumn{1}{c}{$6 < \delta \le 12$}
                & \multicolumn{1}{c}{$\delta \le 6$} & \multicolumn{1}{c}{$6 < \delta \le 12$}
                & \multicolumn{1}{c}{$\delta \le 6$} & \multicolumn{1}{c}{$6 < \delta \le 12$}
                & \multicolumn{1}{c}{$\delta \le 6$} & \multicolumn{1}{c}{$6 < \delta \le 12$} \\
                \midrule
                direct w. $L = 8, H=4$ (baseline)              & 85.7           & 62.7          & 80.9          & 55.4          & 96.9           & 82.8          & 91.8          & 73.5          & 98.4           & 72.9          & 96.9          & 74.9          \\
                direct w. $L = 16$                             & 96.6           & 60.3          & 92.2          & 53.7          & 99.4           & 67.9          & 95.6          & 57.7          & 99.9           & 74.1          & 98.7          & 71.4          \\
                direct w. $L = 24$                             & 99.3           & 70.0          & 96.7          & 60.3          & \textbf{100.0} & 76.7          & 98.5          & 63.0          & \textbf{100.0} & 84.9          & 99.5 & 79.5 \\
                direct w. $L = 32$                             & 99.6           & 72.0          & 97.5          & 59.7          & 100.0          & 80.8          & 98.7          & 65.1          & 100.0          & 90.1          & \textbf{99.7} & 83.7 \\
                direct w. $L = 48$                             & 99.9           & 76.2          & 97.6          & 63.6          & 100.0          & 81.7          & 98.8          & 67.6          & 100.0          & 93.5          & 99.5          & 83.0          \\
                direct w. $L = 64$                             & 99.9           & 81.8          & 98.5          & 71.9          & 100.0          & 89.7          & \textbf{99.5} & 77.4          & 100.0          & 97.4          & 99.6 & 87.1 \\
                direct w. $L = 96$                             & 99.8           & 77.8          & 98.4          & 66.2          & 100.0          & 86.4          & 99.4          & 70.5          & 100.0          & 93.2          & 99.6          & 83.5          \\
                direct w. $L = 128$                            & \textbf{100.0} & \textbf{93.2} & \textbf{98.9} & \textbf{78.0} & 100.0 & \textbf{90.9} & 99.2 & \textbf{79.8} & 100.0 & \textbf{98.5} & 99.5 & \textbf{88.9} \\
                direct w. $H = 1$                              & 68.5           & 56.3          & 65.3          & 52.5          & 81.4           & 62.7          & 76.6          & 57.1          & 88.8           & 75.6          & 82.1          & 62.5          \\
                direct w. $H = 2$                              & 76.9           & 59.5          & 75.0          & 56.6          & 93.7           & 85.5          & 86.9          & 73.1          & 94.5           & 73.4          & 92.4          & 70.0          \\
                direct w. $H = 6$                              & 92.9           & 60.5          & 87.2          & 52.4          & 98.3           & 73.7          & 94.1          & 65.4          & 99.5           & 68.1          & 98.7          & 69.7          \\
                direct w. $H = 8$                              & 89.5           & 60.3          & 85.6          & 52.9          & 98.6           & 79.1          & 92.7          & 66.2          & 99.5           & 74.9          & 98.4          & 74.7          \\
                direct w. $H = 11$                             & 93.0           & 61.8          & 88.0          & 55.0          & 97.9           & 75.8          & 92.8          & 65.7          & 99.3           & 71.7          & 97.8          & 71.1          \\
                direct w. $H = 16$                             & 90.6           & 63.0          & 85.5          & 54.4          & 98.3           & 72.6          & 92.0          & 62.7          & 99.5           & 71.7          & 98.3          & 68.2          \\
                direct w. $L = 8, H=4, N_{\text{epochs}} = 30$ & 84.0           & 66.1          & 78.3          & 53.3          & 94.2           & 75.6          & 86.7          & 62.2          & 97.8           & 70.6 & 96.1 & 67.3 \\
                direct w. $L = 8, H=4, N_{\text{epochs}} = 60$ & 87.2           & 64.6          & 77.3          & 54.8          & 96.1           & 73.2          & 86.3          & 59.9          & 98.2           & 68.3 & 93.1 & 62.8 \\
                \midrule
                cot w. $L = 8, H=4$ (baseline)                 & 97.4           & 87.1          & 88.2          & 60.8          & 99.7           & 98.8          & 94.7          & 79.7          & 99.9           & 96.8          & 95.4          & 75.2          \\
                cot w. $L = 16$                                & 98.7           & 83.5          & 92.4          & 61.0          & 99.8           & 99.0          & 95.6          & 79.1          & 99.9           & 96.6          & 95.6          & 76.0          \\
                cot w. $L = 24$                                & 99.6           & 91.0          & 95.7          & 70.1          & 99.9           & 99.6          & 97.8          & 87.9          & \textbf{100.0} & 99.1          & 97.5          & 83.8          \\
                cot w. $L = 32$                                & 99.8           & 92.3          & 95.4          & 71.3          & 99.9           & \textbf{99.7} & 97.9          & 87.2          & 100.0          & 99.1          & 97.2 & 83.5 \\
                cot w. $L = 48$                                & 99.9           & 93.9          & 96.6          & 75.2          & 99.9           & 99.1          & 98.6          & 89.0          & 99.9           & 99.2          & 97.9          & \textbf{86.2} \\
                cot w. $L = 64$                                & \textbf{100.0} & 98.3          & \textbf{97.8} & 81.6          & \textbf{100.0} & 99.5          & \textbf{99.4} & \textbf{93.5} & 100.0 & \textbf{99.5} & \textbf{98.4} & 85.8 \\
                cot w. $L = 96$                                & 99.9           & 96.0          & 96.7          & 77.3          & 100.0          & 99.3          & 98.8          & 90.4          & 100.0          & 99.1          & 97.6          & 83.4          \\
                cot w. $L = 128$                               & 99.9           & \textbf{98.3} & 97.0          & \textbf{83.0} & 100.0          & 99.5          & 98.6          & 91.8          & 100.0 & 99.4 & 97.0 & 81.0 \\
                cot w. $H = 1$                                 & 87.9           & 65.2          & 78.2          & 52.6          & 95.3           & 78.8          & 84.1          & 64.1          & 96.1           & 75.9          & 87.0          & 57.6          \\
                cot w. $H = 2$                                 & 92.4           & 68.3          & 81.6          & 53.6          & 99.1           & 92.7          & 88.8          & 66.3          & 99.1           & 85.5          & 88.5          & 62.0          \\
                cot w. $H = 6$                                 & 98.3           & 90.1          & 90.5          & 64.3          & 99.9           & 99.4          & 96.2          & 83.7          & 100.0          & 98.0          & 96.4          & 75.9          \\
                cot w. $H = 8$                                 & 95.7           & 80.0          & 88.6          & 56.1          & 99.8           & 98.6          & 95.8          & 81.1          & 99.9           & 95.8          & 97.6          & 82.7          \\
                cot w. $H = 11$                                & 96.0           & 80.1          & 90.1          & 60.7          & 99.2           & 96.5          & 94.3          & 77.4          & 99.7           & 93.7          & 96.7          & 77.1          \\
                cot w. $H = 16$                                & 96.6           & 84.8          & 90.5          & 64.5          & 99.8           & 99.0          & 96.5          & 85.2          & 99.9           & 97.3          & 98.3          & 85.1          \\
                cot w. $L = 8, H=4, N_{\text{epochs}} = 30$    & 96.0           & 82.6          & 85.1          & 58.3          & 99.7           & 98.1          & 93.3          & 77.5          & 99.8           & 95.1 & 93.8 & 71.5 \\
                cot w. $L = 8, H=4, N_{\text{epochs}} = 60$    & 96.5           & 86.0          & 85.3          & 62.5          & 99.2           & 96.7          & 89.7          & 73.0          & 99.5           & 94.2 & 89.1 & 65.4 \\
                \bottomrule
            \end{tabular}
        \end{adjustbox}
    \end{table}

    \newpage

    \section{Tractability of learning to predict directly}
    \label{sec:appendix_direct_tractability}

\begin{figure}[htbp]
        \centering
        \begin{subfigure}{\textwidth}
            \centering
            \includegraphics[width=0.48\linewidth]{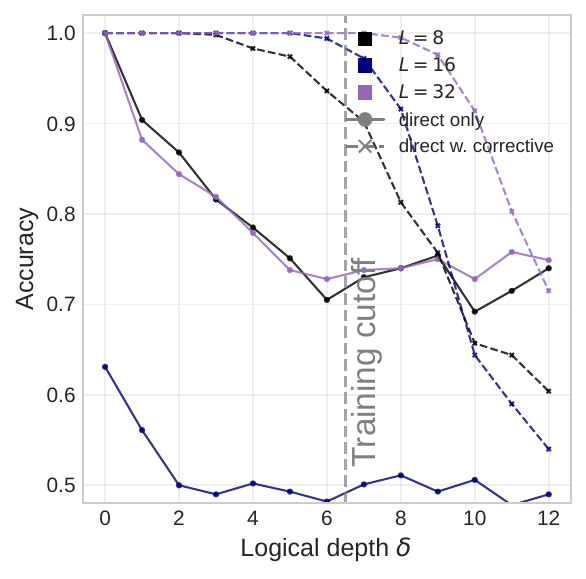}
            \hfill
            \includegraphics[width=0.48\linewidth]{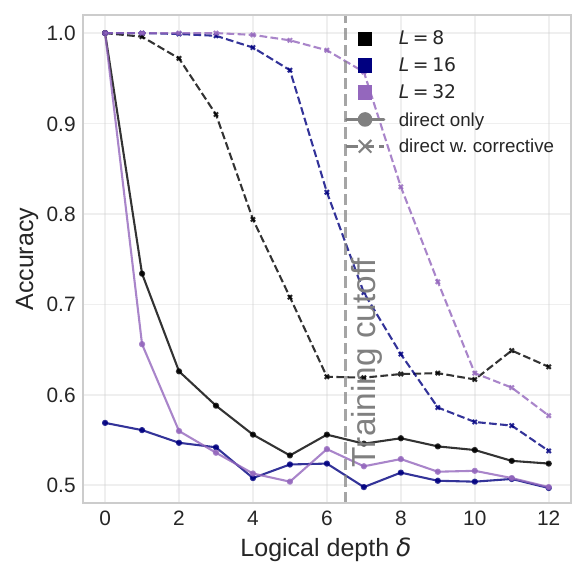}
        \end{subfigure}
        \caption{Direct performance with and without corrective on RP (left) and LP (right) domains.}
        \label{fig:direct_tractability}
    \end{figure}

    To determine whether the corrective objective is required for improvements stemming from increasing Transformer layers $L$,
    we ablated this component across models of varying depth ($L=8, 16, 32$),
    detailed in Figure~\ref{fig:direct_tractability} and Table~\ref{tab:main_direct_rp_tract}.
    Here, deeper models fail to surpass the shallow $L=8$ baseline when corrective component is missing.
    These results confirm that the corrective objective is required to make implicit reasoning tractable.

    \begin{table}[ht!]
        \centering
        \caption{Scaling trends of direct models without \texttt{corrective} component.
        All models trained with \texttt{bidirectional} and \texttt{r2} components on RP samples restricted to $N_{\text{pred}} \le 30$ and a logical depth of $\delta \le 6$,
            and evaluated on problems with depths up to 12 and predicate counts up to 60.
            Column-wise best results are bolded.}
        \label{tab:main_direct_rp_tract}
        \resizebox{\textwidth}{!}{
            \begin{tabular}{lcccccccccccc}
                \toprule
                & \multicolumn{4}{c}{LP} & \multicolumn{4}{c}{LP*} & \multicolumn{4}{c}{RP (train)} \\
                \cmidrule(lr){2-5} \cmidrule(lr){6-9} \cmidrule(lr){10-13}
                & \multicolumn{2}{c}{$N_{\text{pred}} \le 30$} & \multicolumn{2}{c}{$N_{\text{pred}} \le 60$} & \multicolumn{2}{c}{$N_{\text{pred}} \le 30$} & \multicolumn{2}{c}{$N_{\text{pred}} \le 60$} & \multicolumn{2}{c}{$N_{\text{pred}} \le 30$} & \multicolumn{2}{c}{$N_{\text{pred}} \le 60$} \\
                \cmidrule(lr){2-3} \cmidrule(lr){4-5} \cmidrule(lr){6-7} \cmidrule(lr){8-9} \cmidrule(lr){10-11} \cmidrule(lr){12-13}
                Model
                & \multicolumn{1}{c}{$\delta \le 6$} & \multicolumn{1}{c}{$6 < \delta \le 12$}
                & \multicolumn{1}{c}{$\delta \le 6$} & \multicolumn{1}{c}{$6 < \delta \le 12$}
                & \multicolumn{1}{c}{$\delta \le 6$} & \multicolumn{1}{c}{$6 < \delta \le 12$}
                & \multicolumn{1}{c}{$\delta \le 6$} & \multicolumn{1}{c}{$6 < \delta \le 12$}
                & \multicolumn{1}{c}{$\delta \le 6$} & \multicolumn{1}{c}{$6 < \delta \le 12$}
                & \multicolumn{1}{c}{$\delta \le 6$} & \multicolumn{1}{c}{$6 < \delta \le 12$} \\
                \midrule
                direct w. $L = 8$  & \textbf{65.6} & \textbf{53.9} & \textbf{63.4} & 54.3          & \textbf{73.5} & \textbf{55.1} & \textbf{70.1} & \textbf{52.2} & \textbf{83.3} & 72.9 & \textbf{83.2} & 65.5 \\
                direct w. $L = 16$ & 53.9          & 50.4          & 53.1          & \textbf{54.7} & 50.8          & 50.2          & 50.4          & 50.2          & 52.3          & 49.6          & 50.6 & 46.5 \\
                direct w. $L = 32$ & 61.6          & 51.4          & 59.2          & 49.9          & 67.0          & 53.0          & 63.6          & 50.0          & 82.7          & \textbf{74.4} & 82.8 & \textbf{67.8} \\
                \bottomrule
            \end{tabular}
        }
    \end{table}

\end{document}